\pdfoutput=1
\documentclass[11pt]{article}

\usepackage[final]{acl}

\usepackage{times}
\usepackage{booktabs}
\usepackage{latexsym}
\usepackage{pifont}
\usepackage{makecell}
\usepackage{multirow}
\usepackage{spverbatim}
\usepackage[T1]{fontenc}
\usepackage[utf8]{inputenc}
\usepackage{subcaption}
\usepackage{microtype}

\usepackage{inconsolata}

\usepackage{graphicx}
\usepackage{makecell}
\usepackage{float}
\title{Beyond Single Plots: A Benchmark for Question Answering on Multi-Charts}

\author{Azher Ahmed Efat \and Seok Hwan Song \and Wallapak Tavanapong \\
  Department of Computer Science, Iowa State University, Ames, Iowa, USA \\
  \texttt{\{efat, song92, tavanapo\}@iastate.edu}}

\begin{document}
\maketitle
\begin{abstract}
Charts are widely used to present complex information. Deriving meaningful insights in real-world contexts often requires interpreting multiple related charts together. 
\iffalse 
rather than relying on a single one. 

While there has been extensive research on understanding single charts, multi-charts have not been extensively explored. \fi
Research on understanding multi-chart images has not been extensively explored. We introduce PolyChartQA, a mid-scale dataset specifically designed for question answering over {\em multi-chart images}. PolyChartQA comprises 534 multi-chart images (with a total of 2,297 sub-charts) sourced from peer-reviewed computer science research publications and 2,694 QA pairs. We evaluate the performance of nine state-of-the-art Multimodal Language Models (MLMs) on PolyChartQA across question type, difficulty, question source, and key structural characteristics of multi-charts. Our results show a 27.4\% LLM-based accuracy (L-Accuracy) drop on human-authored questions compared to MLM-generated questions, %revealing persistent MLM challenges 
and a 5.39\% L-accuracy gain with our proposed prompting method.
% general-purpose and specialized-chart understanding
\end{abstract}

\section{Introduction}
Charts play a critical role in presenting data, trends, and information in real-world contexts, including scientific research, finance, %healthcare, 
and policy making \cite{graphSlope}. %\cite{cohen2019effectvisualdesignimage,graphSlope}
%As charts contain complex %quantitative 
%information, 
Charts often give rise to questions that require substantial analytical skill, a deep understanding of visual features, precise data extraction, and multi-step reasoning operations based on the chart content \cite{10.1145/3313831.3376467,masry-etal-2022-chartqa,https://doi.org/10.1111/cgf.14573}. % Answering complex questions from these charts requires significant analytical skill, a deep understanding of visual features, precise data extraction, and the ability to perform multi-step reasoning operations \cite{masry-etal-2022-chartqa,https://doi.org/10.1111/cgf.14573}. 
\iffalse
Due to the complexity of chart images compared to natural images, 
%not needed and arguable whether chart images are more complex. Tinychart claim that chart images are less complex.

Existing works on understanding chart images include question answering \cite{masry-etal-2022-chartqa}, summarization \cite{islam-etal-2024-large}, fact checking \cite{akhtar-etal-2023-reading}, chart-to-table translation \cite{liu-etal-2023-deplot}, to name a few.
\fi

Multimodal Language Models (MLMs) have demonstrated strong performance across diverse real-world vision-language tasks, including visual question answering \cite{NEURIPS2023_5abcdf8e,masry-etal-2022-chartqa}, image captioning and generation \cite{10484490,NEURIPS2023_43a69d14}, summarization \cite{islam-etal-2024-large}, and many chart understanding tasks \cite{liu-etal-2023-deplot,akhtar-etal-2023-reading,masry-etal-2025-chartgemma,zhang-etal-2024-tinychart}.

Existing research on chart understanding has been limited to single chart images. However, complex decision-making in real-world contexts also requires an understanding of a multi-chart image with at least two relevant sub-chart images. % Multi-chart images commonly appear in various domains.
Understanding multi-chart images poses unique challenges, requiring integration across related sub-charts, interpretation of diverse visual structures, and resolution of visual ambiguities \cite{https://doi.org/10.1111/cgf.14573}.
\iffalse
To date, MultiChartQA \cite{zhu2024multichartqa} remains the only dataset specifically designed for multi-chart question answering (QA). CharXiv also explores multi-chart settings \cite{CharXiv}.

Pak: Remove from here, but discussed in the related work.
\citeauthor{CharXiv} shows that as the number of sub-charts increases, MLM performance declines, but did not measure the accuracy differences when the same question is posed in a single-chart versus a multi-chart context. {\color{red} Or repeat what was said in the related work: does not explore the influence of multi-chart characteristics or question difficulty}
\fi

\iffalse 
%Pak: May be we move this discussion later. It is unclear how can there be a single chart versus a multi-chart context for the same 

\fi
%While \citet{CharXiv} shows that MLM performance drops as the number of sub-charts increases, they do not explore how the same question is answered differently when presented in a single-chart vs. a multi-chart setting. % \citet{CharXiv} show that MLMs perform significantly better on common chart types (e.g., bar and line charts); however, they do not distinguish between single-chart and multi-chart settings in their analysis.

%To date, MultiChartQA \cite{zhu2024multichartqa} remains the only publicly available dataset specifically designed for multi-chart question answering (QA). CharXiv also explores multi-chart settings \cite{CharXiv}.
MultiChartQA \cite{zhu2024multichartqa} and CharXiv \cite{CharXiv} are two major datasets for evaluation of MLM performance on multi-chart question answering (QA). MultiChartQA has multi-chart sets, each comprising two or three related single- or multi-chart images to simulate multi-chart scenarios. Its questions include explicit chart references (e.g., “first chart”) and answer-format instructions, unlike how end-users phrase questions or how composite figures naturally appear. Performance drops when such references are removed or when actual multi-chart images are  used~\cite{zhu2024multichartqa}. However, the joint effect of these factors on performance relative to single-chart settings has not been investigated. MultiChartQA lacks annotations that describe multi-chart characteristics (e.g., chart type, homogeneity, and the sub-chart count).%, which limits fine-grained analysis.

Similarly, the authors of CharXiv explore descriptive and reasoning questions %—focusing on chart structure or interpretation beyond direct cues—
but omit data retrieval questions, which require precise extraction of numerical values, often under visual ambiguity (e.g., overlapping bars, unclear ticks). It also does not consider  question difficulty or other multi-chart attributes such as homogeneity or chart type, nor provides corresponding annotations. Its questions include positional cues (e.g., “row 1, column 2”) and short, direct answers, diverging from the open-ended nature of real-world multi-chart queries.

\begin{figure*}[]
\centering
\includegraphics[width=\textwidth]
{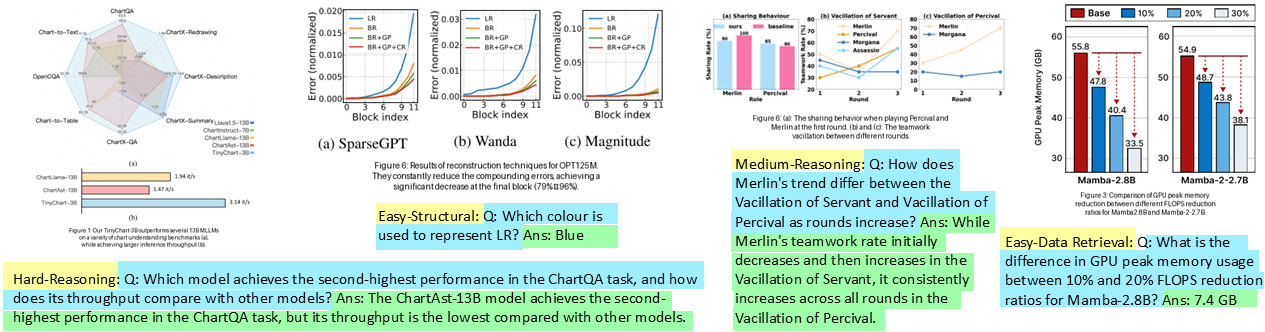}
\caption{Example human-authored QA pairs from PolyChartQA. Multichart images are from \citet{zhang-etal-2024-tinychart,shin-etal-2024-rethinking,lan-etal-2024-llm,zhan-etal-2024-rethinking-token}.}
\label{fig:comb}
\end{figure*}
In addition, due to the scalability limitations of human-authored QA pairs, recent studies have increasingly relied on MLM-generated questions for chart QA tasks \cite{shinoda2024sbsfigurespretrainingfigure,10654963,pramanick2025spiqadatasetmultimodalquestion}. However, this growing reliance on MLM-generated QA raises the question of their suitability for evaluating multi-chart reasoning and how performance differs from human-authored questions in multi-chart contexts. Despite the importance and prevalence of multi-chart images, their complexity remains underexplored in existing research. To bridge this gap, we pose the following research questions.

\begin{itemize}
   \item RQ1: How does MLM performance differ between single-chart and multi-chart images when questions lack explicit chart references?
    \item RQ2: How do question and multi-chart characteristics influence MLMs on multi-chart QA?
    \begin{itemize}
        \item RQ2.1: To what extent does question difficulty (i.e., easy, medium, hard) influence MLM performance?
        \item RQ2.2: Does the question type (structural, data retrieval, or reasoning) affect MLM performance?
        %{\color{red}\item RQ2.3: Is MLM performance sensitive to the number of sub-charts in a multi-chart figure?}
        \item RQ2.3: How does chart type homogeneity (sub-charts are of the same type versus mixed types) affect performance? 
        \item RQ 2.4: Does increasing sub-chart count impact MLM performance?
        %\item RQ2.4: In the case of homogeneous multi-chart figures, how does MLM performance vary by chart type (e.g., bar chart, line chart, scatter plot)?
    \end{itemize}
    \item RQ3. To what extent are human-authored multi-chart questions more challenging for MLMs than MLM-generated ones?%To what extent do human-authored questions over multi-chart images exhibit higher difficulty for MLMs compared to MLM-generated questions?
    \item RQ4. Does stepwise visual decomposition and self-verification prompting improve MLM performance and interpretability in multi-chart QA?%Does incorporating stepwise visual decomposition and self-verification through prompting enhance the accuracy of MLMs in multi-chart QA tasks?
\end{itemize}

To systematically explore these research questions, we introduce \textbf{PolyChartQA}\footnote{\raggedright PolyChartQA Dataset: \url{https://github.com/NRT-D4/PolyChartQA}}, for question answering over multi-chart images. PolyChartQA comprises 534 multi-chart images containing a total of 2,297 sub-charts sourced from peer-reviewed computer science research papers and 2,694 question-answer pairs annotated across diverse question types and difficulty levels. Figure \ref{fig:comb} shows example human-authored QA pairs. 

% PolyChartQA includes eleven chart types.\\ (Human-authored: 519, MLM-generated: 2,175)
%Additionally, our chart images . %, including Bar Chart, Scatter Plot, Spider Chart, Histogram, Line Chart, Box Plot, Point Plot, Surface Plot, Pie Chart, Area Chart, Density Plot, Dot Plot, and Violin Plot. 
% \begin{figure}[!b]
% \vspace{-17pt}
%     \centering
% \includegraphics[width=0.7\columnwidth]{exmp_multi_chart.png}
% \vspace{-6pt}
%     \caption{
%         Multi-chart (source: \citet{10.1145/3597503.3639226})
%     }
%     \vspace{-4pt}
%     \label{exmp_fig}
% \end{figure}

PolyChartQA differs fundamentally from MultiChartQA \cite{zhu2024multichartqa} and CharXiv \cite{CharXiv} in both annotation depth and task formulation. First, PolyChartQA includes explicit multi-chart annotations that are absent in the existing datasets, such as chart type at the sub-chart level, and chart homogeneity. These annotations enable controlled analysis of multi-chart structure and reasoning complexity, which cannot be supported by the existing datasets. Second, PolyChartQA does not have explicit chart references in the questions (e.g., “first chart”). In contrast, existing datasets rely on such identifiers. This design choice forces MLMs to localize and select the relevant sub-chart(s) before reasoning, making the task closer to how questions naturally arise and substantially more challenging. Third, PolyChartQA preserves composite multi-chart figures as single images with embedded sub-charts, whereas MultiChartQA primarily represents multi-chart examples as sets of separate chart images. Finally, PolyChartQA explicitly categorizes questions into difficulty levels. PolyChartQA contains 941 more open-ended questions than MultiChartQA, expanding both task diversity and evaluation coverage.% Together, these differences make PolyChartQA well suited for realistic multi-chart reasoning.}

% Our dataset is significantly different from MultiChartQA \cite{zhu2024multichartqa} and CharXiv \cite{CharXiv} datasets. Our chart images naturally have multiple sub-charts in them and have annotations of multi-chart characteristics. In addition, the questions in our PolyChartQA dataset are formulated without chart identifiers (e.g., “the first chart”) or additional instructions, thereby more closely reflecting how questions naturally arise. This design requires models to autonomously infer the relevant sub-charts before answering, making the task more realistic and challenging. Our dataset contains 941 more open-ended questions than MultiChartQA, substantially expanding the evaluation space.

\iffalse
Pak: Remove due to redundant with the below.
We then benchmarked and evaluated nine open- and closed-source general-purpose and specialized-chart understanding MLMs on our PolyChartQA dataset and a modified MultichartQA dataset.
\fi
In summary, we make five primary contributions: (1) Formulation and investigation of the aforementioned research questions on MLM performance in multi-chart QA, focusing on chart structure, question difficulty, and human-authored vs. MLM-generated QA; (2) PolyChartQA, a benchmark dataset specifically designed to evaluate question answering over multi-chart images; (3) A comprehensive evaluation of nine state-of-the-art MLMs on the multi-chart QA task on two datasets; (4) A stepwise visual decomposition and self-verification pipeline to enhance MLM performance and interpretability on multi-chart QA; and (5) Key findings that MLMs perform significantly worse on multi-chart images compared to single-chart ones (up to 36.98\% L-accuracy drop), and that human-authored questions are substantially more challenging than MLM-generated ones, with performance gaps as large as 27.4\% L-accuracy. Our results highlight the challenges that MLMs face when reasoning over multi-chart images. Our curated dataset complements existing datasets to evaluate and improve MLM performance on multi-charts.
%reveal a significant performance gap, 
\section{Literature Review}
\subsection{Chart Question-Answering Datasets}
\iffalse
Understanding real-world charts is difficult as they are varied in terms of data and visualization. Answering questions on real-world charts requires a great understanding of language, domain knowledge, and reasoning. 

Earlier datasets primarily rely on synthetic data to generate chart images \citep{8578690,kahou2017figureqa}. 
To address the lack of real-world data complexity,
\citet{9093523} introduced PlotQA and \citet{9093269} created LEAF-QA. Both datasets are based on real-world data but corresponding chart images were synthetically generated, which do not reflect the diverse styles of many real-world charts \cite{masry-etal-2022-chartqa}. 
ChartQA \cite{masry-etal-2022-chartqa} was curated to have real-world chart images from various sources. 
\fi

\noindent \textbf{Single-chart QA datasets:} Most chart QA datasets contain single-chart images as summarized in Table~\ref{dataset_comparison}.
\iffalse
{\em Synthetic Data \& Chart Image Datasets} include FigureQA~\cite{kahou2017figureqa}, DVQA~\cite{8578690}, SynChart~\cite{liu2024synchartsynthesizingchartslanguage}, and SBS Figures~\cite{shinoda2024sbsfigurespretrainingfigure} and Omni-Chart-600K~\cite{wang-etal-2025-omni} utilizing LLMs to generate chart images.

{\em Real-world Data \& Corresponding Synthetic Chart Image Datasets}  include PlotQA \cite{9093523} and LEAF-QA  \cite{9093269}.

{\em Real-world Chart Datasets} include ChartQA \cite{masry-etal-2022-chartqa} with images curated from various sources, EvoChart-QA~\cite{Huang_Lai_Zhang_Wu_Ma_Zhang_Liu_2025} of real-world charts from 140 websites.
 \fi
 The questions of these datasets can be broadly categorized into (1) template-based \citep{8578690,9093269,kahou2017figureqa,9093523}, (2) human-authored \cite{masry-etal-2022-chartqa,Huang_Lai_Zhang_Wu_Ma_Zhang_Liu_2025}, and (3) MLM-generated questions \cite{shinoda2024sbsfigurespretrainingfigure,10654963}. %ChartQA \cite{masry-etal-2022-chartqa} used the T5 language model to automatically generate questions alongside a set of human-authored questions.  
 \iffalse
 Similarly, \citet{10654963} generates question-answer pairs for chart understanding using MLMs.
 
 However, existing chart understanding tasks focus heavily on single chart scenarios.
 \fi
 
 \noindent \textbf{Multi-chart QA datasets:} MultiChartQA \cite{zhu2024multichartqa} has 1,753 open-ended questions and 247 multiple-choice questions over 500 multi-chart sets ($\approx$2,168 sub-charts). %However, its questions include direct chart reference, and 
 However, approximately 75.2\% of the multi-chart sets consist solely of two or three single charts. MMC-Benchmark \cite{liu-etal-2024-mmc} contains 52 multi-chart images, which is inadequate to thoroughly assess model performance
on multi-chart tasks \cite{zhu2024multichartqa}. MMC-benchmark also lacks an in-depth exploration of the question-level
and chart-level complexity in multi-chart tasks. CharXiv \cite{CharXiv} contains 1,427 multi-chart
images.% and shows that the performance decreases with an increasing number of sub-charts.% However,
%CharXiv does not explore the influence of multi-chart characteristics or question difficulty.% {\color{red} TODO}
\subsection{Existing Models for Chart Understanding} %{\color{red} TODO}
Early chart understanding primarily relied on Optical Character Recognition and deep neural networks \cite{8578690,9093269}, followed by transformer-based models \cite{singh-shekhar-2020-stl}. Works on MLMs for chart-to-table conversion and evaluation include \cite{9093523,masry-etal-2023-unichart,liu-etal-2023-deplot,kim2025simplotenhancingchartquestion,masry-etal-2022-chartqa,zhou-etal-2023-enhanced,Liu_Li_Rao_Gao_Guan_Li_Ma_2025,meng-etal-2024-chartassistant}. %MLMs have further advanced the state of visual language understanding. 
Recent works focus on developing MLMs tailored for chart understanding \cite{zhang-etal-2024-tinychart,masry-etal-2024-chartinstruct,liu-etal-2024-mmc,masry-etal-2025-chartgemma} and evaluation of MLMs on chart reasoning tasks \cite{islam-etal-2024-large,zhu2024multichartqa,mukhopadhyay-etal-2024-unraveling,Huang_Lai_Zhang_Wu_Ma_Zhang_Liu_2025,CharXiv,ChartMuseum,masry-etal-2025-chartqapro,wu-etal-2024-chartinsights,hutchinson2025chartquestionansweringrealworld}. 

%With the introduction of language models, tasks like question answering and summarization have seen substantial improvements due to enhanced contextual comprehension and generation capabilities. \citet{9093523} introduced a method that converts charts into structured tables to facilitate question answering, particularly for out-of-vocabulary questions. More recent models, including UniChart, DePlot, and SIMPLOT, have also adopted chart-to-table conversion as a core strategy for chart understanding tasks \cite{masry-etal-2023-unichart,liu-etal-2023-deplot,kim2025simplotenhancingchartquestion}. \citet{masry-etal-2022-chartqa} uses language models combined with data tables for chart question answering tasks. 
%\citet{masry-etal-2024-chartinstruct,liu-etal-2024-mmc,masry-etal-2025-chartgemma} use instructions generated by MLMs to improve MLM's performance on chart understanding. 
\section{Proposed PolyChartQA}
We present PolyChartQA, a benchmark dataset for multi-chart question answering. PolyChartQA comprises 534 multi-chart images and 2,694 QA pairs (Human-authored: 519, MLM-generated: 2,175). % annotated across diverse question types and difficulty levels. 
MLM-generated QA pairs were generated using GPT-4.1 and manually verified. Table \ref{dataset_comparison} compares our dataset with the existing datasets.
\begin{table*}[t]
\centering
\resizebox{\textwidth}{!}{%
\begin{tabular}{llcccccc}
\toprule
\textbf{Chart Setting}&\textbf{Dataset} & \textbf{Image Type} & \textbf{\# Images} & \textbf{\# QA}  & \textbf{Question Generation} &\textbf{Sub-chart Ref. in Q} &\textbf{Chart Annotation}\\
\midrule
\multirow{7}{*}{Single-chart} & LEAF-QA~\cite{9093269}       & \makecell{Synthetic from\\real data} & 250K & 2M   & Template-based& -   &Chart type, Bounding box\\
&PlotQA~\cite{9093523}         & \makecell{Synthetic from\\real data} & 224.3K & 28.9M  & Template-based  & -&Chart type Bounding box\\
&ChartQA~\cite{masry-etal-2022-chartqa}       & Real-world             & 21.9K & 32.7K  & \makecell{Human-authored +\\Machine-generated} &-& None\\
&SBS Figures~\cite{shinoda2024sbsfigurespretrainingfigure}    & Synthetic & 1M & 4.2M  & LLM-generated (Verified 100 QAs) &-&Chart type \\
&SynChart~\cite{liu2024synchartsynthesizingchartslanguage} & Synthetic & 4M & 59.7M & LLM-generated  &-&Chart type\\
&EvoChart-QA~\cite{Huang_Lai_Zhang_Wu_Ma_Zhang_Liu_2025} & Real-world & 625 & 1,250 & Human-authored & -&Chart type\\
&CharXiv~\cite{CharXiv}    & Real-world & 896 & 3,584$^*$ & \makecell{Human-authored +\\Verified GPT-inspired \& generated}&- & \# of sub-chart\\
\hline
Multi-chart (Simulated)&MultiChartQA~\cite{zhu2024multichartqa}  & Real-world& \makecell{500 multi-chart sets\\($\approx$ 2,168 sub-charts)} & 2K & Human-authored   &Yes&None\\\hline
\multirow{2}{*}{\textbf{Multi-chart}} &CharXiv~\cite{CharXiv}    & Real-world & 1,427$^1$ & 5,708$^*$ & \makecell{Human-authored +\\Verified GPT-inspired \& generated}&Yes & \# of sub-chart\\
&\textbf{PolyChartQA (Ours)} & \makecell{\textbf{Real-world }\\ \textbf{Computer Science paper}} & \makecell{\textbf{534 divided into}\\ \textbf{2,297 sub-charts}} & \textbf{2.69K}  & \makecell{\textbf{Human-authored} +\\\textbf{Verified MLM-generated}} &\textbf{No} &\makecell{\textbf{\# of sub-chart, sub-chart type,} \\\textbf{chart homogeneity}}\\
\bottomrule
\end{tabular}}
\caption{Comparison of PolyChartQA with existing chart question answering datasets; $^1$of which 161 images and 161 reasoning QA are from computer science; $^*$ denotes \#reasoning and \#descriptive questions. 
}
\label{dataset_comparison}
\end{table*}
% \multirow{9}{*}{Single-chart} &FigureQA~\cite{kahou2017figureqa}  & Synthetic  & 140K & 1.8M    & Template-based  &-  & \\
% &DVQA~\cite{8578690}          & Synthetic & 300K & 3.49M   & Template-based   &  -&\\
\subsection{Charts Collection}
We curated a collection of multi-chart images from 168 research papers published in 2024 across 4 top-tier computer science conferences: EMNLP, ICSE, ICML, and SIGCOMM. %A total of 168 papers were collected. 
We then used PDFFigures 2.0 \cite{10.1145/2910896.2910904} to extract all the figures from these articles. Following figure extraction, a manual filtering process was conducted to identify and retain only multi-chart images, resulting in a dataset of 534 multi-chart images. Each multi-chart image was manually annotated with the number of sub-charts, chart type homogeneity, and specific sub-chart types by the first author. Of these images, 85.58\% are homogeneous multi-chart images where all sub-charts are of the same type. The remaining images have multiple types of sub-charts. PolyChartQA includes 11 chart types.  Although this reflects an imbalance in chart homogeneity, the imbalance originates from real-world sources.% A similar trend is observed in other real-world datasets such as CharXiv \cite{CharXiv} containing 23.7\% images with five or more sub-charts. 
%Figure \ref{sub_chart_count} shows the distribution of sub-charts in the images.} 

To validate the annotations, given each multi-chart image, we prompted GPT-4o to annotate with the number of charts, the type of sub-charts, and the chart type homogeneity. We validated annotations for all 534 multi-chart images. The agreement between the first author and the GPT-4o annotations was 91.95\%. The disagreement was due to sub-chart count (4.49\%), sub-chart type (3.93\%), and chart homogeneity (0.75\%), with some instances involving overlaps across these categories. The first author ensured the accuracy and consistency of the analysis. As GPT-4.1 became available later, we used it for question-and-answer generation. Appendix \ref{Appendix_distribution_chart_image} summarizes multi-chart image distributions by conference, sub-chart count, and chart type.%Appendix  \ref{Appendix_distribution_chart_image} describes the distribution of multi-chart images across the different conferences, number of sub-charts in an image, and the type of charts.% in homogeneous charts.

% \begin{figure}[ht]
% \centering
% \includegraphics[width=\linewidth]{chart_component_count.png}
% \caption{Distribution of numbers of sub-charts in multi-chart figures.}
% \label{sub_chart_count}
% \end{figure}
\subsection{Question Generation}

We define three difficulty levels and three question types for our experiments. Our questions for each multi-chart image are open-ended and can be answered from the image. All question criteria were established before the annotation process began and were applied uniformly during both QA generation and subsequent manual verification. This strict adherence to predefined guidelines ensured consistency in question formulation, reducing individual annotator stylistic differences and intent-level bias.
\\~\\
\noindent{\textbf{Question Difficulty Levels}
     \begin{itemize}
         \item \textbf{Easy:} Easy questions focus on a single sub-chart with straightforward retrieval or trend analysis. %Easy questions require inspection of a single sub-chart only. %The information needed to answer the question, such as visual elements or explicit values, is directly visible and unambiguous within the individual sub-chart.
         %These questions typically involve straightforward data retrieval or simple trend identification within the individual sub-chart. %They do not require integrating information from any other sub-chart in the multi-chart figure. While 
         Our Easy questions are similar to the Direct questions from MultiChartQA.% \cite{zhu2024multichartqa}.%, the latter includes explicit chart references that guide the model to the relevant sub-chart. Our Easy questions do not provide such references, requiring the model to identify the relevant chart.% autonomously.
         \item \textbf{Medium:} Medium questions require comparing, aggregating, or synthesizing information spread across two or more sub-charts. %Although the reasoning involved is more complex than that for the easy questions,
         The reasoning is relatively direct and does not require recursive visual inference.
         %The reasoning remains relatively direct, and the logical steps can usually be followed without recursive inference %or multi-stage transformations of 
         %visual features.
         \item \textbf{Hard:} Hard questions demand multi-step reasoning across multiple sub-charts. These include tasks that require chaining information across sub-charts, for example, extracting a label or legend from one chart and applying it to interpret symbols or data points in another. % Hard questions may also involve complex data retrieval (e.g., retrieving a numeric value from one sub-chart to conditionally extract or interpret another from a separate sub-chart) or reasoning. 
         Our Hard questions are inspired by the Sequential Reasoning questions in MultiChartQA \cite{zhu2024multichartqa}. However, MultiChartQA’s sequential reasoning questions only focus on content-based tasks, such as extracting specific numbers or phrases from given clues, and do not include structural questions. In contrast, our Hard questions span structural, data retrieval, and reasoning types, all without providing explicit chart references.%, thereby increasing the complexity. 
         %, but differ in two important ways. First, while MultiChartQA's Sequential Reasoning questions involve multi-hop reasoning and stepwise interpretation across charts, they include explicit chart references that guide the model through the reasoning path. Second, these questions are only focused on content-based tasks such as extracting specific numbers or phrases based on provided clues, and do not encompass structural questions. Our Hard questions span structural, data retrieval, and reasoning types, all without providing explicit chart references, thereby increasing the complexity. % and chart selection burden on the model.
     \end{itemize}
\noindent \textbf{Question Types:} We use the types of questions following existing works \cite{9093523}.% {\color{red} Can all our questions be answered using the figure and the caption without external knowledge? See CharXiv for their description. Specify the question format. Open-ended answers or multiple choices or what. Should we include a measure of the length of the question for each type? See the CharXiv statistic table.}
     \begin{itemize}
         \item \textbf{Structural:} Focuses on visual or structural elements, e.g., colors and labels.
        \item  \textbf{Data Retrieval:} Focuses on retrieving explicit numeric or categorical values, e.g., reading data points, and extracting counts.
        \item \textbf{Reasoning:} Involves inference, comparison, aggregation, trend analysis, or contradiction.
        \end{itemize}

% We generated two sets of open-ended question-answer pairs: (1) Human-authored, (2) MLM-generated using GPT-4.1. We explain both of the generation processes below.% Recent work on chart understanding shows an increasing trend of using MLM-generated questions for scalability \cite{pramanick2025spiqadatasetmultimodalquestion,10654963,shinoda2024sbsfigurespretrainingfigure}. 
\paragraph{Human-authored PolyChartQA:} The human-authored QA pairs were manually crafted by the first author, a computer science graduate student. Initially, 520 QA pairs were created from 238 multi-chart images. These QA pairs were then evenly distributed between the second author (a computer science graduate student) and the third author (a professor of computer science) for validation. Each question was validated by a single annotator (either second or third author). The validation focused on two criteria: (i) clarity of the question and (ii) correctness of the QA pair. Based on the feedback, one image and its associated {\color{cyan}}one question–answer pair were excluded as both the question and the answer were incorrect. Additionally, 98 QA pairs were revised for clarity and correctness. %The final dataset comprises 519 human-authored QA pairs derived from 237 multi-chart images.
\paragraph{MLM-generated PolyChartQA:} We prompted GPT-4.1 to generate QA pairs from all the multi-chart images. Our prompt includes the criteria for the desired question difficulty level, question type, multi-chart image, and instructions to generate QA pairs per the given criteria. Initially, we generated a total of 4,308 QA pairs from 534 images. \textbf{Manual Verification:} All the generated QA pairs were checked by the first author to ensure correctness according to our criteria given in \ref{criteria_to_filter_ques} Table \ref{qa_removal_criteria}. About 49.5\% of the QA pairs (2,133 QA pairs) and 3 multi-chart images were removed. Next, the first author verified the correctness of the question difficulty, question type, and relevant chart annotations for the remaining questions according to our definitions of difficulty and question types. After manual checking, annotation of 18.99\%, 11.26\%, and 7.59\% QA pairs was revised for the question difficulty, question type, and relevant chart. For the relevant charts, most of the edits were changing specific chart names to `Any chart'. Additionally, we found 16 QA pairs for which GPT-4.1 assigned more than one question type (e.g., Structural/Reasoning instead of Reasoning). We asked an independent human annotator (an undergraduate CS research assistant) to correct the labeling mistakes for these 16 QA pairs. The annotator was briefed with detailed definitions of question types, difficulty levels, and task instructions. 

%We provided the annotator with detailed definitions of question types and difficulty levels and thoroughly explained the task during a dedicated briefing session. %0.74\% or
\begin{table}[t]
\small
\centering
\resizebox{0.88\columnwidth}{!}{
\begin{tabular}{lr}
\hline
\textbf{Statistic} & \textbf{Number} \\
\hline
\multicolumn{2}{l}{\textbf{Human-authored PolyChartQA}} \\
Images & 237 \\
Sub-charts & 894 \\
Homogeneous charts & 209 \\
Non-homogeneous charts & 28 \\
%Unique papers & 66 \\
Total questions & 519 \\
\textit{Question Difficulty:} & \\
\quad Easy & 293 \\
\quad Medium & 168 \\
\quad Hard & 58 \\
\textit{Question Type:} & \\
\quad Structural & 89 \\
\quad Data Retrieval & 141 \\
\quad Reasoning & 289 \\
%Average question length (words) & 14.82 \\
%Median question length (words) & 13.0 \\
%Average answer length (words) & 6.94 \\
% Median answer length (words) & 2.0 \\
% Min questions per image & 1 \\
% Max questions per image & 8 \\
Median questions per image & 1.0 \\
Distinct-2 (Question)& 0.4516 \\
Avg. Token Length (Question)& 15.48\\
Min - Max Token Length (Question)& 5 - 51\\\hline
%Average questions per image & 2.05 \\
\multicolumn{2}{l}{\textbf{Verified MLM-generated PolyChartQA}} \\
Images & 531 \\
Sub-charts & 2,290 \\
Homogeneous charts & 454 \\
Non-homogeneous charts & 77 \\
%Unique papers & 258 \\
Total questions & 2,175 \\
\textit{Question Difficulty:} & \\
\quad Easy & 1,208 \\
\quad Medium & 867 \\
\quad Hard & 100 \\
\textit{Question Type:} & \\
\quad Structural & 763 \\
\quad Data Retrieval & 302 \\
\quad Reasoning & 1,109 \\
%Average question length (words) & 17.79 \\
%Median question length (words) & 17.0 \\
%Average answer length (words) & 10.52 \\
%Median answer length (words) & 9.0 \\
%Min questions per image & 1 \\
%Max questions per image & 9 \\
Median questions per image & 4.0 \\
Distinct-2 (Question)& 0.3356 \\
Avg. Token Length (Question)& 18.49\\
Min - Max Token Length (Question)& 8 - 56\\\hline
%Average questions per image & 4.15 \\
\multicolumn{2}{l}{\textbf{Total (Combined)}} \\
Images & 534 \\
Sub-charts & 2,297 \\
Median sub-charts per image & 3\\
Homogeneous charts & 457 \\
Non-homogeneous charts & 77 \\
Unique papers & 168 \\
Unique chart types & 11\\
Total questions & 2,694 \\
\hline
\end{tabular}}
\caption{PolyChartQA dataset statistics.}
\label{polychartqa_stat}
\end{table}
Although scaling the dataset size through MLM-generated questions is an option, we prioritized quality by manually verifying all QA pairs generated by the MLM. This decision inevitably limited scalability but ensured correctness, which is particularly critical given the substantially greater complexity of multi-chart images and harder question types compared to single-chart. Prior large-scale efforts illustrate this inherent trade-off between manual validation and scalability. 
QA pairs from less than 1\% of the number of papers in the SPIQA dataset were manually verified \cite{pramanick2025spiqadatasetmultimodalquestion}, of which about 11\% were discarded. Similarly, about 52.21\% of all QA pairs were discarded after manual verification~\cite{hutchinson2025chartquestionansweringrealworld}.
\iffalse In SPIQA \cite{pramanick2025spiqadatasetmultimodalquestion}, QA pairs were generated from figures and tables across 25,859 papers, but were manually verified only in 200 papers. About 11\% of the questions were then discarded. Such a drop rate would be even higher for multi-chart figures. Similarly, \citet{hutchinson2025chartquestionansweringrealworld} generated 429 QAs from interactive charts using MLM and discarded 224 (a 52.21\% drop rate) after manually verifying all QA pairs. 
\fi
We chose to retain fewer higher-quality questions by validating every pair. Table \ref{polychartqa_stat} presents the statistics of our dataset. The Distinct-2 \cite{li-etal-2016-diversity} (unique bigram ratio) of the human-authored questions (0.4516) in Table \ref{polychartqa_stat} is higher than those of the MLM-generated ones and  MultiChartQA (0.2257), indicating greater lexical diversity. We compute Distinct-2 using only the core question text, excluding answer instructions. Beyond Figure \ref{fig:comb},
additional human- and MLM-generated examples are in \ref{appendix_humanQA_examples} - \ref{appendix_MLMQA_examples}. \ref{appendix_prompt_ex} presents the prompts for validating chart annotation, and generating QA pairs.

%The generated QA pairs appear promising. However, after manual checking of the results, we found that the MLM assigned more than one question type (e.g Structural/Reasoning instead of Reasoning) for 471 QA pairs. We asked an independent human annotator to correct the labeling mistakes for the 471 QA pairs. For the rest of the generated QA pairs {\color{red} Q}, we kept the label as is.
% \begin{figure*}[]
% \centering
% %\resizebox{\linewidth}{!}{!
% \begin{subfigure}[c]{0.49\textwidth}
%     \centering
% \includegraphics[width=\linewidth]{H-Q-E-2.png}
%     \caption{The multi-chart is collected from \citet{chen-etal-2024-new}.}
% \end{subfigure}
% %\hspace{0.01\textwidth}
% \begin{subfigure}[c]{0.49\textwidth}
%     \centering
%     \includegraphics[width=\linewidth]{H-Q-E-5.png}
%     \caption{The multi-chart is from \citet{furniturewala-etal-2024-thinking}.}
% \end{subfigure}
% % }
% \caption{Example of Human-authored QA pairs from PolyChartQA.} 
% \label{fig:comb}
% \end{figure*}
% \begin{figure*}[!t]
%     \centering
% \includegraphics[width=0.7\textwidth]{H-Q-E-2.png}
%     \caption{
%         }
%     \label{H_Q_E_2}
% \end{figure*}
% \begin{figure*}[!t]
%     \centering
% \includegraphics[width=0.7\textwidth]{Q_ex_M-Text.png}
%     \caption{
%       {\color{red}  Examples of human-authored Easy-Structural, Medium-Reasoning, and Hard-Reasoning QA pairs from our PolyChartQA. The multi-chart is from \citet{JMLR:v25:21-1343}.} %{\color{red} Using the same chart, add difficult reasoning, or medium data retrieval.}
%     }
%     \label{human_qa}
% \end{figure*}
\section{Proposed Visual Decomposition and Self-Verification Pipeline (VDSP)}
To improve MLM performance and interpretability in complex multi-chart QA, we propose a modular 3-stage prompt-based pipeline that decomposes the reasoning process while leveraging the visual understanding capabilities of MLMs.  Figure \ref{vdsp_fig} shows our pipeline with explicit visual decomposition, structured reasoning, and self-verification, aligned with the observed error types.% Figure \ref{vdsp_fig} provides an overview of our pipeline. The three stages are:% of the pipeline are described below:
\begin{figure}[]
    \centering
\includegraphics[width=\columnwidth]{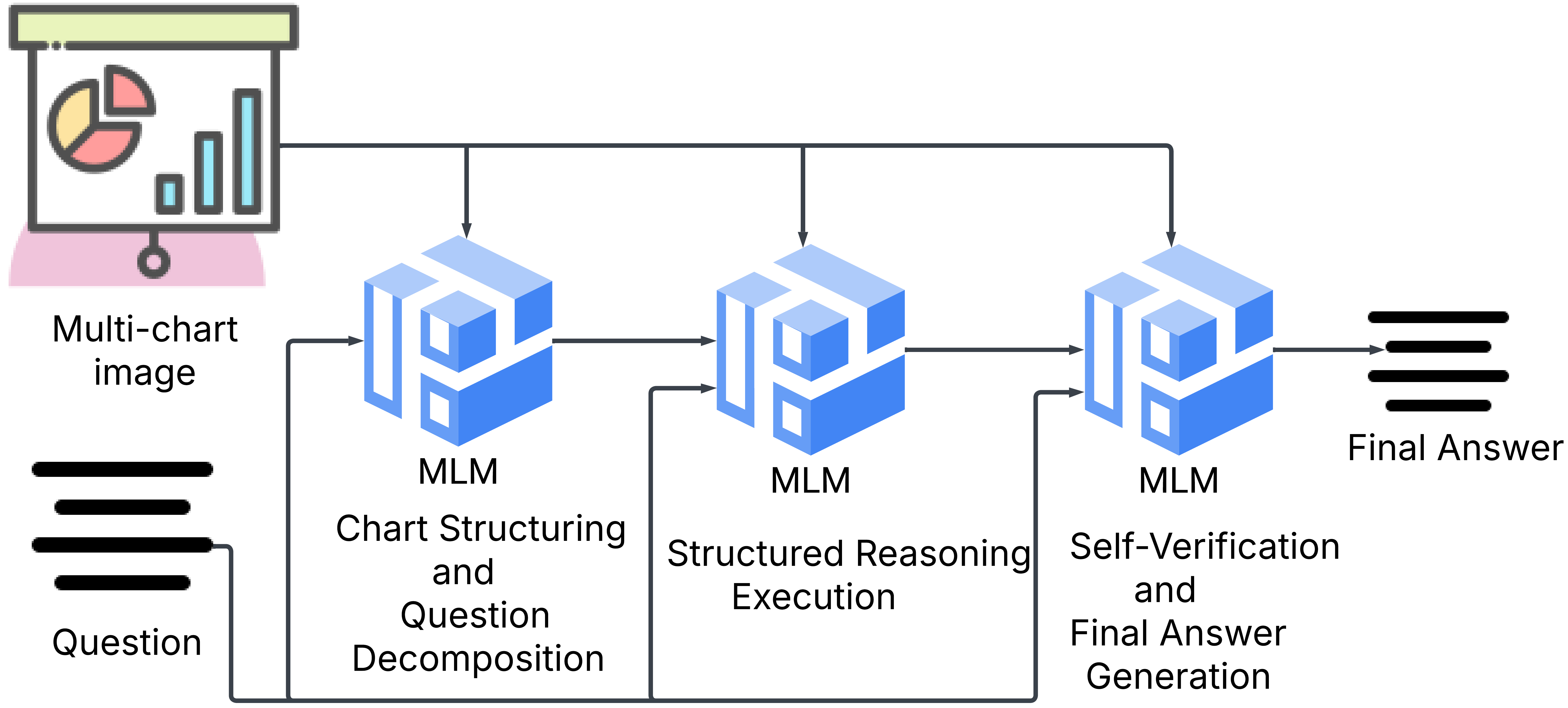}
    \caption{
        VDSP pipeline for multi-chart QA.
    }
    \label{vdsp_fig}
\end{figure}
\paragraph{Stage 1. Chart Structuring and Question Decomposition:} Given a multi-chart image and the question, the MLM is prompted to first parse the image into individual sub-charts, identifying their types (e.g., bar, line) and key characteristics (e.g., axis labels, legends). Next, the question is decomposed into a sequence of sub-tasks, grounded in chart structure. This mimics a form of visual Chain of Thought reasoning, but with explicit mapping between visual elements and reasoning steps.% This step establishes an interpretable semantic layout of the visual input and a task-aware plan that guides subsequent steps.
\paragraph{Stage 2. Structured Reasoning Execution:} Given the multi-chart image, question, and Stage 1 output, the MLM extracts relevant values from specific visual components (e.g., bar heights) and performs computations or comparisons as needed. Finally, the MLM combines intermediate outcomes to formulate a preliminary answer.% This step enforces explicit grounding of reasoning in visual evidence, thereby minimizing hallucination.% and encouraging traceable logic.
\paragraph{Stage 3. Self-Verification and Final Answer Generation:} Given the multi-chart image, the question, and the output from Stage 2, the model performs a self-reflective check on the reasoning in Stage 2 by verifying whether all chart references and visual details are accurate, followed by detecting common reasoning errors (e.g., incorrect axis readings, incorrect chart selection). It returns either a validated or a corrected final answer.% Figures \ref{stage1} - \ref{stage3} in Appendix \ref{appendix_prompt_ex} list the prompts for each stage.
\section{Experimental Setup}
\noindent \textbf{Nine compared models:} We evaluated three state-of-the-art closed-source MLMs (Claude-3.7-Sonnet, GPT-4.1, and Gemini-2.0-Flash), four open-source MLMs (Pixtral-12B, Llama-3.2-11B-Vision, Llava-1.5-7b, and Gemma-3-4b-it), and two chart-specialized MLMs, ChartGemma \cite{masry-etal-2025-chartgemma} and MatCha \cite{liu-etal-2023-matcha}. See the configurations in \ref{model_config}.

\noindent \textbf{Two datasets:} PolyChartQA was used for testing RQ2-4 while   MultiChartQA-RQ1 was used for RQ1. 
MultiChartQA-RQ1 consists of all 365 ``Direct'' questions from the MultiChartQA \citep{zhu2024multichartqa} dataset in which all the images in the multi-chart sets are single-chart images. Direct questions refer to those for which an MLM can derive the correct answer using information from a single chart. We replaced explicit sub-chart identifiers (e.g., “first chart”) in the question with the generic term “chart”. The same QA pairs were used for both the single-chart and multi-chart conditions. The two conditions differ only in the input image configuration. In the single-chart setting, the MLM was provided with only the relevant sub-chart needed to answer the question. In the multi-chart setting, we constructed a composite multi-chart image by merging all charts from the corresponding set into a single image. As the questions and answers were identical across both settings, any observed performance differences directly reflect the added difficulty of sub-chart localization and reasoning in a multi-chart visual context, thereby isolating the effect of multi-chart complexity.

\textbf{Prompting methods:} We used Zero-shot prompting for RQ1, Zero-shot and Chain-of-Thought (CoT) for RQ2–RQ4, and also the proposed VDSP prompting on the human-authored QA pairs of PolyChartQA for RQ4.

%To simulate our multi-chart setting, for each multi-chart set in MultiChartQA, we merged the individual images for that set into one multi-chart image and replaced the sub-chart identifier (first/ second/ third chart) with `chart' in the question. 
%{\color{red} For RQ3, we randomly selected 100 pairs of multi-chart images from MultiChartQA and one randomly selected question from each pair for the human-authored question. Then, we asked GPT-4.1 to generate one question-answer pair from each multi-chart pair. We call this dataset MultiChartQA-RQ2.}\\

\textbf{Evaluation Metrics:} We use H-Accuracy (Human-evaluation), L-Accuracy (LLM-based accuracy) introduced by the previous works \cite{pramanick2025spiqadatasetmultimodalquestion, liu-etal-2024-mmc,ChartMuseum,CharXiv}, and BERTScore \cite{zhang2020bertscoreevaluatingtextgeneration}. BERTScore measures semantic similarity between model predictions and references. H-accuracy was obtained through human evaluation of model-predicted answers for human-authored questions in a Zero-shot setting. As human evaluation is difficult to scale, for all other cases, we use L-accuracy by prompting a selected LLM as a judge to assess whether the ground-truth and model-generated answers are similar or not. 

% {\color{red} TODO: Use Cohen's Kappa agreement is better.}
\textbf{Judge selection:} We randomly sampled 100 QA pairs from PolyChartQA, where seven general-purpose MLM-predicted answers were evaluated independently by a human and three LLMs: Claude-3.7-Sonnet, GPT-4.1, and Gemini-2.0-flash. The results showed substantial to almost perfect Cohen’s Kappa agreement between the human and Claude-3.7-Sonnet (0.73–0.92), the human and GPT-4.1 (0.65–0.94), and moderate to substantial agreement between the human and Gemini-2.0-flash (0.46-0.84). Given its highest alignment with human evaluation, Claude-3.7-Sonnet was selected as the judge to compute L-Accuracy. Detailed agreement statistics are provided in \ref{agreement_llm_human}.

%The results showed a strong alignment, with 89–99\% agreement between human and GPT-4.1, 93–98\% between human and Claude-3.7-Sonnet, and 81–93\% between human and Gemini-2.0-flash evaluations. Given its highest alignment with human evaluation, Claude-3.7-Sonnet was selected as the judge to compute L-Accuracy for the remaining experiments. {\color{cyan}Cohen’s Kappa values  indicate substantial to almost perfect agreement between human and Claude-based evaluations.} Detailed agreement statistics are provided in Appendix~\ref{agreement_llm_human}. {\color{red} Pick one agreement to discuss. I think using Cohen's Kappa throughout the discussion is better.}

%For the final computation of L-Accuracy, Claude-3.7-Sonnet was asked to assess whether the ground truth and the model’s predicted answer were similar or not, following the prompt design of \citet{pramanick2025spiqadatasetmultimodalquestion}. 
Following previous studies \cite{CharXiv,pramanick2025spiqadatasetmultimodalquestion,liu-etal-2024-mmc,ChartMuseum}, we report all metrics based on a single evaluation run. While minor variability may occur across runs or model versions, Claude-3.7-Sonnet showed strong alignment with human judgments, ensuring reliable evaluation. The prompt for L-Accuracy is provided in \ref{appendix_prompt_ex}.% Figure \ref{acc_prompt}.%, computed using the bertscore Python package (v0.3.13) with the bert-base-uncased model and default parameters.

\section{Experimental Results}
Underlined scores indicate the best performance within each category. Performance drops are computed as the score difference across categories for each individual MLM.%Bold values denote the best overall performance, unless stated otherwise. 

% In all the results, the underlined scores indicate the best performance within their category. The results are based on single-run evaluations of MLMs using a zero-shot prompt on MultiChartQA-RQ1 (RQ1), zero-shot, and chain-of-thought (COT) prompts on PolyChartQA (RQ2–RQ4), and proposed VDSP prompting on the human-authored QA pairs of PolyChartQA for RQ4. Bold font indicates best overall performance.

\paragraph{\textbf{RQ1 Findings: Several MLMs perform  significantly worse on multi-chart QA compared to single-chart QA.}} %We explore RQ1 using a zero-shot prompting approach on three closed-source models on MultiChartQA-RQ1. We selected Gemini-2.0-flash as it is the latest model among the Gemini models under study. 
Figure~\ref{single_vs_multi} shows that L-accuracy drops from single- to multi-chart settings by 26.85\%–36.98\% for closed-source, 3.56\%–36.44\% for open-source, and 4.66\%–6.30\% for chart-specific models. MatCha and ChartGemma perform poorly, likely due to their pre-training on single-chart datasets. BERTScore decreases by up to 12.38\% when processing multi-chart images, indicating reduced semantic alignment. 

 % The L-Accuracy of closed-source models decreases substantially when moving from the single-chart to the multi-chart setting, with drops ranging from 26.85\% to 36.98\%. Similarly, open-source models exhibit reductions from 3.56\% to 36.44\%, while chart-specific models show declines of 4.66\%–6.3\%. The transition to the multi-chart setting results in a BERTScore reduction of up to 12.38\%, indicating a notable decline in semantic alignment. Table \ref{single_vs_multi} shows the detailed results.
\begin{figure}[]
\centering
\includegraphics[width=\columnwidth]{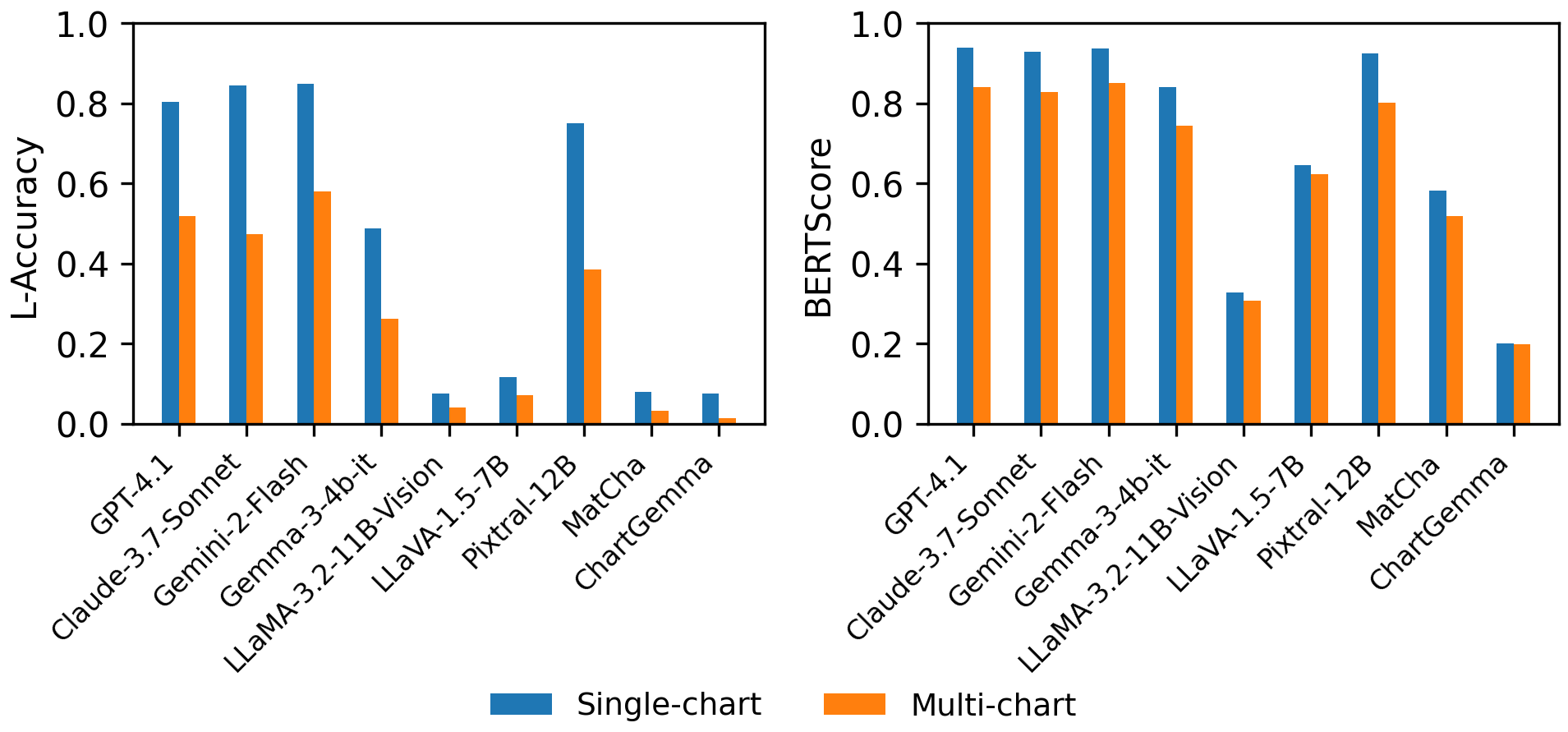}
\caption{Single-chart vs multi-chart QA performance.}
\label{single_vs_multi}
\end{figure}
\paragraph{\textbf{RQ2.1 Findings: {\em Performance consistently declines with increasing question difficulty.}}}
This trend is consistent across all MLMs and question sources, underscoring the difficulty MLMs face with multi-step reasoning and cross-chart aggregation.  Figure \ref{tab:lacc_zeroshot_difficulty} reports Zero-shot L-Accuracy, with other results in \ref{appendix_results_qdiff}. \textbf{Human-authored questions:} Both H- and L-accuracy consistently decline from Easy to Hard across all MLMs. For L-Accuracy under Zero-shot, the drop ranges are 17.24\%–29.34\% (closed-source MLMs), 15.02\%–27.69\% (open-source MLMs), and 10.58\%–23.89\% (chart-specific MLMs). For H-Accuracy, the corresponding drops are 23.09\%–27.94\% (closed-source MLMs), 16.04\%-31.10\% (open-source MLMs), and 10.92\%–26.28\% (chart-specific MLMs). %The performance drop ranges from 17.24\%–29.34\% for closed-source MLMs, 15.02\%–31.10\% for open-source MLMs, and 10.59\%–26.28\% for chart-specific MLMs under a zero-shot setting. 
CoT prompting slightly mitigates the decline, but substantial gaps remain. \textbf{MLM-generated questions:} A similar trend is observed, with declines of 4.62\%–30.35\% (Zero-shot), and 5.79\%-28.27\% (CoT) L-Accuracy. 
\begin{figure}[]
\centering
\includegraphics[width=\columnwidth]{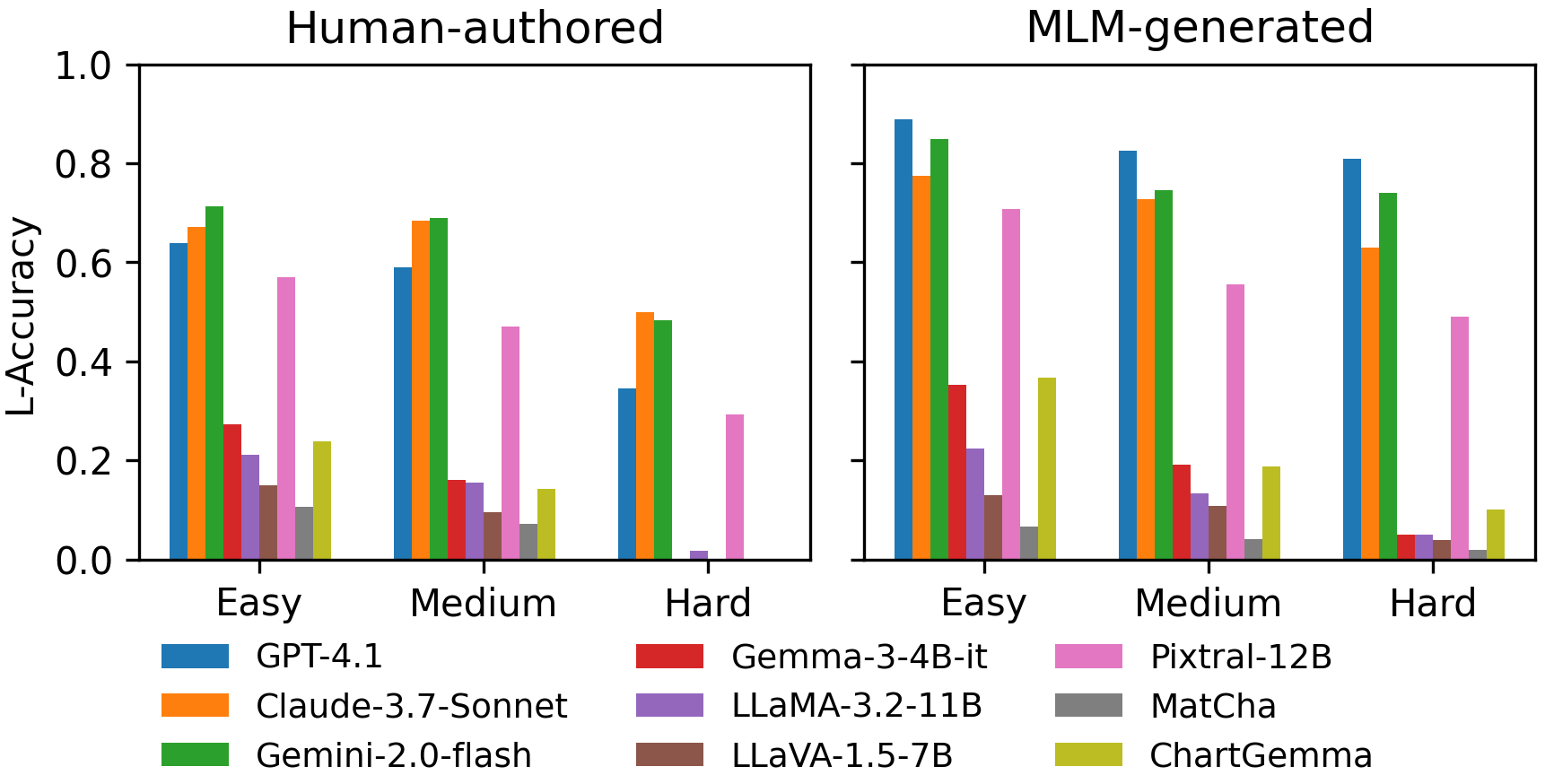}
\caption{L-Accuracy on PolyChartQA under Zero-shot prompting across difficulty levels.}
\label{tab:lacc_zeroshot_difficulty}
\end{figure}
\paragraph{\textbf{RQ2.2 Findings: {\em MLM performances vary by question types.}}}
The models consistently perform best on structural and worst on data retrieval questions. \textbf{Human-authored questions:} Both H- and L-accuracy consistently decline from structural, reasoning, to data retrieval types across MLMs except for MatCha for both Zero-shot and CoT prompting. Figure \ref{l_acc_qtype_zero} and \ref{appendix_results_qtype} show detailed results. %Structural questions use clear visual cues, reasoning leverages relative comparisons, while data retrieval demands exact value extraction—often hindered by ambiguous or overlapping chart–axis alignments. 
In summary, for L-Accuracy under Zero-shot, the drop ranges are 20.5\%-35.09\% (closed-source MLMs), 17.16\%-28.35\% (open-source MLMs), and 9.88\%-34.31\% (chart-specific MLMs).
For H-Accuracy, the corresponding drops are 20.5\%-37.76\% (closed-source MLMs), 10.42\%-28.89\% (open-source MLMs), and 8.46\%-35.13\% (chart-specific MLMs). %the performance drop ranges from 20.45\%–37.76\% for closed-source MLMs, 10.42\%–28.89\%  for open-source MLMs, and 8.46\%–35.13\% for chart-specific MLMs under the zero-shot setting. %CoT also exhibits a large performance gap. %CoT prompting slightly mitigates the decline, but substantial gaps remain. 
\textbf{MLM-generated questions:} A similar trend is observed for all closed-source models and Pixtral, %with somewhat smaller declines, 
with declines of 10.74\%–22.89\% (Zero-shot), 11.73\%-25.59\% (CoT). The rest of the MLMs perform the best on structural questions, but sometimes better on data retrieval than on reasoning questions. 
\begin{figure}[]
\centering
\includegraphics[width=\columnwidth]
{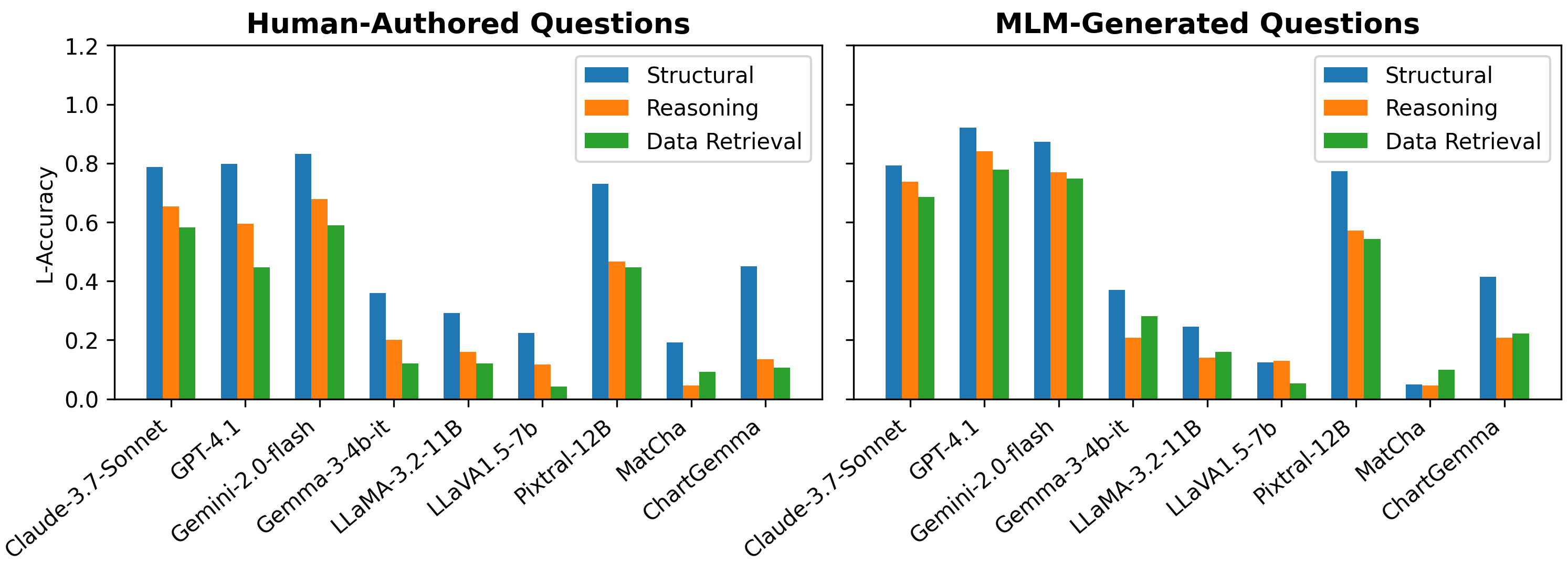}
\caption{L-Accuracy on PolyChartQA under Zero-shot setting across question types.}
\label{l_acc_qtype_zero}
\end{figure}

\paragraph{\textbf{RQ2.3 Findings: {\em Most MLMs perform better on non-homogeneous charts.}}}
 The detailed results are in \ref{appendix_results_homo}. %{\color{yellow} Recheck the results, GPT4.1, Gemma, Llava do not follow the trend indicated.}
In summary, under Zero-shot on human-authored questions, both H- and L-Accuracy decline from non-homogeneous to homogeneous charts, with drops of up to 9.31\% (L-Accuracy) and 8.40\% (H-Accuracy) for closed-source MLMs, 17.41\% (L-Accuracy) and 14.58\% (H-Accuracy) for open-source MLMs, and 1.26\% (H-Accuracy) for chart-specific MLMs %both H- and L-accuracies decline from non-homogeneous to homogeneous 
%charts up to 9.31\%(L-Accuracy), 8.4\%(H-Accuracy) for closed-source, 17.41\% for 
%open-source (L-Accuracy), 14.58\%(H-Accuracy), and 1.26\%(H-Accuracy) for chart-specific MLMs 
%under the zero-shot setting 
except GPT-4.1, Gemma, and ChartGemma. The downward trend holds for MLM-generated questions, up to 9.02\% (L-Accuracy). CoT mitigates the gap mostly.
\paragraph{\textbf{RQ2.4 Findings: {\em MLM performance generally declines as the number of sub-charts increases.}}}
This trend is more pronounced for human-authored QAs than  MLM-generated QAs. Detailed results are in \ref{appendix_result_subchart}.

\paragraph{\textbf{RQ3 Findings: Human-authored questions are significantly harder than MLM-generated ones.}} Table \ref{tab:chi_square_combined} reveals a statistically significant accuracy gap between human-authored and MLM-generated questions for most MLMs using the Pearson’s chi-squared statistical test. With Zero-shot prompting, L-accuracy drops from MLM-generated QAs to human-authored QAs by 9.24–27.02\% (closed-source), up to 13.1\% (open-source), and up to 10.16\% (chart-based). With CoT, L-accuracy decreases is similar (up to 27.4\%).
\iffalse 8.24\%–27.4\% (closed-source), up to 10.17\% (open-source), and up to 4.24\% (chart-based), respectively. \fi 
BERTScore is presented in \ref{appendix_results_poly_3}.
\begin{table}[]
\centering
\small
\resizebox{\columnwidth}{!}{%
\begin{tabular}{lcccc}
\toprule
\multirow{2}{*}{\textbf{Model}} &
\multicolumn{2}{c}{\textbf{Zero-shot}} &
\multicolumn{2}{c}{\textbf{CoT}} \\
\cmidrule(lr){2-3} \cmidrule(lr){4-5}
 & \textbf{Human} & \textbf{MLM} 
 & \textbf{Human} & \textbf{MLM}  \\
\midrule
Claude-3.7-Sonnet & \textbf{0.6570} & \textbf{0.7494} & \textbf{0.6647} & \textbf{0.7471}  \\
GPT-4.1 & \textbf{0.5896} & \textbf{\underline{0.8598}} & \textbf{0.5954} & \underline{\textbf{0.8694}}  \\
Gemini-2.0-flash & \textbf{\underline{0.6802}} & \textbf{0.8028}  & \textbf{\underline{0.6936}} & \textbf{0.8106} \\
Gemma-3-4b-it & \textbf{0.2062} & \textbf{0.2749}  & 0.2158 & 0.2529 \\
LLaMA-3.2-11B-Vision & 0.1715 & 0.1794 & \textbf{0.0848} & \textbf{0.1559}  \\
LLaVA1.5-7b & 0.1156 & 0.1168  & 0.1195 & 0.1159 \\
Pixtral-12B & \textbf{0.5067} & \textbf{0.6377} & \textbf{0.5337} & \textbf{0.6354}  \\
MatCha & \textbf{0.0829} & \textbf{0.0543}  & 0.0617 & 0.0441  \\
ChartGemma & \textbf{0.1811} & \textbf{0.2828}  & 0.1911 & 0.2336  \\
\bottomrule
\end{tabular}}
\caption{L-Accuracy on Human vs. MLM-generated questions. Bold pairs indicate statistically significant differences at a significance level of 0.05.}
\label{tab:chi_square_combined}
\end{table}
% \begin{table}[]
% \centering
% \small
% \resizebox{\columnwidth}{!}{%
% \begin{tabular}{lcccccc}
% \toprule
% \multirow{2}{*}{\textbf{Model}} &
% \multicolumn{3}{c}{\textbf{Zero-shot}} &
% \multicolumn{3}{c}{\textbf{CoT}} \\
% \cmidrule(lr){2-4} \cmidrule(lr){5-7}
%  & \textbf{Human} & \textbf{MLM} & \textbf{p-value}
%  & \textbf{Human} & \textbf{MLM} & \textbf{p-value} \\
% \midrule
% Claude-3.7-Sonnet & 0.6570 & 0.7494 & \textbf{0.0001} & 0.6647 & 0.7471 & \textbf{0.0005} \\
% GPT-4.1 & 0.5896 & \underline{0.8598} & \textbf{0.0000} & 0.5954 & \underline{\textbf{0.8694}} & \textbf{0.0000} \\
% Gemini-2.0-flash & \underline{0.6802} & 0.8028 & \textbf{0.0000} & \underline{0.6936} & 0.8106 & \textbf{0.0000} \\
% Gemma-3-4b-it & 0.2062 & 0.2749 & \textbf{0.0054} & 0.2158 & 0.2529 & 0.1899 \\
% LLaMA-3.2-11B-Vision & 0.1715 & 0.1794 & 0.8297 & 0.0848 & 0.1559 & \textbf{0.0002} \\
% LLaVA1.5-7b & 0.1156 & 0.1168 & 0.9303 & 0.1195 & 0.1159 & 0.9036 \\
% Pixtral-12B & 0.5067 & 0.6377 & \textbf{0.0000} & 0.5337 & 0.6354 & \textbf{0.0001} \\
% MATCHA & 0.0829 & 0.0543 & \textbf{0.0440} & 0.0617 & 0.0441 & 0.2293 \\
% ChartGemma & 0.1811 & 0.2828 & \textbf{0.0000} & 0.1911 & 0.2336 & 0.1013 \\
% \bottomrule
% \end{tabular}}
% \caption{Human vs. MLM QA L-accuracy.}
%  \vspace{-10pt}
% \label{tab:chi_square_combined}
% \end{table}
\paragraph{\textbf{RQ4 Findings: Incorporating stepwise visual decomposition and self-verification enhances the
L-accuracy and interpretability.}} 
Table \ref{L-Acc_rq4} shows L-accuracy of the closed-source models and Pixtral-12B, the best-performing open-source model on human-authored questions from PolyChartQA. VDSP improves L-accuracy by up to 5.39\% over Zero-shot and 4.62\% over CoT. While GPT-4.1 exhibits similar accuracy under CoT and VDSP, VDSP provides interpretability. First, VDSP explicitly decomposes the task into the 1) sub-chart identification; 2) reasoning; and 3) verification, making the model’s failure points transparent (e.g., mis-localization, incorrect reasoning). Second, VDSP reveals exactly which sub-chart the model selected and how it used it. This level of structured interpretability and diagnostic clarity is not available in CoT and is particularly valuable for multi-chart tasks, where transparency and traceability are important. A qualitative example illustrating the interpretability benefit of VDSP is given below.

\textbf{Qualitative Example:} Question: \emph{``Which colour is used to represent the optimized prompt for phi-2 model?''}; CoT: \emph{``Blue''}; VDSP: \emph{``The reasoning process is correct. The bottom-left chart is labeled `phi-2', and the legend clearly shows that `Optimized' is represented by blue. The blue bars in the phi-2 chart correspond to the optimized prompt. Therefore, the color used to represent the optimized prompt for the phi-2 model is blue.''}

Figures \ref{ex_step_out_vdsp}–\ref{int_step_ex_q} in \ref{VDSP_int_steps_output} show an example intermediate output. BERTScore is reported in  \ref{appendix_results_poly_4}.
 The prompts for Zero-shot, CoT, and VDSP used across RQ1-4, are listed in Figures \ref{zero}-\ref{stage3} in  \ref{appendix_prompt_ex}. 
%Table \ref{L-Acc_rq4} shows that VDSP outperforms both baselines, improving L-accuracy by up to 5.39\% over zero-shot and 4.62\% over CoT. 

\begin{table}[]
\centering
\small
\resizebox{\columnwidth}{!}{%
\begin{tabular}{lccc}
\hline
\textbf{Model} & \textbf{Zero-shot} & \textbf{CoT} & \textbf{VDSP} \\
\hline
GPT-4.1              & 0.5896& 0.5954  &0.5954  \\
Claude-3.7-Sonnet    & 0.6570 & 0.6647 & \underline{0.7110} \\
Gemini-2.0-flash     & \underline{0.6802} &\underline{0.6936}  &0.7013 \\
Pixtral-12B          &0.5067 & 0.5337 &  0.5145\\
\hline
\end{tabular}}
\caption{L-Accuracy of Zero-shot, CoT, and VDSP.}
 \label{L-Acc_rq4}% prompting methods on the human-authored QA.}\label{L-Acc_rq4}
\end{table}
% \begin{figure}[!ht]
%     \centering
%     \includegraphics[width=\columnwidth]{RQ4.png}
%     \caption{
%         Performance comparison of Zero-shot, Chain-of-Thought (CoT), and VDSP prompting methods on the human-authored QA of PolyChartQA.% The top plot shows L-Accuracy across different models, while the bottom plot presents corresponding BERT Scores.
%     }
%     \label{rq4}
% \end{figure}

% \begin{table}[t]
% \centering
% \resizebox{\columnwidth}{!}{%
% \begin{tabular}{lcc}
% \toprule
% \textbf{Model} & \makecell{\textbf{Accuracy} \\ \textbf{(Human-authored)}} & \makecell{\textbf{Accuracy} \\ \textbf{(MLM-generated)}} \\
% \midrule
% Claude-3.7-Sonnet       & 0.38 & 0.80 \\
% GPT-4.1                 & 0.33 & 0.89 \\
% Gemini-1.5-Flash        & 0.24 & 0.69 \\
% Gemini-2.0-Flash        & 0.34 & 0.75 \\
% Pixtral-12B             & 0.31 & 0.57 \\
% Llama-3.2-11B-Vision    & 0.03 & 0.21 \\
% llava-1.5-7b            & 0.05 & 0.08 \\
% Gemma-3-4b-it           & 0.10 & 0.26 \\
% \bottomrule
% \end{tabular}}
% \caption{Comparison of MLM performance (accuracy) on human-authored vs. MLM-generated questions from the \textbf{MultiChartQA} dataset.}
% \label{tab:human_vs_llm_questions}
% \end{table}
\paragraph{Error Analysis on Human-Authored Questions from PolyChartQA:}
\iffalse
The error analysis was done on the output of human-authored questions from PolyChartQA. 
\fi
We analyzed one-third of the human-authored questions that all MLMs answered incorrectly under human evaluation. %Manual inspection revealed that 
Most errors stem from line selection errors in line plots (21–26\%), visual value misreading (15–26\%), and sub-chart misidentification (15–21\%). Other error types include axis misalignment, calculation mistakes, legend association errors, multi-step reasoning failures, and incorrect value comparisons. Open-source MLMs often produce incomplete answers. Examples are provided in \ref{appendix_error}.

\section{Ablation Studies}
% We conducted ablation studies on the VDSP prompting strategy using two modified versions. In the first, the Stage 1 prompt was altered to convert each sub-chart into a structured table instead of a textual description, while Stages 2 and 3 remained unchanged; this resulted in up to a 4.8\% performance drop. {\color{red} I am lost. you said all three stages were changed. How can they be called an ablation study?} In the second variant, all three stages were modified, where Stage 2 followed a dual-persona design—an Analyst generating an answer and a Reviewer verifying it—while Stages 1 and 3 retained tasks similar to VDSP. This configuration yielded performance comparable to the zero-shot setting but with a performance drop of 2.7\% in Pixtral-12B, indicating that Stage 1 is the most critical component of the framework. Detailed results are provided in Appendix \ref{appendix_VDSP_V2} - \ref{appendix_VDSP_V3}, along with H-Accuracy on the human-authored PolyChartQA under the zero-shot setting in Appendix \ref{appendix_results_h-acc_H_poly} Table \ref{tab:hacc_human_zeroshot}.}
 \textbf{Variants of the VDSP prompting strategy:} In the first variant, Stage 1 was modified to generate structured tables instead of textual descriptions, while Stages 2 and 3 remained unchanged, resulting in a performance drop of up to 4.8\% L-Accuracy. In the second variant, all three prompts were revised, though Stages 1 and 3 retained their original roles, Stage 2 adopted a dual-persona (Analyst–Reviewer) design. This led to performance comparable to the Zero-shot setting, with a 2.7\% L-Accuracy drop in Pixtral-12B, confirming Stage 1 as the most critical component. Detailed results are provided in \ref{appendix_VDSP_V2} - \ref{appendix_VDSP_V3}. 
 
 \noindent\textbf{Overall H-Accuracy:} H-Accuracy ranges from 9.06\%-70.52\% under Zero-shot setting, and 6.17\%-71.87\% under CoT setting on human-authored PolyChartQA as presented in Table \ref{tab:hacc_human_zeroshot}. As shown in Table \ref{tab:ha_la_accuracy_cot} in \ref{appendix_results_h-acc_H_poly}, except for Gemini-2.0-flash and ChartGemma, the absolute difference in accuracy between the human and the LLM judges is $\leq$ 1.2\% across all seven models under the CoT setting.
 \begin{table}[hbpt]
\centering
\begin{tabular}{lcc}
\hline
\textbf{Model} & \textbf{Zero-shot} & \textbf{CoT} \\
\hline
Claude-3.7-Sonnet & 0.6532 &0.6609 \\
GPT-4.1 & 0.6108 &0.5954\\
Gemini-2.0-flash & \underline{0.7052}&\underline{0.7187} \\
Pixtral-12B & 0.5318&0.5376 \\
LLaMA-3.2-11B & 0.1522 &0.0809\\
LLaVA-1.5-7B & 0.1233 &0.1175\\
Gemma-3-4b-it & 0.2119 &0.2274 \\
MatCha & 0.0906 &0.0617\\
ChartGemma & 0.1927&0.2351 \\
\hline
\end{tabular}
\caption{H-Accuracy on human-authored PolyChartQA under the Zero-shot and Chain-of-Thought settings.\label{tab:hacc_human_zeroshot}}
\end{table}

 \noindent\textbf{Robustness of our findings across LLM judges:} The human–vs–MLM QA performance trend remains consistent when using GPT-4.1 and Gemini-2.0-flash as LLM-judges. In Zero-shot human QA, closed-source MLMs reach 63.2\%–72.1\% (GPT-4.1 judge) and 67.4\%–74.8\% (Gemini-2.0-flash judge), while open-source MLMs reach 12.9\%–53.4\% and 17.0\%–56.3\%, respectively. %For human-authored QA in the zero-shot setting, closed-source models achieve 63.2–72.1\% accuracy under GPT-4.1 and 67.4–74.8\% under Gemini-2.0-flash, while open-source models perform substantially lower (12.9–53.4\% and 17.0–56.3\%, respectively). 
On MLM-generated QA, closed-source MLMs reach 78.9\%–89.7\% (GPT-4.1) and 81.5\%–90.8\% (Gemini-2.0-flash), and open-source MLMs obtain 15.8\%–69.2\% and 23.1\%–72.0\%. Detailed results are in \ref{multi_judge_comp_gpt_gemini} Table \ref{tab:human_mlm_judge_comparison}.

% MLM performance chart type (Appendix XX) and L-accuracy computed using GPT-4.1 (Appendix XXX)}.
\section{Conclusion}
We introduce PolyChartQA, a benchmark for multi-chart QA, showing that the task is substantially harder and exhibits large performance gaps between human- and MLM-authored questions. Task decomposition and self-verification prompting help improve MLM performance and highlight key reasoning and  failures.
Our work facilitates further studies on multi-chart QA. %Future work will address multi-chart complexity.%Future work should examine why MLMs generate correct questions but incorrect answers.
%Future work should explore why MLMs often generate valid questions but incorrect answers.
% We introduced PolyChartQA, a large-scale benchmark designed for evaluating question answering capabilities over multi-chart figures. Unlike prior datasets focused primarily on single-chart scenarios, PolyChartQA emphasizes the unique reasoning and visual challenges posed by multi-chart settings. Our extensive evaluation reveals that multi-chart question answering is substantially more difficult for MLMs to answer than its single-chart counterpart.
% A significant performance gap between human-authored versus MLM-generated questions highlights a limitation in the models’ ability to generalize to real-world and diverse question formulations. 
% We found that the decomposition of the task and self-evaluating of the answer helps MLMs improve performance. Our error analysis reveals  
% primary error sources and can help future researchers in designing their model. While manually filtering MLM-generated questions, we found many questions that were correct and interesting, but the MLM-generated answers were wrong. For future work, it will be worthwhile to investigate why MLM generates a wrong answer while simultaneously generating a correct question.
 %Although the closed-source models demonstrate relatively higher L-accuracy, most open-source models, with the exception of Pixtral-12B, exhibit poor performance on the multi-chart question answering task. %Chart homogeneity also influences performance. %that MLMs primarily struggle with the interpretation of visual values
\section{Limitations}
Despite its contributions, PolyChartQA has a few limitations. First, human-authored QA creation is costly and time-intensive; thus, the dataset includes 519 human-authored QAs, complemented by MLM-generated QAs that are fully manually verified, prioritizing quality over scale. 
%Despite the contributions of this work, some limitations remain. First, crafting human-authored question-answer pairs is both time-intensive and costly, as acknowledged in prior works. Our dataset contains 519 human-authored QA pairs. To increase the number of questions, we used MLM to generate additional QA pairs and manually verified them. This process remains time-consuming, limiting the number of generated questions, but ensuring the quality of the questions. While previous works generate large-scale QA pairs using MLM, they do not perform manual verification of all the generated QA pairs \cite{pramanick2025spiqadatasetmultimodalquestion,shinoda2024sbsfigurespretrainingfigure}. Although our dataset is not large-scale, it is fully human-verified, ensuring high-quality and reliable annotations.

% Second, while LLM-based evaluations are inherently non-deterministic and may exhibit model-specific variations in ambiguous cases, we mitigated this by employing an LLM judge that demonstrated exceptionally high alignment with human evaluations, thereby ensuring reliable assessment. 

Second, our dataset exhibits limited diversity along two dimensions. (1) All images are drawn from computer science research papers published in 2024, which may constrain external validity beyond academic figures and other disciplines. However, the generated questions do not require domain-specific knowledge and can be answered solely using the information presented in the charts. Moreover, the chart types used (e.g., bar charts, line charts) are common across domains and widely used in real-world applications. (2) The dataset is predominantly homogeneous by chart type (85.58\%). Although this distribution reflects the real-world distribution in contemporary research papers, our findings are therefore more conclusive for homogeneous multi-chart images, with relatively limited coverage of heterogeneous multi-chart reasoning. %it limits coverage of heterogeneous multi-chart reasoning.
Addressing broader disciplinary sources and increasing chart-type diversity are important directions for future extensions of PolyChartQA.

\section{Ethical Considerations and Potential Risks}
All charts used in our study were sourced from open-access computer science research papers. No personally identifiable information or sensitive content was used in the construction of the dataset.\\
ACL materials are licensed under a Creative Commons Attribution 4.0 International License, which covers papers collected from EMNLP. We selected multi-chart images from open-access EMNLP, ICML, ICSE, and SIGCOMM publications or corresponding preprints in arXiv with a Creative Commons International License.  We list the image names from the articles in a separate file and give credit to the original authors. 

While we make our question-answers available under Commons Attribution License (CC BY 4.0), all chart images are subject to their respective copyrights by the authors of the respective papers.\\
Additionally, we used PDFFigures 2.0 to extract the multi-chart images from the papers. PDFFigures 2.0 is publicly available under Apache License 2.0, allowing free use. We have cited the creators. The MultiChartQA dataset is available under the Creative Commons Attribution-NonCommercial 4.0 International (CC BY-NC 4.0) license that permits noncommercial use, sharing, and adaptation of the material, provided appropriate credit is given, changes are indicated, and the use remains noncommercial. We cited the creator of the artifacts and discussed the changes we made for our use.

Beyond the generation of QA pairs using MLMs, AI assistants were employed solely for non-substantive tasks such as language refinement, formatting, and basic debugging.\\
 
\section*{Acknowledgments}
This work is partially supported by the National Science Foundation under Grant No. 2152117. Any opinions, findings, and conclusions or recommendations expressed in this material are those of the author(s) and do not necessarily reflect the views of the National Science Foundation.

We acknowledge the valuable contribution of our independent annotator Tanisha Ravindran for the effort in reviewing and correcting the question-answer annotations.

\bibliography{custom}

\appendix

\section{Appendix}
\subsection{Distribution of Multi-chart Images in PolychartQA \label{Appendix_distribution_chart_image}} The dataset has these percentages of images from different conferences: EMNLP (65\%), SIGCOMM (17\%), ICSE (10\%), and ICML (8\%).
% \iffalse
% Figure \ref{domain_distribution} presents the image distribution of multi-chart images by conferences in PolyChartQA.\\
% \begin{figure}[!h]
% \centering
% \includegraphics[width=\linewidth]{image_distribution.png}
% \caption{Distribution of multi-chart images by conferences.}
% \label{domain_distribution}
% \end{figure}
% \fi
Figure \ref{sub_chart_count} presents the distribution of sub-chart counts in PolyChartQA. The highest number of sub-charts in an image is 72. PolyChartQA includes eleven chart types: Bar Chart, Scatter Plot, Spider Chart, Histogram, Line Chart, Box Plot, Point Plot, Surface Plot, Pie Chart, Area Chart, and Violin Plot. The distribution of chart types across homogeneous multi-chart images from PolyChartQA is shown in Figure \ref{chart_type}.\\

\begin{figure*}[ht]
\centering
\includegraphics[width=\textwidth]
{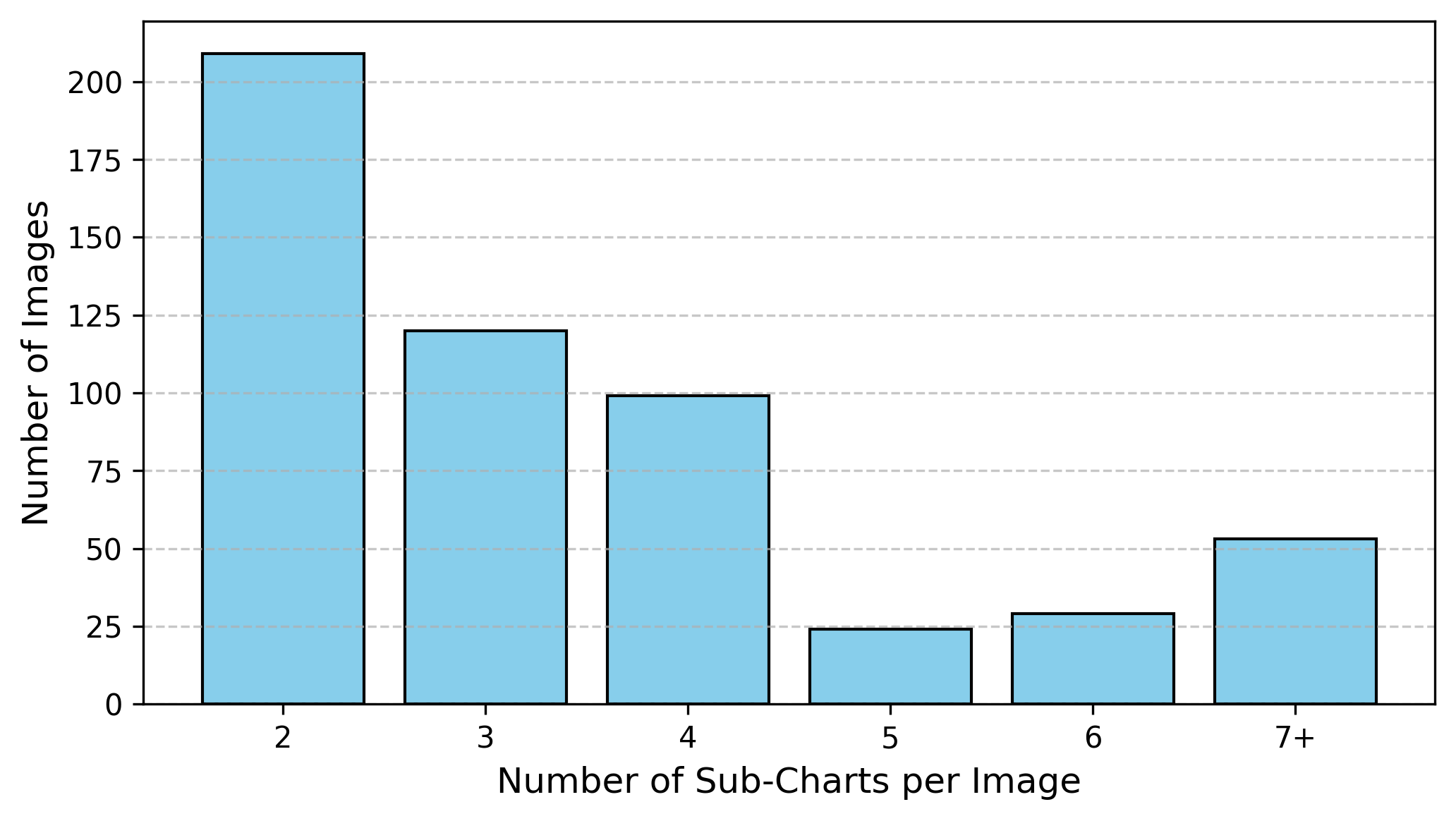}
\caption{Distribution of sub-chart counts in PolyChartQA multi-chart images.}
\label{sub_chart_count}
\end{figure*}

\begin{figure*}[ht]
\centering
\includegraphics[width=\textwidth]
{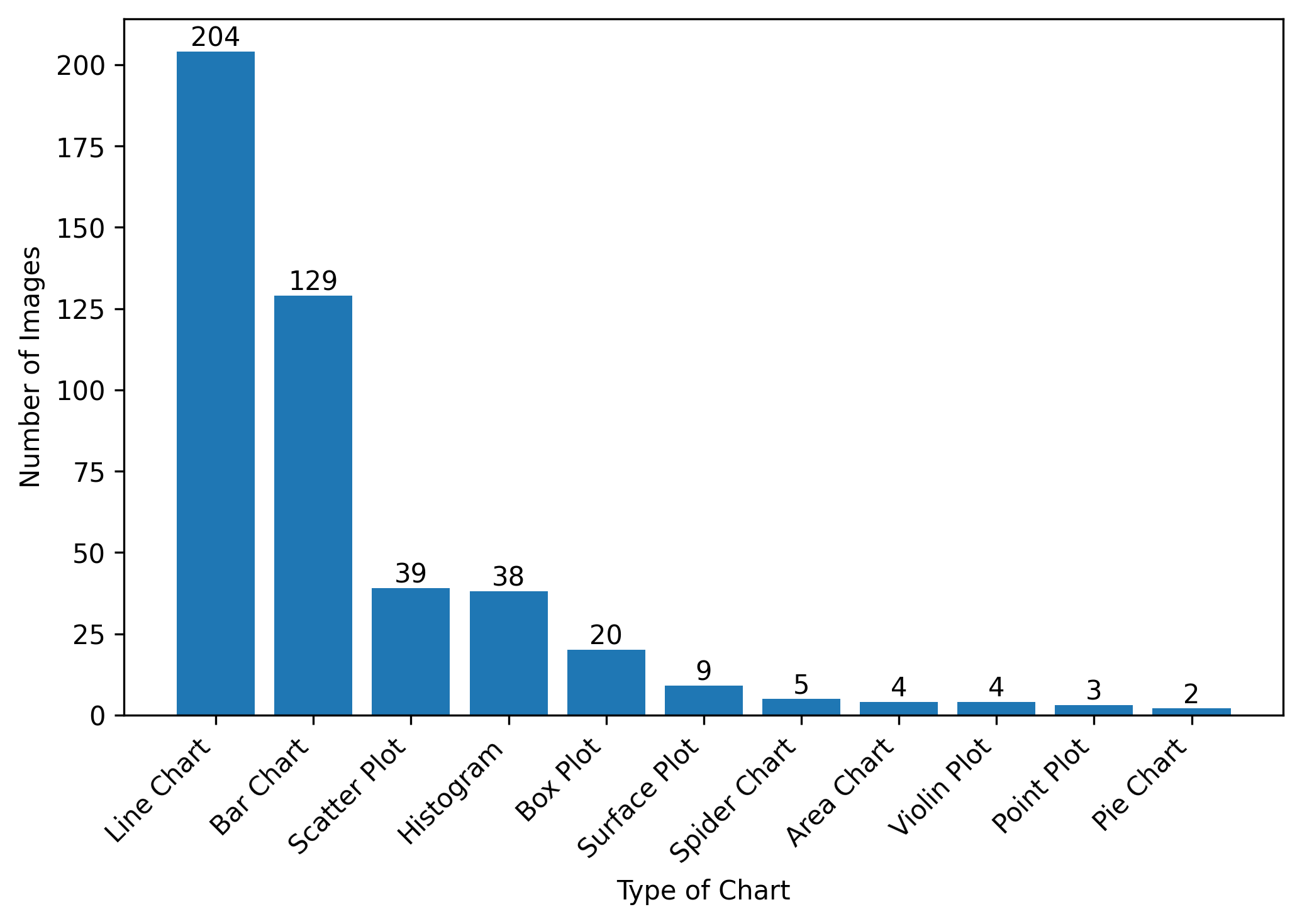}
\caption{Distribution of chart types across homogeneous multi-chart images.}
\label{chart_type}
\end{figure*}

\subsection{Criteria for Removing MLM-generated Questions}\label{criteria_to_filter_ques}
Table \ref{qa_removal_criteria} presents the criteria used to remove MLM-generated QA. 
\begin{table}[hbt]
\centering
\resizebox{\columnwidth}{!}{
\begin{tabular}{l}
\hline
\textbf{Criterion}  \\
\hline
QA explicitly mentions chart position (e.g., "left chart") \\
QA pair is partially or fully incorrect \\
Question is derived from the caption, not the chart itself \\
Question has multiple valid answers or is unclear \\
Errors in symbols (e.g., $\gamma, \lambda$) or formatting \\
Hallucinated content \\
QA does not follow the instructions\\
QA based on existing knowledge instead of chart \\
\hline
\end{tabular}}
\caption{Criteria for removing MLM-generated QA pairs.}
\label{qa_removal_criteria}
\end{table}

\subsection{Agreement between LLM Judges and Human Evaluation}\label{agreement_llm_human}
Table \ref{tab:kappa_agreement} reports the Cohen’s Kappa agreement between LLM-judge and human evaluation on randomly sampled 100 QA pairs from PolyChartQA. Table~\ref{tab:agreement} reports the percent agreement between human evaluations and different LLM judges across all evaluated models, illustrating the reliability of each LLM judge.

\begin{table*}[ht]
\centering
\resizebox{\textwidth}{!}{%
\begin{tabular}{lccccccc}
\toprule
\textbf{Judge Model} & \textbf{Claude-3.7-Sonnet} & \textbf{GPT-4.1} & \textbf{Gemini-2.0-flash} & \textbf{Gemma-3-4b-it } & \textbf{LLaMA-3.2-11B} & \textbf{LLaVA-1.5-7B } & \textbf{Pixtral-12B} \\
\midrule
Claude-3.7-Sonnet & 0.852 & 0.920 & 0.884 & 0.834 & 0.734 & 0.864 & 0.900 \\
GPT-4.1           & 0.870 & 0.774 & 0.829 & 0.777 & 0.649 & 0.942 & 0.860 \\
Gemini-2.0-flash  & 0.780 & 0.730 & 0.753 & 0.455 & 0.481 & 0.684 & 0.840 \\
\bottomrule
\end{tabular}}
\caption{Cohen’s Kappa agreement between LLM-judge and human accuracy under Zero-shot setting. Higher positive values indicate stronger agreement.}
\label{tab:kappa_agreement}
\end{table*}

\begin{table*}[ht]
\centering
\resizebox{\textwidth}{!}{%
\begin{tabular}{lcccccccc}
\toprule
\textbf{Judge} & \textbf{Claude-3.7-Sonnet} & \textbf{GPT-4.1} &  \textbf{Gemini-2.0-flash} & \textbf{Gemma-3-4b-it } & \textbf{LLaMA-3.2-11B} & \textbf{LLaVA-1.5-7B } & \textbf{Pixtral-12B} \\
\midrule
\textbf{Claude-3.7-Sonnet}   & 93 & 96  & 95 & 96 & 95 & 98 & 95 \\
\textbf{GPT-4.1}      & 94 & 89  & 93 & 94 & 92 & 99 & 93 \\
\textbf{Gemini-2.0-flash}   & 90 & 87  & 90 & 81 & 84 & 93 & 92 \\
\bottomrule
\end{tabular}}
\caption{Percent agreement (\%) between human evaluation and LLM judges across evaluated models. Columns represent the evaluated models, and rows indicate the LLM acting as the evaluation judge.}\label{tab:agreement}
\end{table*}
\subsection{L-Accuracy Comparison with Different Judges}\label{multi_judge_comp_gpt_gemini}
Table \ref{tab:human_mlm_judge_comparison} presents the L-accuracy on PolyChartQA when judged by GPT-4.1 and Gemini-2.0-flash.
\begin{table*}[]
\centering
\small
\setlength{\tabcolsep}{6pt}
\begin{tabular}{lcccc}
\hline
\multirow{2}{*}{\textbf{Model}} & 
\multicolumn{2}{c}{\textbf{Human-Authored QA}} & 
\multicolumn{2}{c}{\textbf{MLM-Generated QA}} \\
\cmidrule(lr){2-3} \cmidrule(lr){4-5}
 & \textbf{GPT-4.1 Judge} & \textbf{Gemini-2.0-flash Judge} & \textbf{GPT-4.1 Judge} & \textbf{Gemini-2.0-flash Judge} \\
\hline
Claude-3.7-sonnet        & 0.692 & 0.719 & 0.789 & 0.815 \\
GPT-4.1           & 0.632 & 0.674 & 0.897 & 0.908 \\
Gemini-2.0-flash      & 0.721 & 0.748 & 0.851 & 0.865 \\
Gemma-3-4b-it         & 0.241 & 0.326 & 0.340 & 0.463 \\
LLaMA-3.2-11B         & 0.225 & 0.303 & 0.267 & 0.328 \\
LLaVA1.5-7b         & 0.129 & 0.170 & 0.158 & 0.231 \\
Pixtral-12B   & 0.534 & 0.563 & 0.692 & 0.720 \\
\hline
\end{tabular}
\caption{L-Accuracy comparison on Human-authored QA and MLM-generated QA, evaluated using GPT-4.1 and Gemini-2.0-flash judges.}
\label{tab:human_mlm_judge_comparison}
\end{table*}

\subsection{Model Configuration}\label{model_config} For reproducibility, we set the temperature = 0, and seed = 5 (GPT-4.1, Pixtral-12B) or sample = false (Llama-3.2-11B-Vision, Llava-1.5-7b, and Gemma-3-4b-it). For open-sourced MLMs (except Pixtral-12B), we ran these models in Google Colab with A100/L4 GPU. GPT-4.1, Claude-3.7-Sonnet, Gemini-2.0-flash, and Pixtral-12B were used through API calls. BERTScore is computed using the bertscore Python package (v0.3.13) with the bert-base-uncased model and default parameters.
\subsection{Results across Question Difficulty Levels on PolyChartQA\label{appendix_results_qdiff}}
Figure \ref{h_acc_diff} presents the detailed human evaluation result on human-authored QAs from PolyChartQA under the Zero-shot setting across question difficulty levels. Figures \ref{l_acc_diff_cot}-\ref{bert_acc_diff_cot} show the L-accuracy under CoT and BERTScore under the Zero-shot and CoT settings on PolyChartQA across question difficulty levels.

\begin{figure*}[]
\centering
\includegraphics[width=\textwidth]
{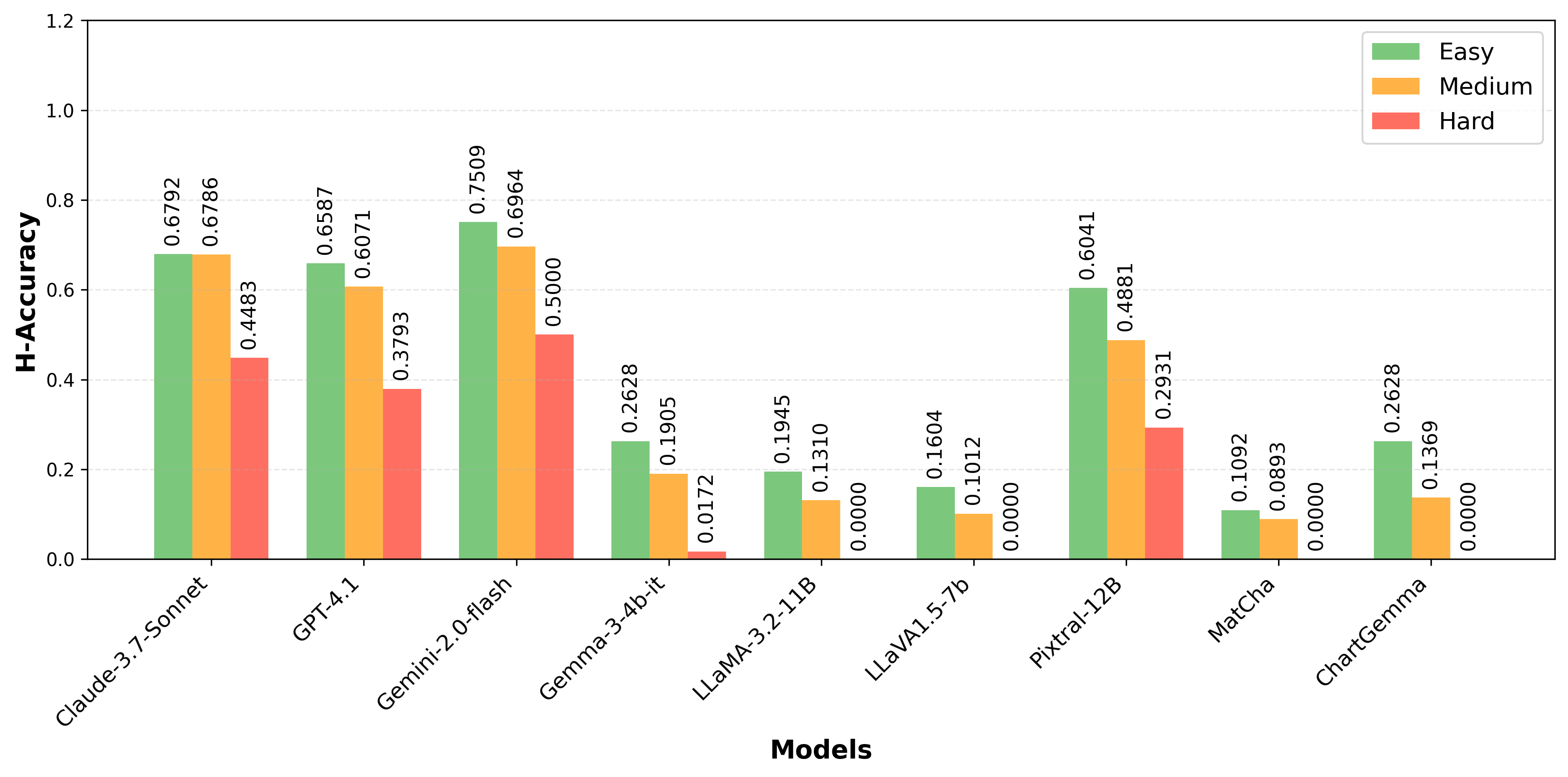}
\caption{H-Accuracy on human-authored QAs from PolyChartQA under the Zero-shot setting across question difficulty levels.}
\label{h_acc_diff}
\end{figure*}

% \begin{figure*}[ht]
% \centering
% \includegraphics[width=\textwidth]
% {l_accuracy_zero_2.1.png}
% \caption{L-Accuracy on PolyChartQA under the zero-shot setting across question difficulty levels.}
% \label{l_acc_diff_zero}
% \end{figure*}

\begin{figure*}[]
\centering
\includegraphics[width=\textwidth]
{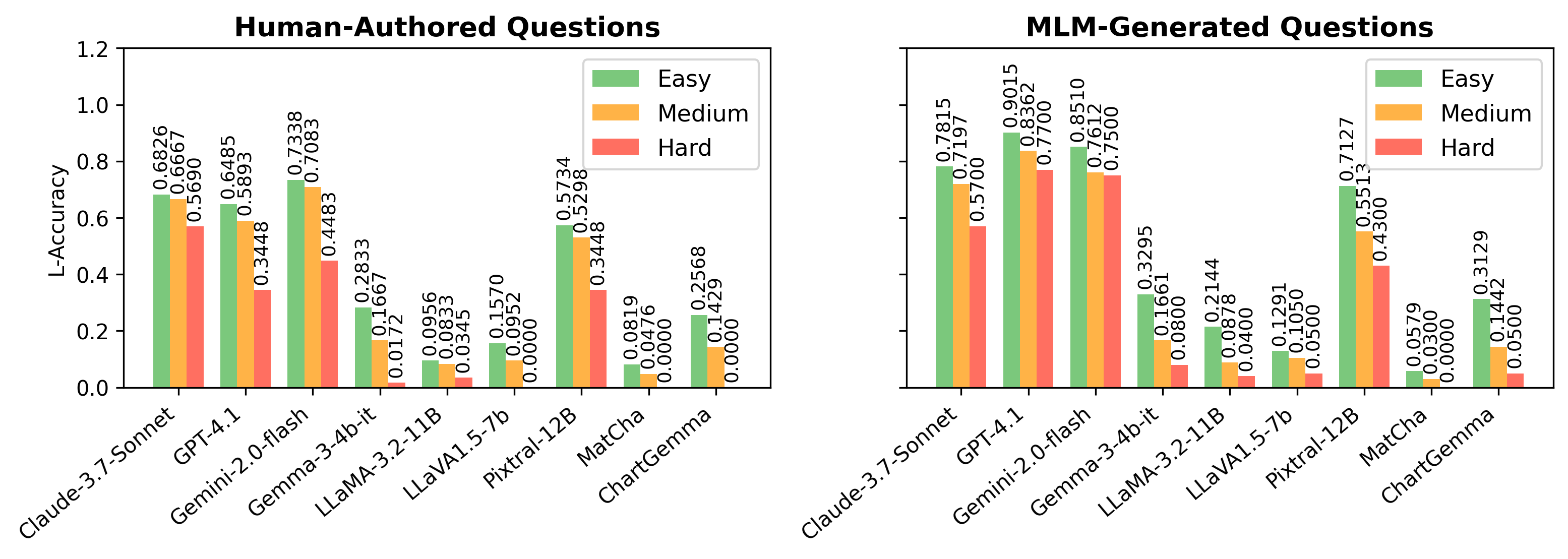}
\caption{L-Accuracy on PolyChartQA under the chain-of-thought setting across question difficulty levels.}
\label{l_acc_diff_cot}
\end{figure*}

\begin{figure*}[]
\centering
\includegraphics[width=\textwidth]
{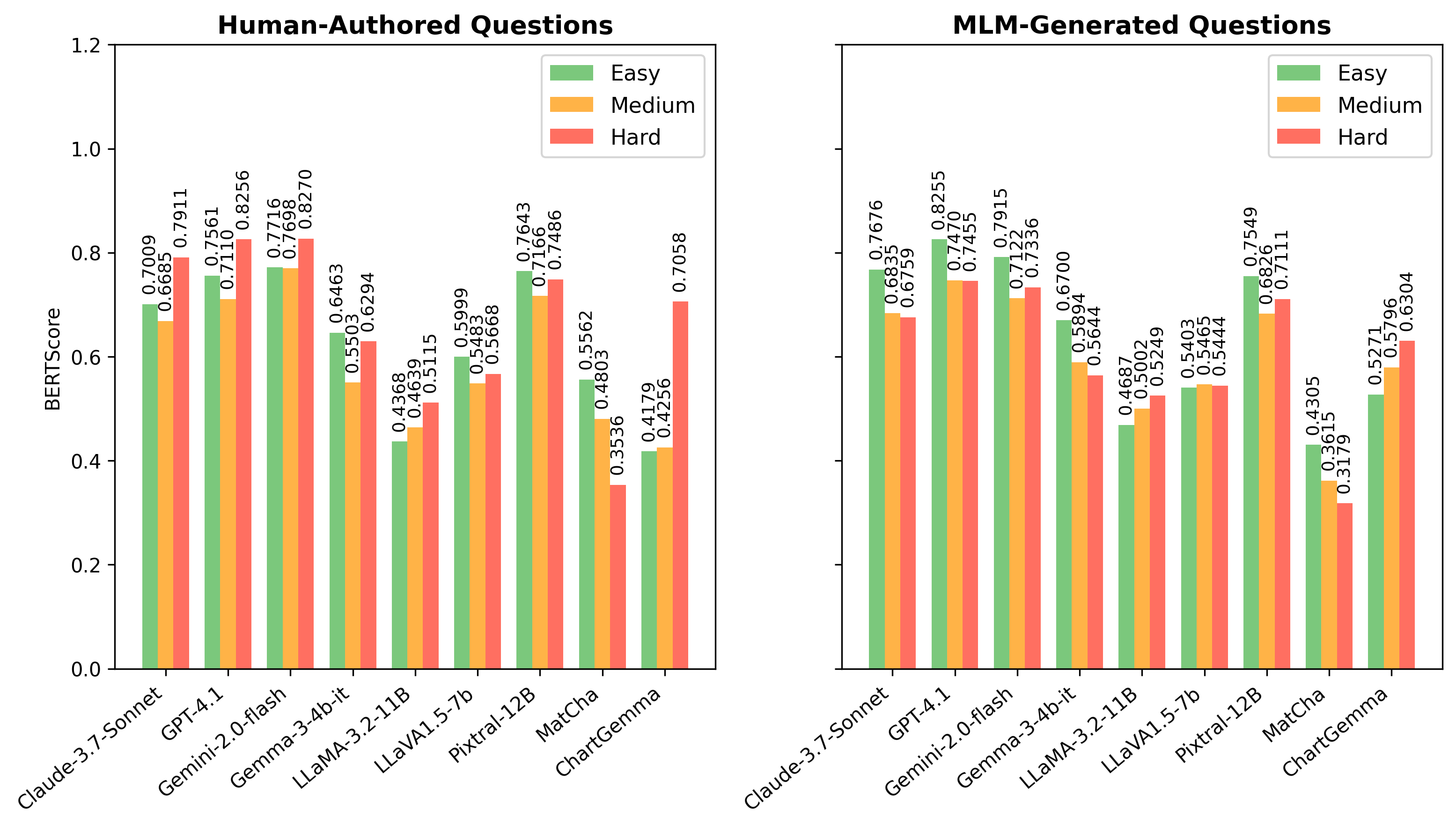}
\caption{BERTScore on PolyChartQA under Zero-shot setting across question difficulty levels.}
\label{bert_acc_diff_zero}
\end{figure*}

\begin{figure*}[]
\centering
\includegraphics[width=\textwidth]
{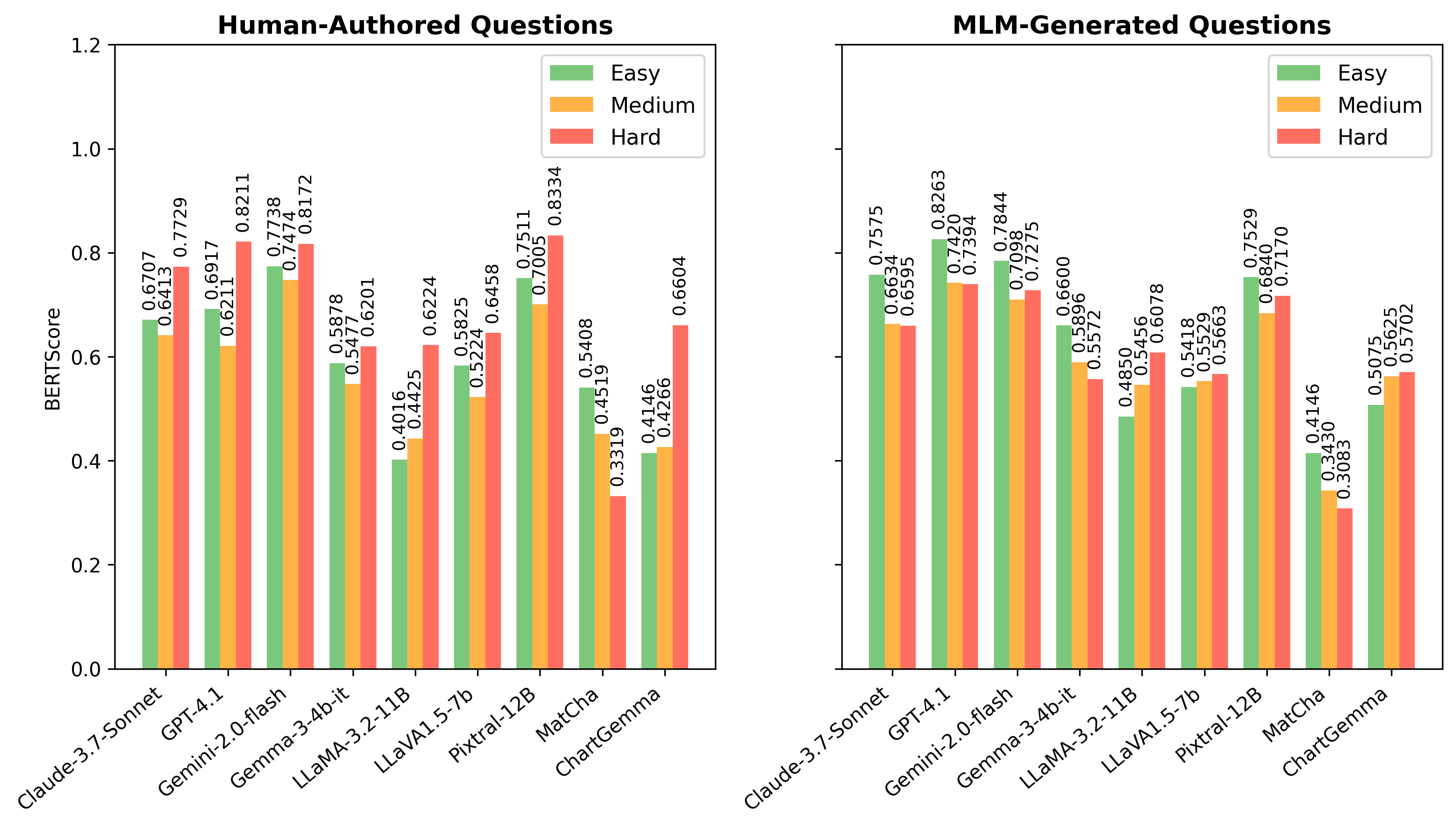}
\caption{BERTScore on PolyChartQA under chain-of-thought setting across question difficulty levels.}
\label{bert_acc_diff_cot}
\end{figure*}

\subsection{Results across Question Types on PolyChartQA\label{appendix_results_qtype}}

Figure \ref{h_acc_qtype} presents the H-Accuracy result on human-authored QAs from PolyChartQA under the Zero-shot setting across question types. Figures \ref{l_acc_qtype_cot}-\ref{bert_acc_qtype_cot} show the L-accuracy under CoT and BERTScore under the Zero-shot and CoT settings on PolyChartQA across question types.

\begin{figure*}[]
\centering
\includegraphics[width=\textwidth]
{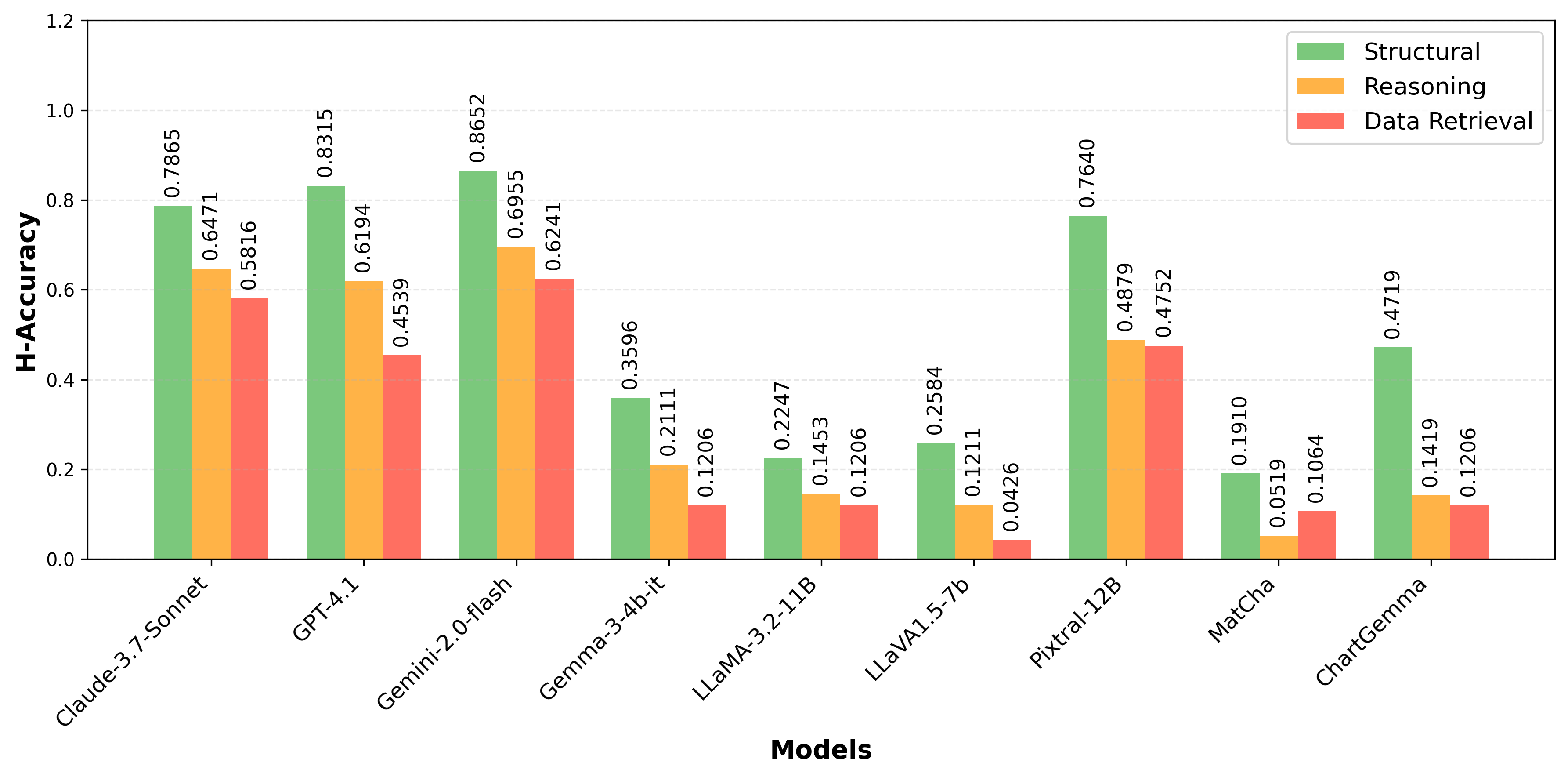}
\caption{H-Accuracy on human-authored QAs from PolyChartQA under Zero-shot setting across question types.}
\label{h_acc_qtype}
\end{figure*}

% \begin{figure*}[ht]
% \centering
% \includegraphics[width=\textwidth]
% {l_accuracy_zero_2.2.png}
% \caption{L-Accuracy on PolyChartQA under zero-shot setting across question types.}
% \label{l_acc_qtype_zero}
% \end{figure*}

\begin{figure*}[]
\centering
\includegraphics[width=\textwidth]
{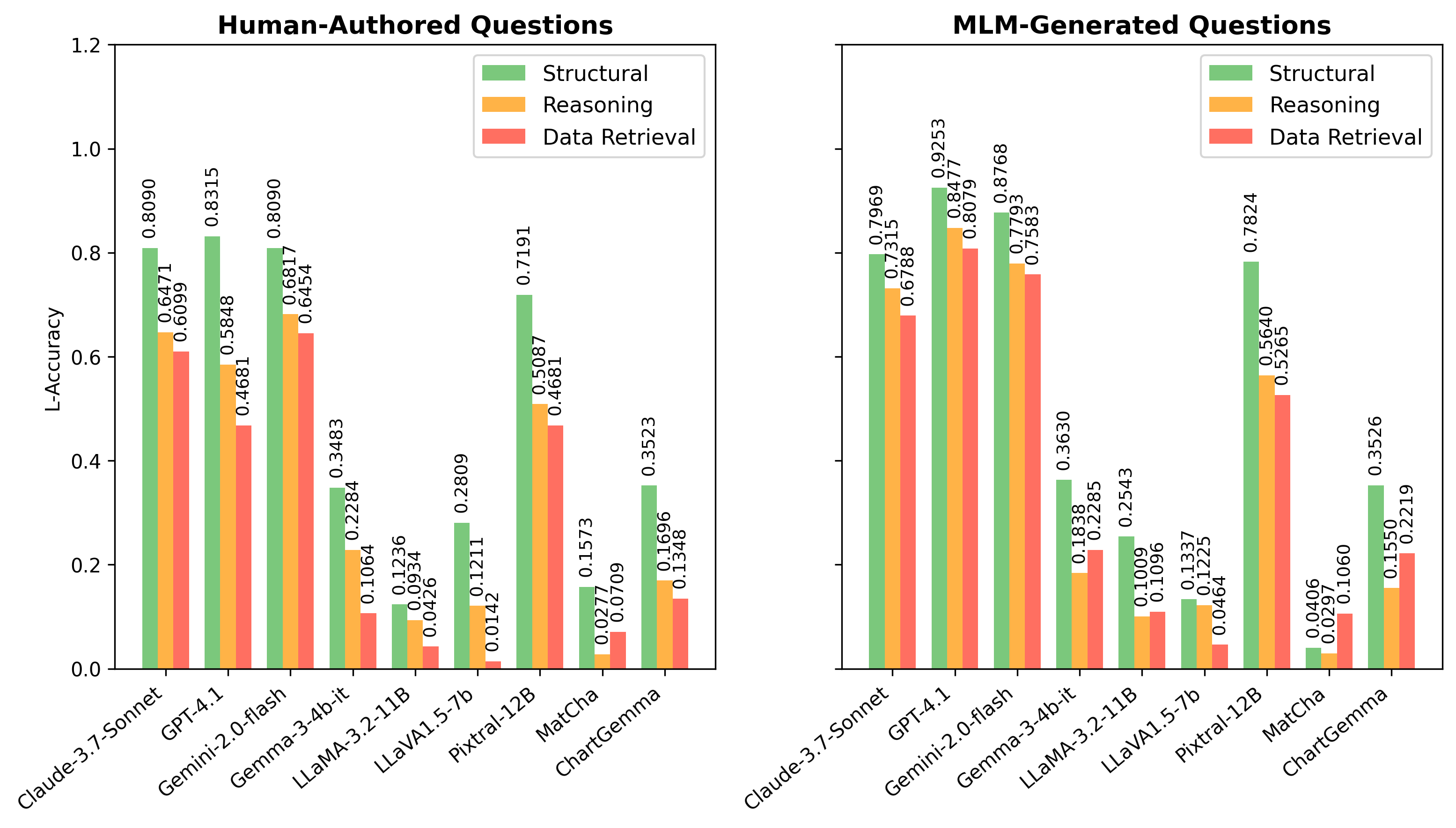}
\caption{L-Accuracy on PolyChartQA under CoT setting across question types.}
\label{l_acc_qtype_cot}
\end{figure*}

\begin{figure*}[]
\centering
\includegraphics[width=\textwidth]
{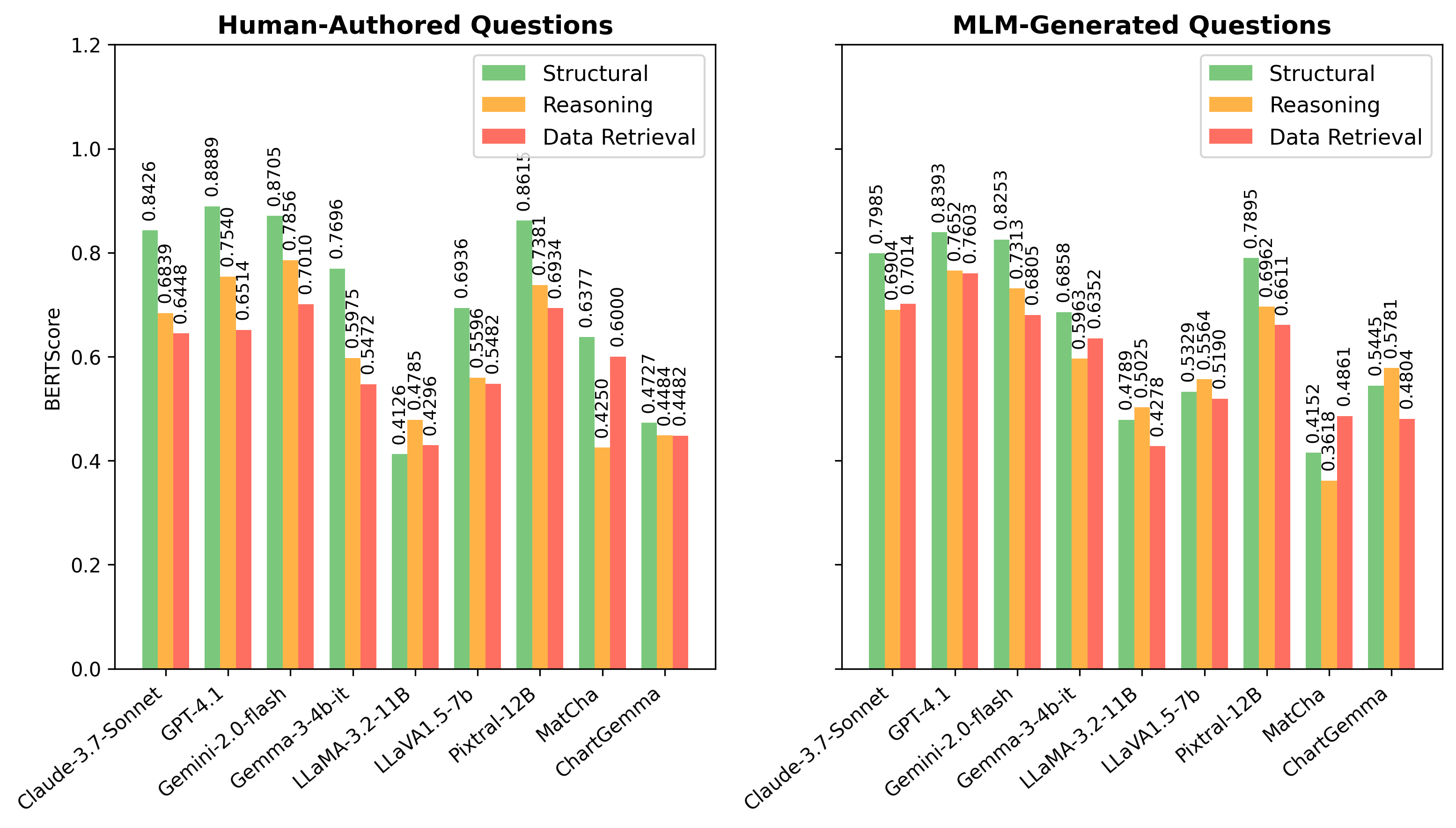}
\caption{BERTScore on PolyChartQA under Zero-shot setting across question types.}
\label{bert_acc_qtype_zero}
\end{figure*}

\begin{figure*}[]
\centering
\includegraphics[width=\textwidth]
{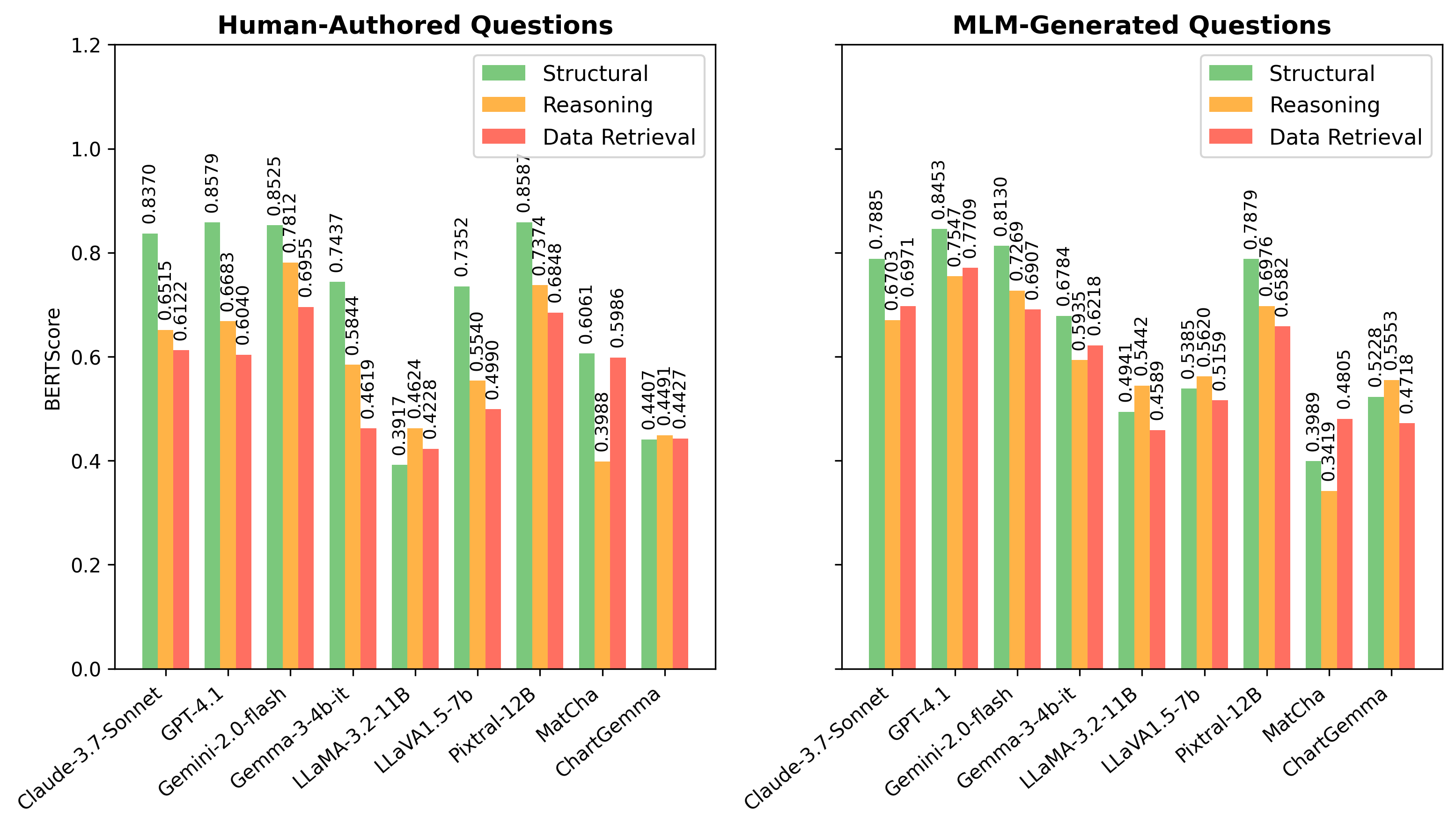}
\caption{BERTScore on PolyChartQA under CoT setting across question types.}
\label{bert_acc_qtype_cot}
\end{figure*}

\subsection{Results across Chart Homogeneity on PolyChartQA\label{appendix_results_homo}}

Figure \ref{h_acc_homo} presents the detailed human evaluation result on human-authored QAs from PolyChartQA under the Zero-shot setting across chart homogeneity. Figures~\ref{l_acc_homo_zero}-\ref{bert_acc_homo_cot} show the L-accuracy and BERTScore under Zero-shot and CoT settings on PolyChartQA across chart homogeneity.

\begin{figure*}[]
\centering
\includegraphics[width=\textwidth]
{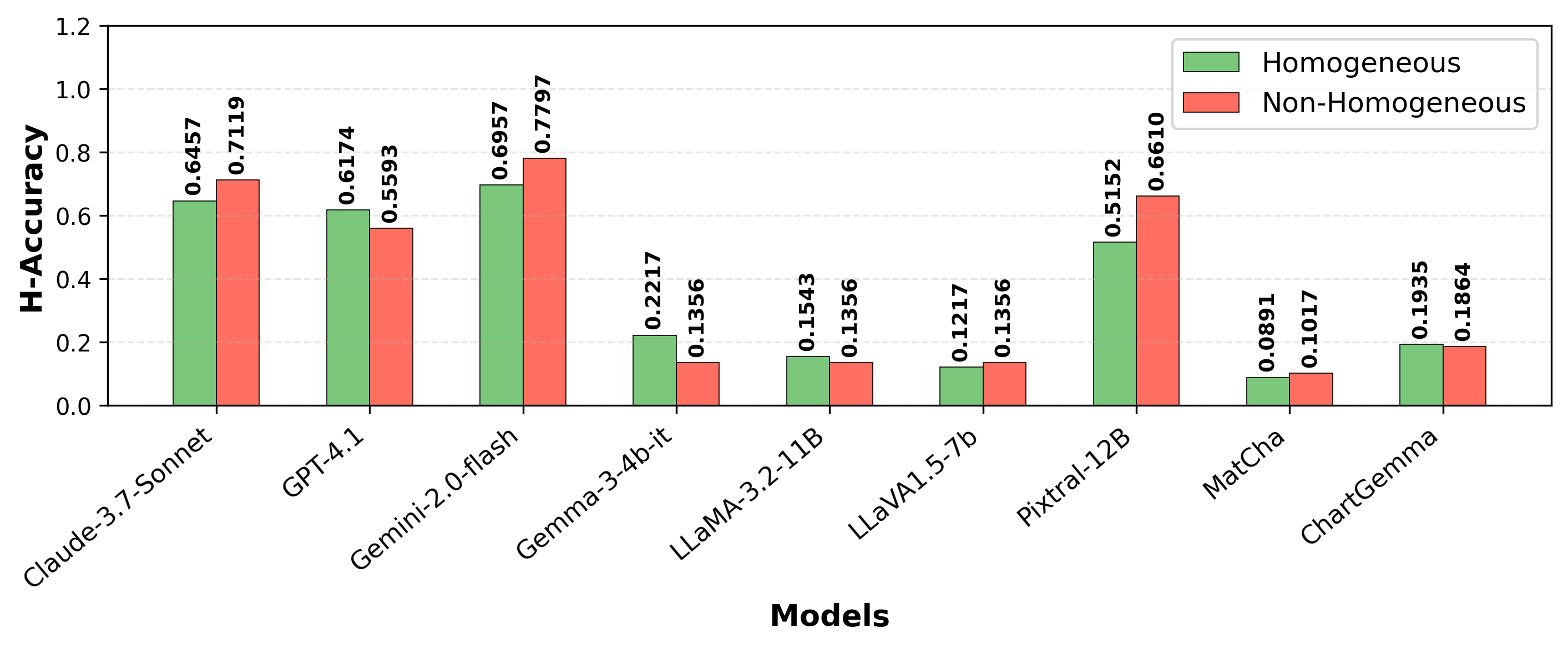}
\caption{H-Accuracy on human-authored QAs from PolyChartQA under Zero-shot setting across chart homogeneity.}
\label{h_acc_homo}
\end{figure*}

\begin{figure*}[]
\centering
\includegraphics[width=\textwidth]
{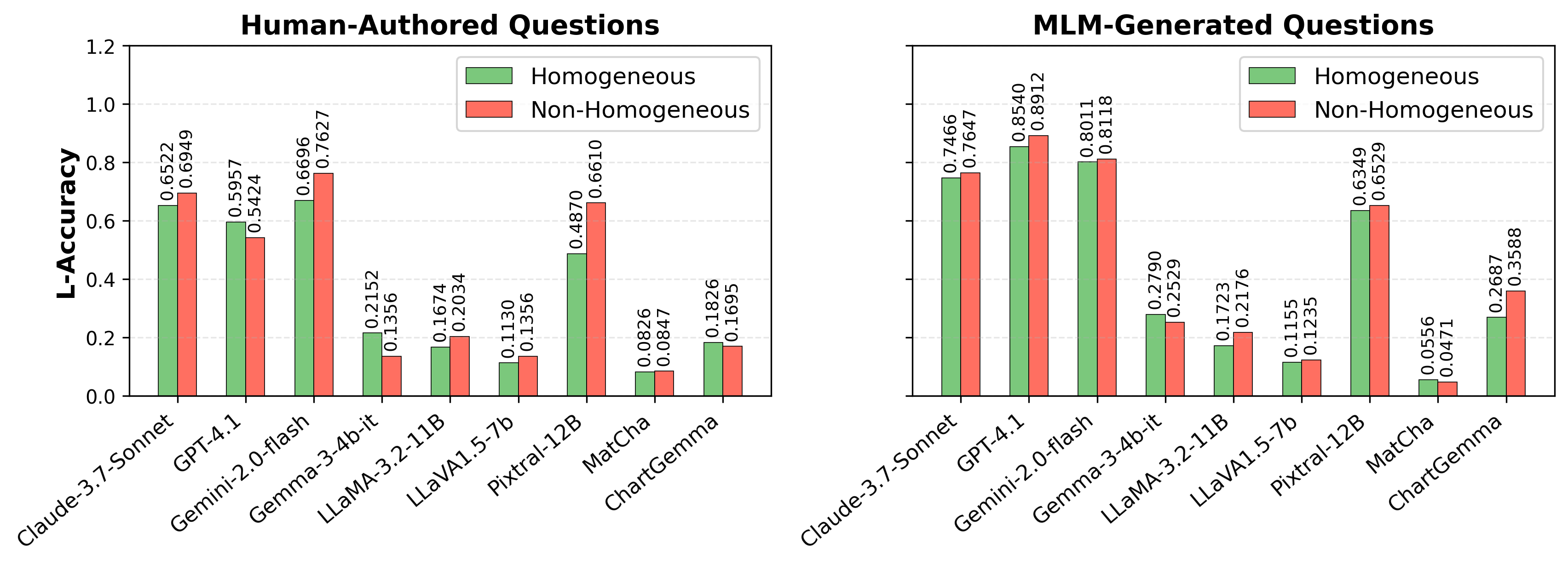}
\caption{L-Accuracy on PolyChartQA under Zero-shot setting across chart homogeneity.}
\label{l_acc_homo_zero}
\end{figure*}

\begin{figure*}[]
\centering
\includegraphics[width=\textwidth]
{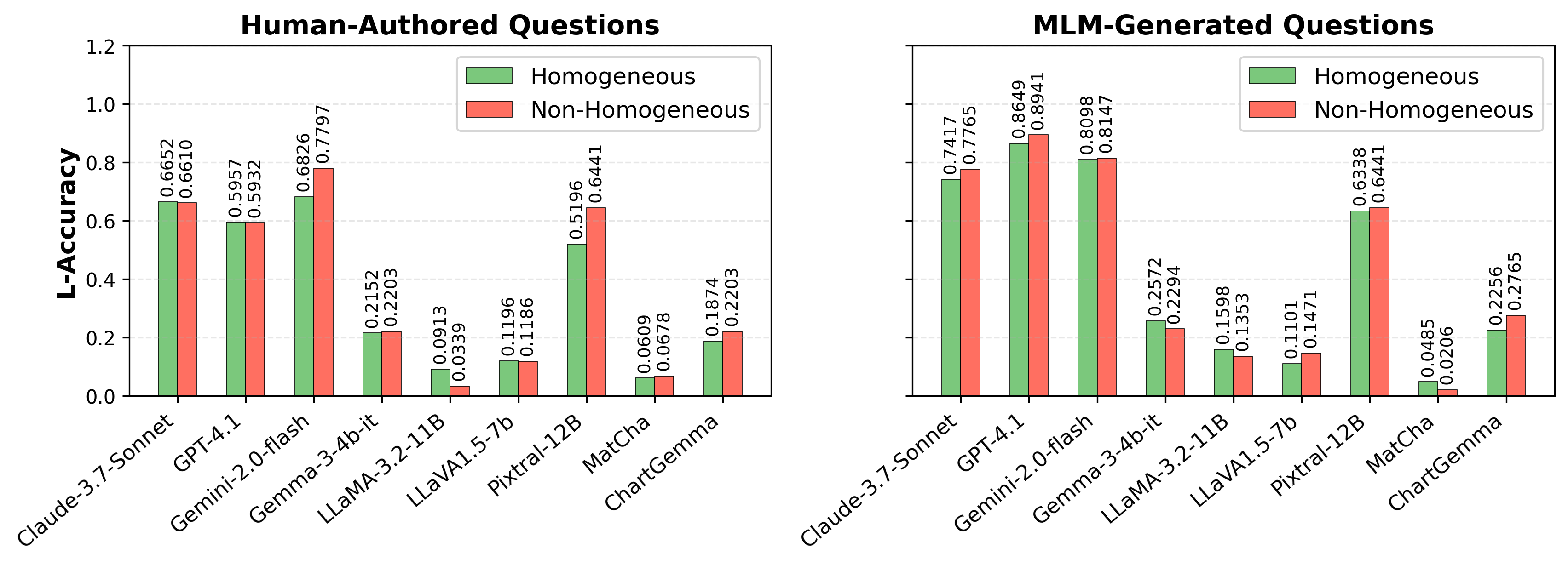}
\caption{L-Accuracy on PolyChartQA under CoT setting across chart homogeneity.}
\label{l_acc_homo_cot}
\end{figure*}

\begin{figure*}[]
\centering
\includegraphics[width=\textwidth]
{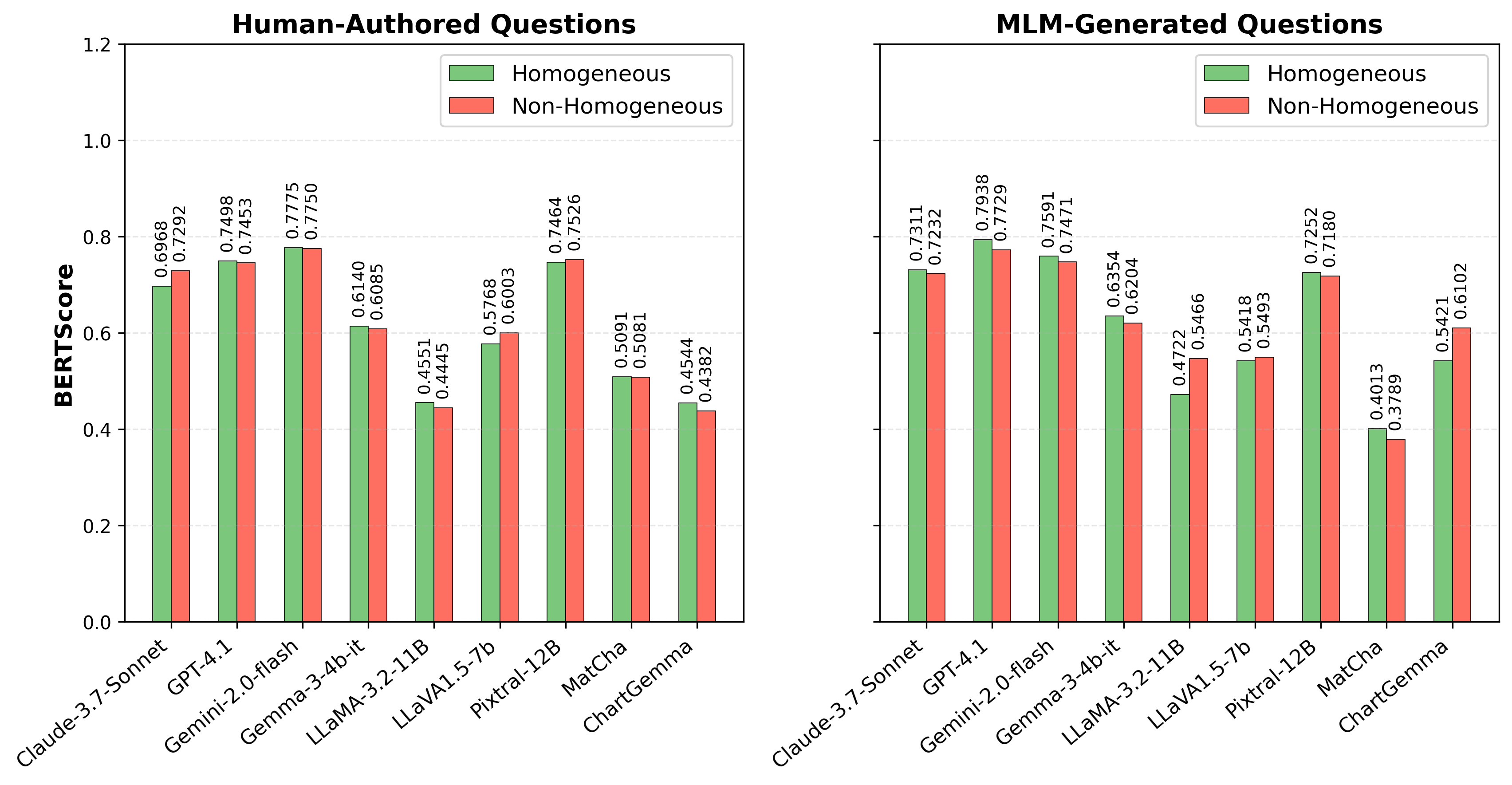}
\caption{BERTScore on PolyChartQA under Zero-shot setting across chart homogeneity.}
\label{bert_acc_homo_zero}
\end{figure*}

\begin{figure*}[]
\centering
\includegraphics[width=\textwidth]
{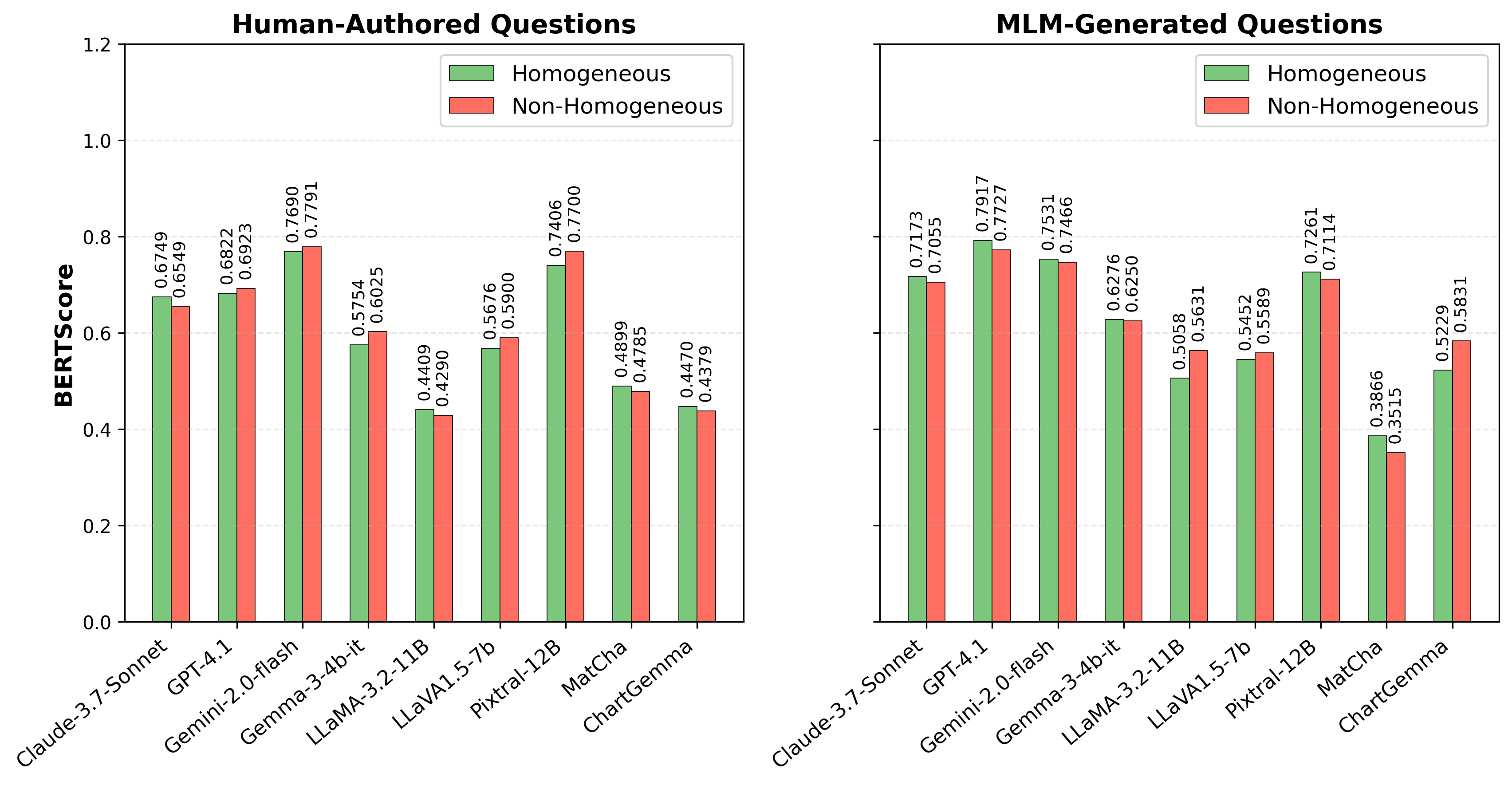}
\caption{BERTScore on PolyChartQA under CoT setting across chart homogeneity.}
\label{bert_acc_homo_cot}
\end{figure*}

\subsection{MLM-Performance based on the Number of Sub-charts}\label{appendix_result_subchart}
Figure \ref{H_acc_sub_zero_hum} presents the detailed human evaluation result on human-authored QAs from PolyChartQA under the Zero-shot setting, grouped by the number of sub-charts per image. Figures \ref{l_acc_sub_zero_hum}–\ref{l_acc_sub_cot_mlm} present the L-Accuracy trends under the Zero-shot and CoT settings on PolyChartQA, grouped by the number of sub-charts per image. 

\begin{figure*}[]
\centering
\includegraphics[width=\textwidth]
{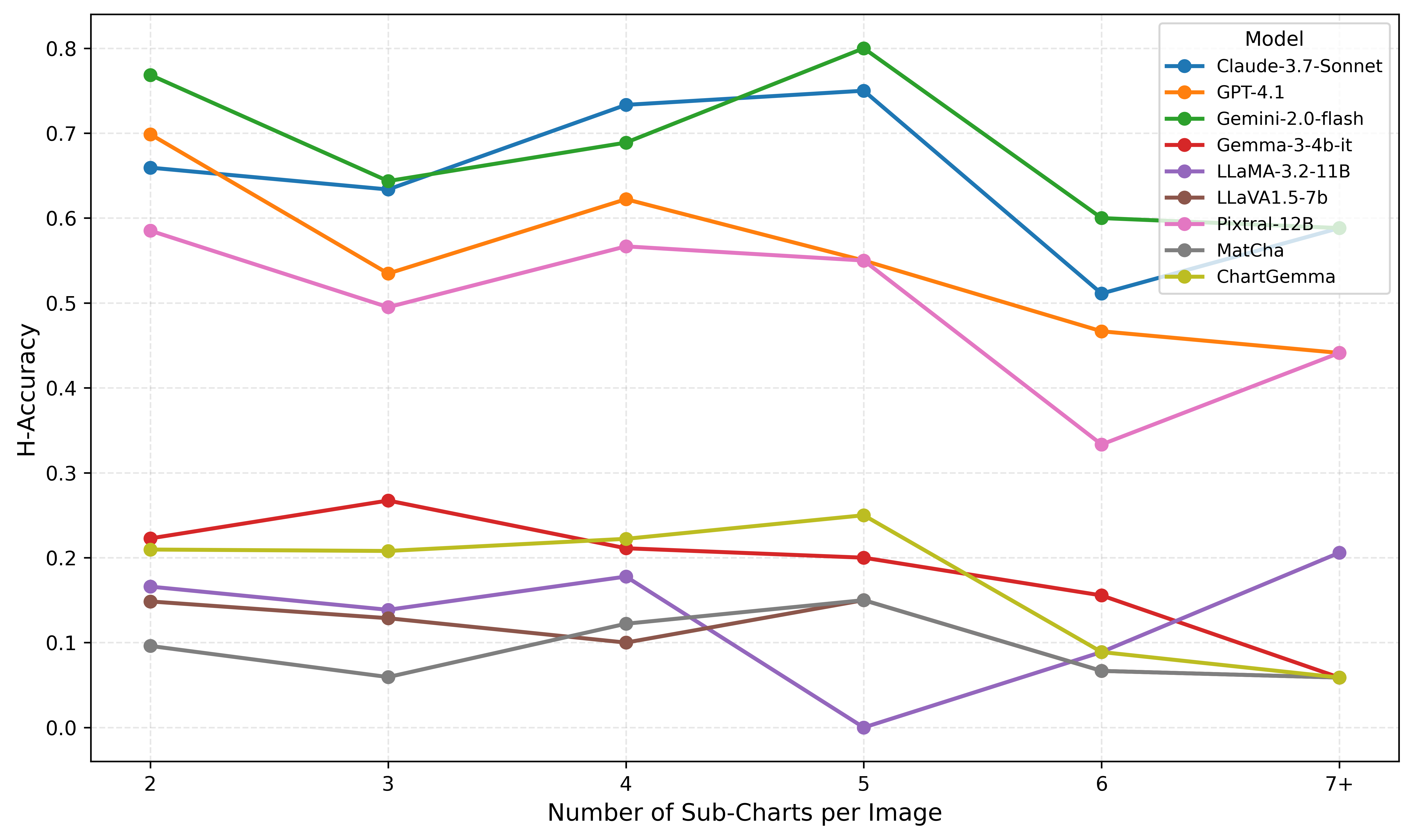}
\caption{H-Accuracy on Human-authored PolyChartQA under Zero-shot setting based on the number of sub-charts in an image.}
\label{H_acc_sub_zero_hum}
\end{figure*}

\begin{figure*}[]
\centering
\includegraphics[width=\textwidth]
{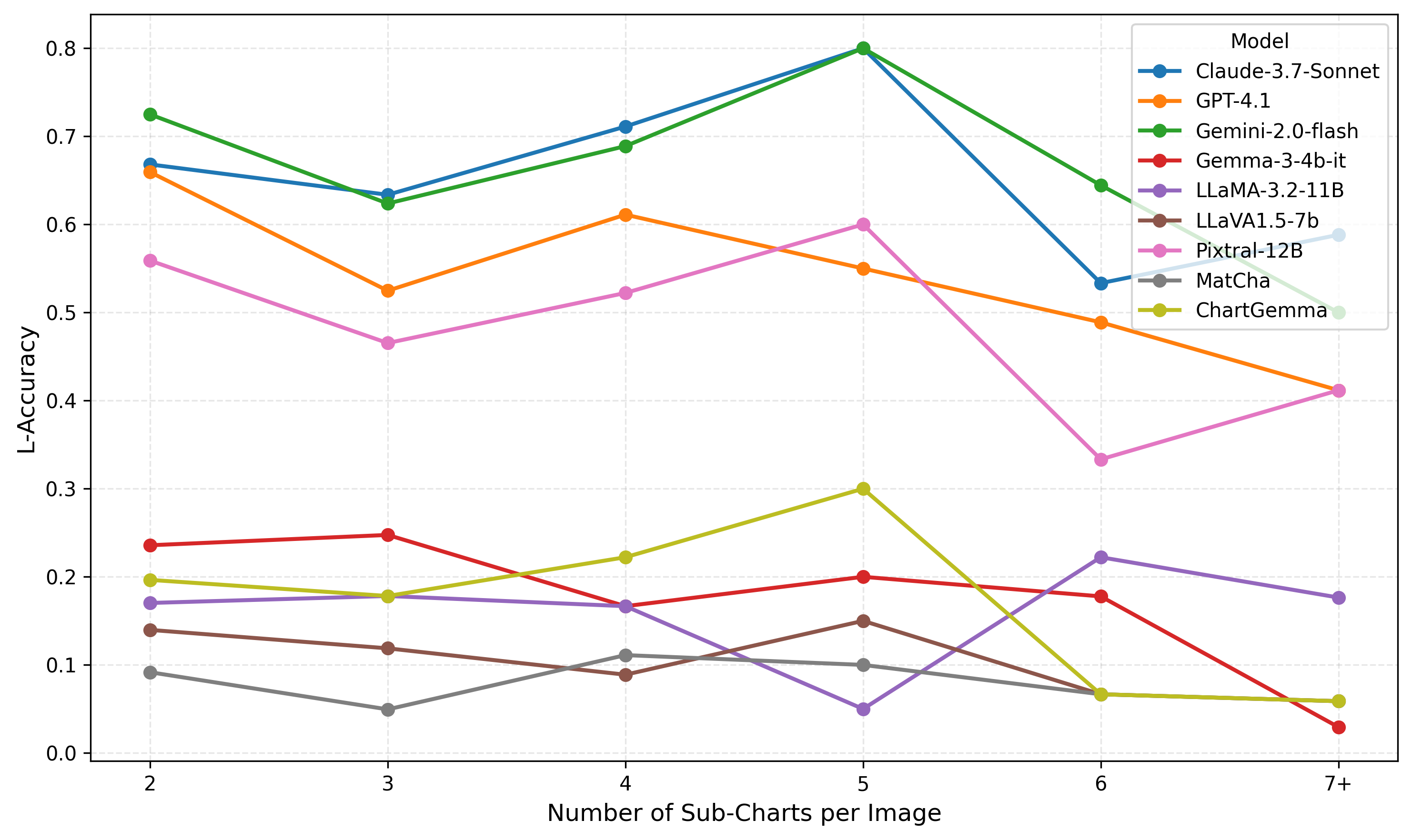}
\caption{L-Accuracy on Human-authored PolyChartQA under Zero-shot setting based on the number of sub-charts in an image.}
\label{l_acc_sub_zero_hum}
\end{figure*}

\begin{figure*}[]
\centering
\includegraphics[width=\textwidth]
{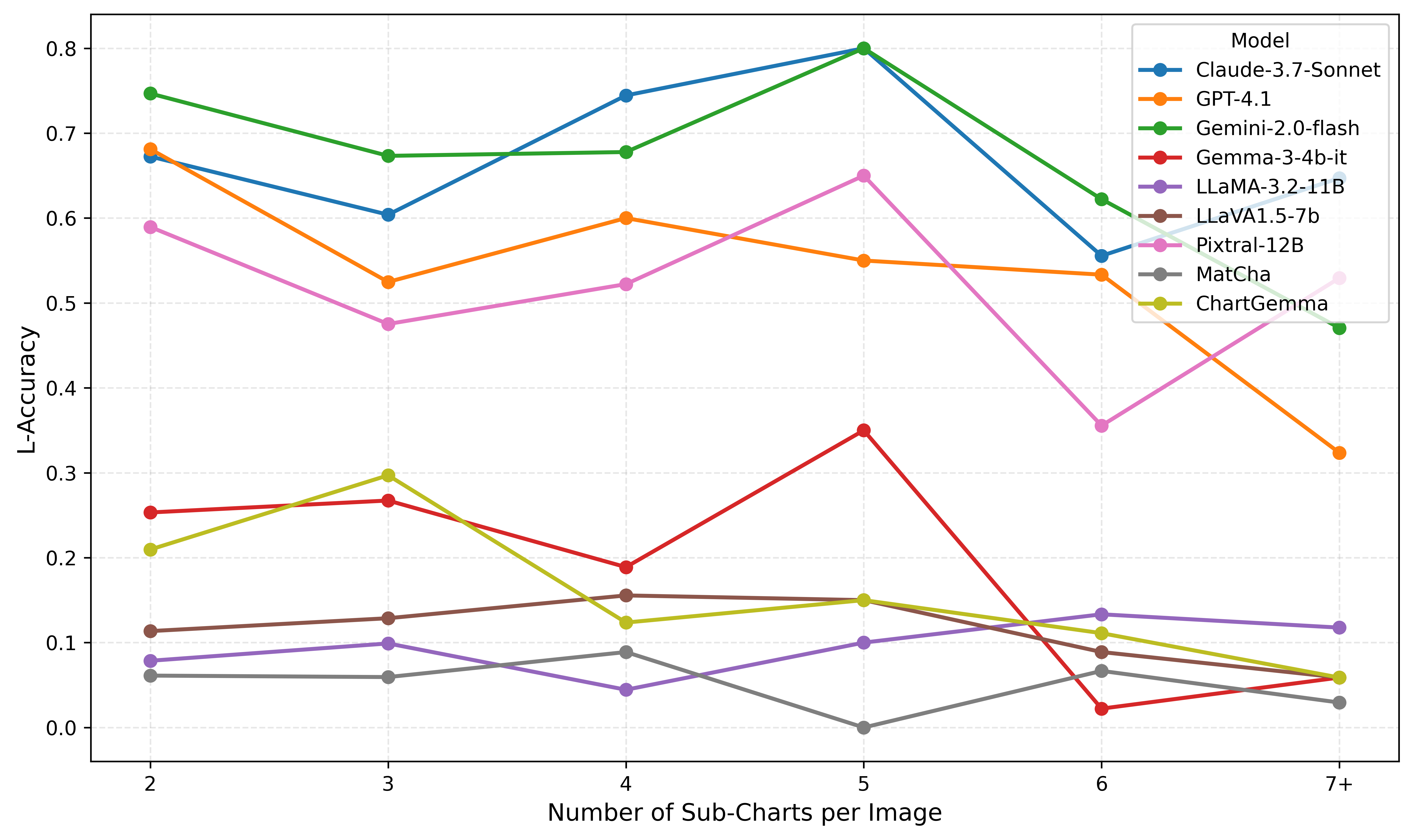}
\caption{L-Accuracy on Human-authored PolyChartQA under CoT setting based on the number of sub-charts in an image.}
\label{l_acc_sub_cot_hum}
\end{figure*}

\begin{figure*}[]
\centering
\includegraphics[width=\textwidth]
{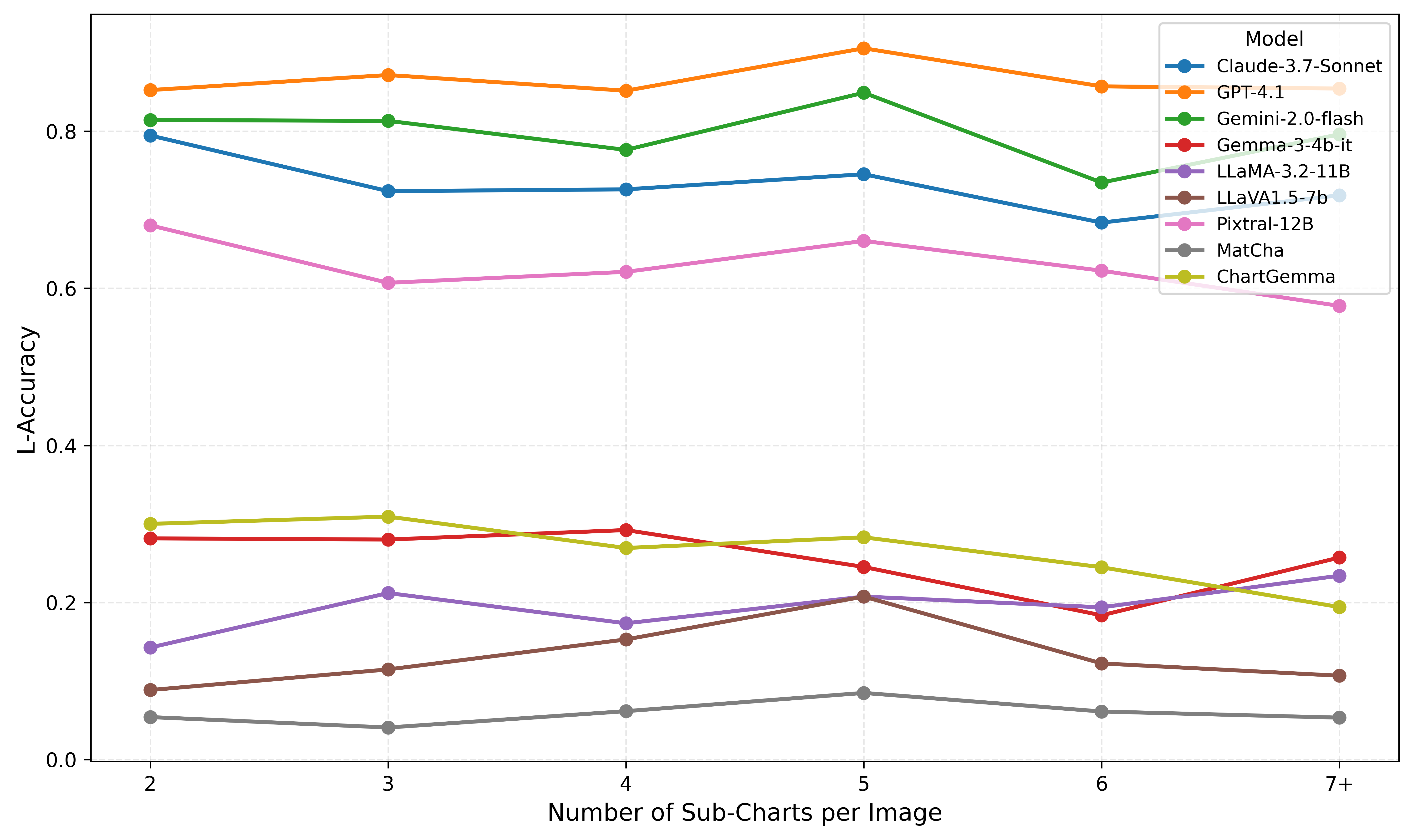}
\caption{L-Accuracy on MLM-generated PolyChartQA under Zero-shot setting based on the number of sub-charts in an image.}
\label{l_acc_sub_zero_mlm}
\end{figure*}

\begin{figure*}[]
\centering
\includegraphics[width=\textwidth]
{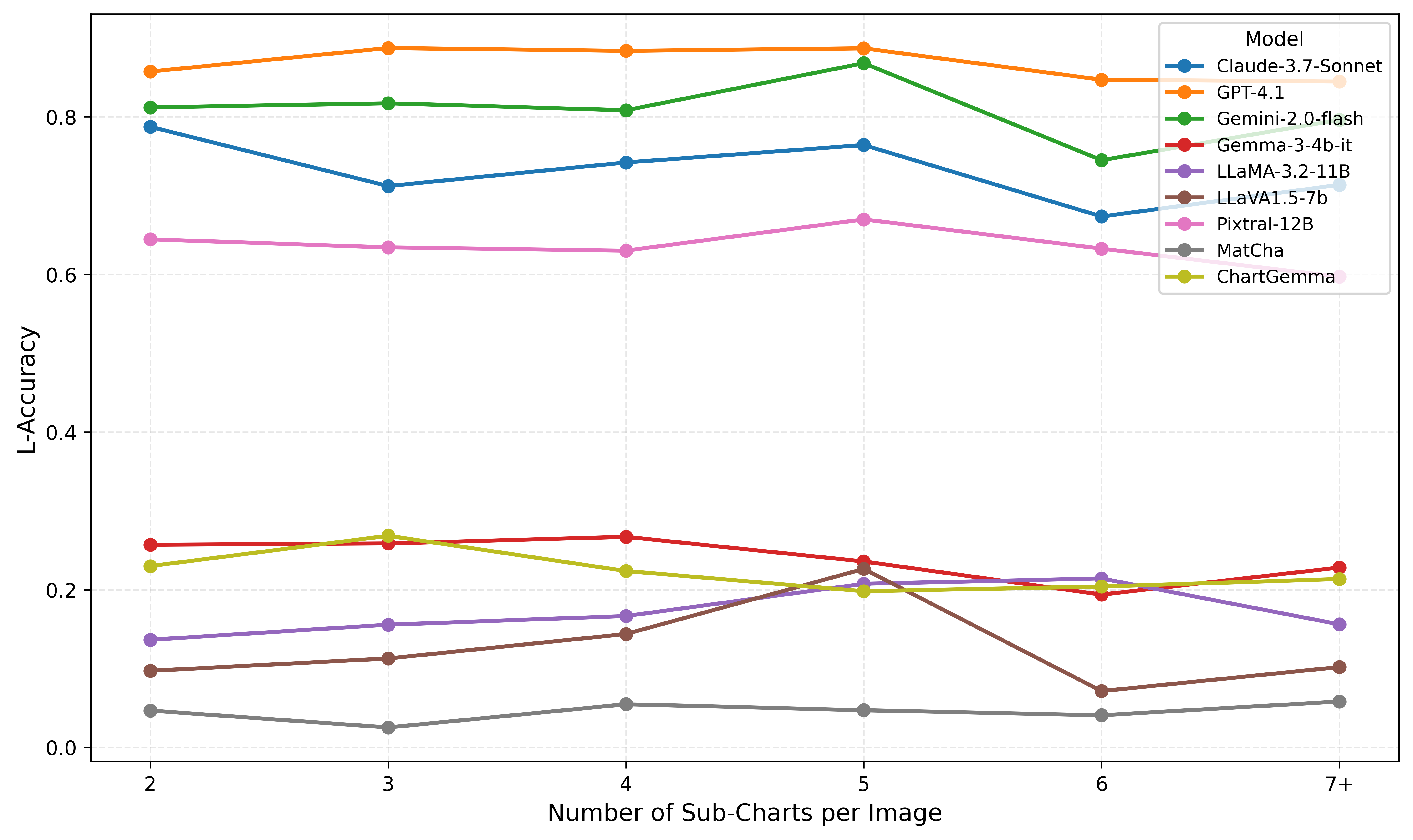}
\caption{L-Accuracy on MLM-generated PolyChartQA under CoT setting based on the number of sub-charts in an image.}
\label{l_acc_sub_cot_mlm}
\end{figure*}

\subsection{Results on Human-authored vs MLM-generated PolyChartQA\label{appendix_results_poly_3}}
%\ref{l_acc_3}-
Figure \ref{bert_acc_3} shows BERTScore for both Zero-shot and CoT settings on human-authored and MLM-generated questions from PolyChartQA.

% \begin{figure*}[ht]
% \centering
% \includegraphics[width=\textwidth]
% {l_accuracy_zero_3.png}
% \caption{L-Accuracy on Human-authored vs MLM-generated PolyChartQA under zero-shot and CoT setting.}
% \label{l_acc_3}
% \end{figure*}

\begin{figure*}[ht]
\centering
\includegraphics[width=\textwidth]
{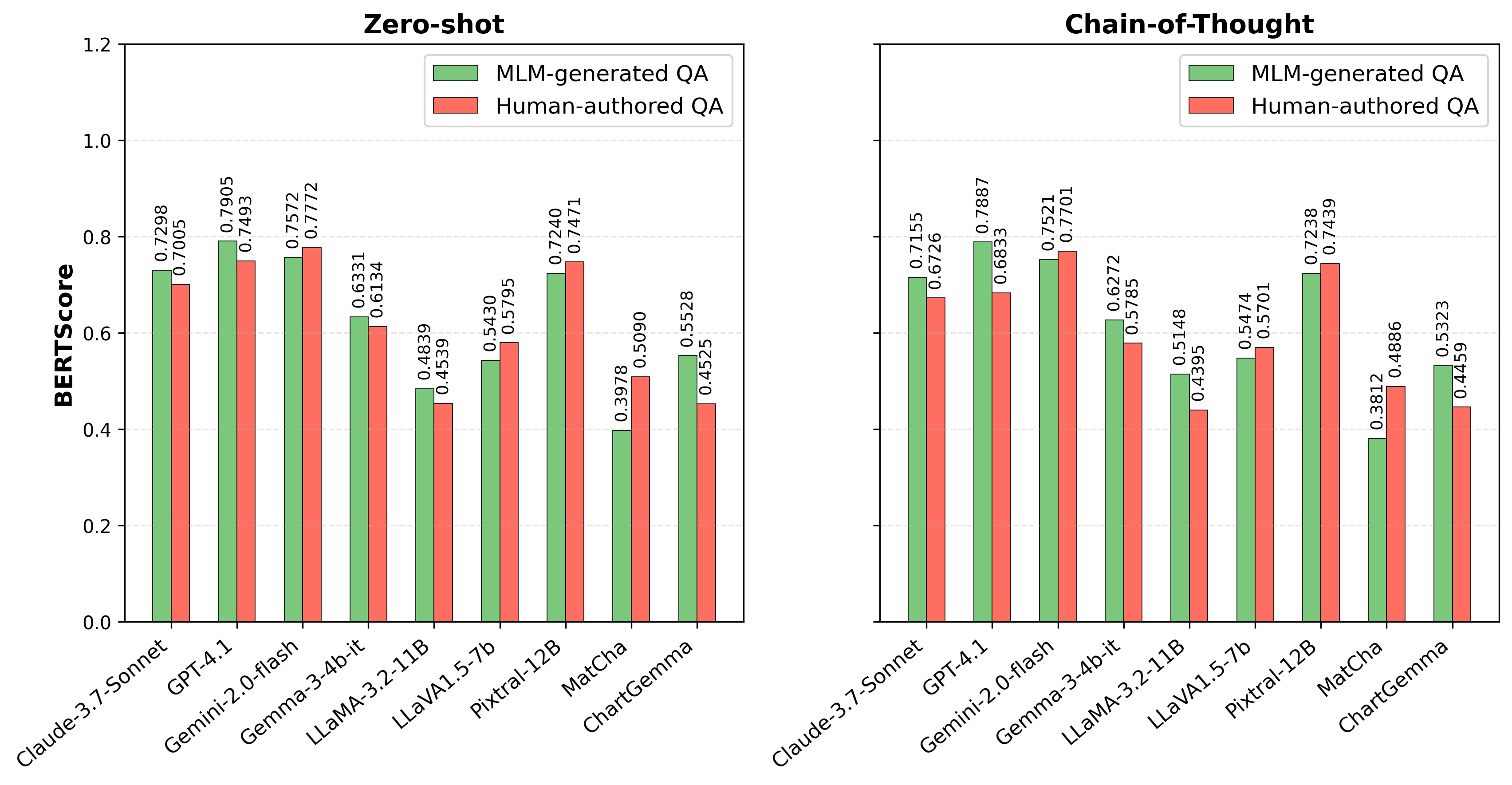}
\caption{BERTScore on the human-authored vs MLM-generated questions from PolyChartQA under Zero-shot and CoT settings.}
\label{bert_acc_3}
\end{figure*}

\subsection{Performance Comparison of Zero-shot, CoT, and VDSP Prompting Strategies}\label{appendix_results_poly_4}
Table \ref{bert_score_rq4} presents the BERTScore of Zero-shot, CoT, and VDSP on the human-authored questions.
BERTScore decreases across all models for VDSP prompting, likely due to the longer, reasoning-heavy responses. 
\begin{table}[hbpt]
\centering
\resizebox{\columnwidth}{!}{%
\begin{tabular}{lccc}
\hline
\textbf{Model} & \textbf{Zero-shot} & \textbf{CoT} & \textbf{VDSP} \\
\hline
Claude-3.7-Sonnet    & 0.7298 & 0.6726 & 0.5046 \\
GPT-4.1              &\textbf{\underline{0.7905}}  & 0.6833 &0.6033  \\
Gemini-2.0-flash &  0.7572   &0.\underline{7701}  & \underline{0.6812} \\
Pixtral-12B          & 0.7240 & 0.7434 & 0.4591 \\
\hline
\end{tabular}}
\caption{BERTScore of Zero-shot, CoT, and VDSP prompting methods on human-authored QA from PolyChartQA.}\label{bert_score_rq4}
\end{table}

% \begin{figure*}[!h]
%     \centering
%     \includegraphics[width=\textwidth]{diff.png}
%     \caption{
%         Performance comparison of models across difficulty levels on the human-authored and the MLM-generated questions from PolyChartQA. The bar charts represent L-Accuracy, while the overlaid line plots represent BERTScore. H = Human-authored, M = MLM-generated.
%     }
%     \label{diff}
% \end{figure*}

% \begin{figure*}[!ht]
%     \centering
%     \includegraphics[width=\textwidth]{q_type.png}
%     \caption{
%         Performance comparison of models across question types on Human-authored and MLM-generated questions from PolyChartQA. Bar charts represent L-Accuracy, while overlaid line plots represent BERTScore. H = Human-authored, M = MLM-generated, S = Structural, D = Data Retrieval, R = Reasoning.
%     }
%     \label{q_type_img}
% \end{figure*}

% \begin{figure*}[!ht]
%     \centering
%     \includegraphics[width=\textwidth]{homog.png}
%     \caption{
%         Performance comparison of models on homogeneous vs. non-homogeneous multi-chart images across Human-authored and MLM-generated questions from PolyChartQA. Bar charts represent L-Accuracy, while overlaid line plots represent BERTScore. H-H = Human-authored Question and homogeneous chart, H-NH = Human-authored Question and non-homogeneous chart, M-H = MLM-generated Question and homogeneous chart, M-NH = MLM-generated Question and non-homogeneous chart.  
%     }
%     \label{homog}
% \end{figure*}
\subsection{Performance Comparison of MLMs on Human-authored PolyChartQA with Human Evaluation}\label{appendix_results_h-acc_H_poly}
Table \ref{tab:ha_la_accuracy_cot} presents H-Accuracy vs L-Accuracy on human-authored questions from PolyChartQA under the CoT setting.
% \begin{table}[hbpt]
% \centering
% \begin{tabular}{l c c}
% \hline
% \textbf{Model} & \textbf{H-Accuracy} \\
% \hline
% Claude-3.7-Sonnet & 0.6532 \\
% GPT-4.1 & 0.6108 \\
% Gemini-2.0-flash & \textbf{0.7052} \\
% Pixtral-12B & 0.5318 \\
% LLaMA-3.2-11B & 0.1522 \\
% LLaVA-1.5-7B & 0.1233 \\
% Gemma-3-4b-it & 0.2119 \\
% MatCha & 0.0906 \\
% ChartGemma & 0.1927 \\
% \hline
% \end{tabular}
% \caption{H-Accuracy on human-authored PolyChartQA under Zero-shot settings.\label{tab:hacc_human_zeroshot}}
% \end{table}

\begin{table}[hbpt]
\centering
\resizebox{\columnwidth}{!}{%
\begin{tabular}{lccc}
\hline
\textbf{Model} & \textbf{H-Accuracy} & \textbf{L-Accuracy} & \textbf{Difference} \\
\hline
Claude-3.7-Sonnet      & 0.6609 & 0.6647 & 0.0038 \\
GPT-4.1                & 0.5954 & 0.5954 & 0.0000 \\
Gemini-2.0-flash       & \textbf{0.7187} & \textbf{0.6936} & 0.0251 \\
Pixtral-12B            & 0.5376 & 0.5337 & 0.0039 \\
LLaMA-3.2-11B-Vision   & 0.0809 & 0.0848 & 0.0039 \\
LLaVA1.5-7b            & 0.1175 & 0.1195 & 0.0020 \\
Gemma-3-4b-it          & 0.2274 & 0.2158 & 0.0116 \\
MatCha                 & 0.0617 & 0.0617 & 0.0000 \\
ChartGemma             & 0.2351 & 0.1911 & \textbf{0.0440} \\
\hline
\end{tabular}}
\caption{H-Accuracy vs L-Accuracy using CoT prompting on the human-authored PolyChartQA.}
\label{tab:ha_la_accuracy_cot}
\end{table}

\subsection{Ablation Study: VDSP version 2}\label{appendix_VDSP_V2}
In this setting, the Stage 1 prompt was altered to convert each sub-chart into a structured table instead of a textual description, while Stages 2 and 3 remained unchanged.
Table \ref{VDSP V2}  presents the L-Accuracy Zero-shot, CoT, and VDSP version 2 on the human-authored questions. Figure \ref{vdsp_v2_s1} shows the prompt used in stage 1.

\begin{table}[hbpt]
\centering
\resizebox{\columnwidth}{!}{%
\begin{tabular}{lccc}
\hline
\textbf{Model} & \textbf{Zero-shot} & \textbf{CoT} & \textbf{VDSP v2} \\
\hline
GPT-4.1              & 0.5896& 0.5954  &0.5742  \\
Claude-3.7-Sonnet    & 0.6570 & 0.6647 & 0.6686 \\
Gemini-2.0-flash     & \underline{0.6802} &\textbf{\underline{0.6936}}  &\textbf{\underline{0.6936}} \\
Pixtral-12B          &0.5067 & 0.5337 &  0.4547\\
\hline
\end{tabular}}
\caption{L-Accuracy of Zero-shot, CoT, and VDSP version 2 prompting methods on the human-authored QA from PolyChartQA.}\label{VDSP V2}
\end{table}

\subsection{Ablation Study: VDSP version 3}\label{appendix_VDSP_V3}
In this variant, all three stages were modified, where Stage 2 followed a dual-persona design—an Analyst generating an answer and a Reviewer verifying it—while Stages 1 and 3 retained tasks similar to the original VDSP setting.
Table \ref{VDSP V3}  presents the L-Accuracy Zero-shot, CoT, and VDSP version 3 on the human-authored questions. Figure \ref{vdsp_v3_s2} shows the prompt used in stage 2.

\begin{table}[hbpt]
\centering
\resizebox{\columnwidth}{!}{%
\begin{tabular}{lccc}
\hline
\textbf{Model} & \textbf{Zero-shot} & \textbf{CoT} & \textbf{VDSP v3} \\
\hline
GPT-4.1              & 0.5896& 0.5954  &0.5934  \\
Claude-3.7-Sonnet    & 0.6570 & 0.6647 & 0.6763 \\
Gemini-2.0-flash     & \underline{0.6802} &\textbf{\underline{0.6936}}  &\underline{0.6801} \\
Pixtral-12B          &0.5067 & 0.5337 &  0.4798\\
\hline
\end{tabular}}
\caption{L-Accuracy of Zero-shot, CoT, and VDSP version 3 prompting methods on the human-authored QA from PolyChartQA.}\label{VDSP V3}
\end{table}
\subsection{Computational Overhead of VDSP}
As a multi-stage prompting strategy, VDSP requires multiple model calls, resulting in higher token consumption and inference latency than single-pass prompting. This introduces a three-way trade-off among interpretability, cost, and accuracy. By transforming a single prediction into a traceable reasoning pipeline, VDSP is particularly valuable in analytical and high-stakes settings where understanding the model’s decision process is as important as the final answer. While single-pass prompting may be preferable in latency- and cost-sensitive applications, VDSP provides structured intermediate reasoning, improving transparency and enabling error diagnosis in complex multi-chart tasks.
\subsection{Human-Authored Question-Answer Examples \label{appendix_humanQA_examples}}
Figures \ref{H_Q_E_2} - \ref{H_Q_E_5} show examples of human-authored question-answer pairs from PolyChartQA.

\begin{figure*}[]
    \centering
\includegraphics[width=\linewidth]{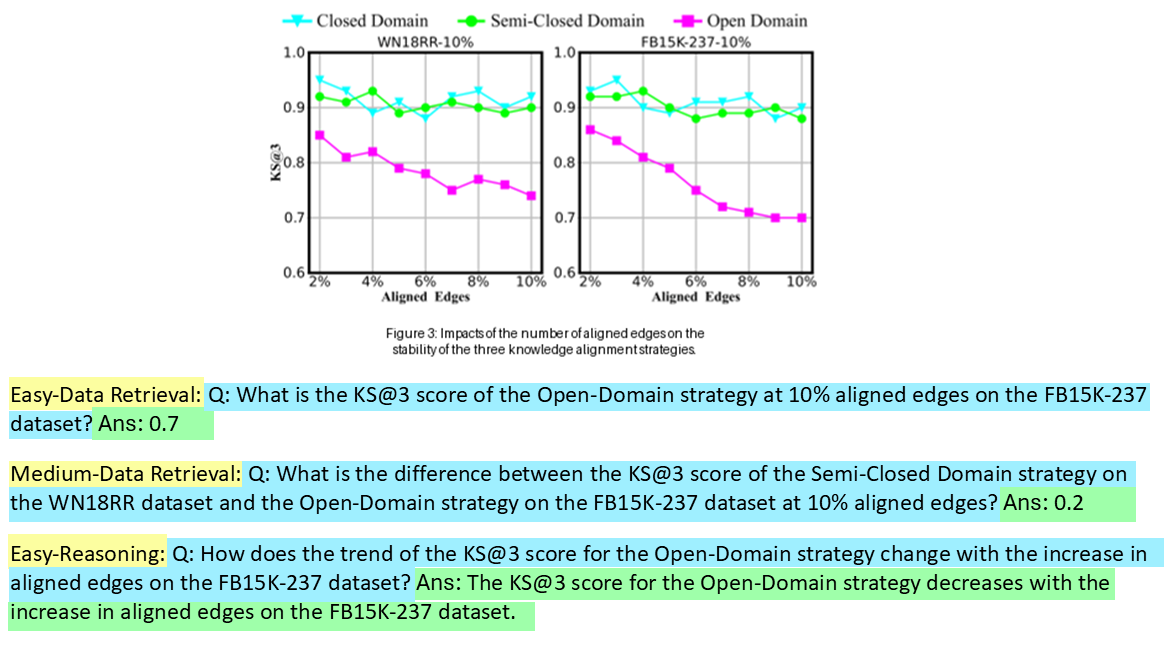}
     \caption{Example human-authored Easy-Data Retrieval, Medium-Data Retrieval, and Easy-Reasoning QA pairs from PolyChartQA. The multi-chart is collected from \citet{chen-etal-2024-new}.}
    
    \label{H_Q_E_2}
\end{figure*}
\begin{figure*}[]
    \centering
\includegraphics[width=\textwidth]{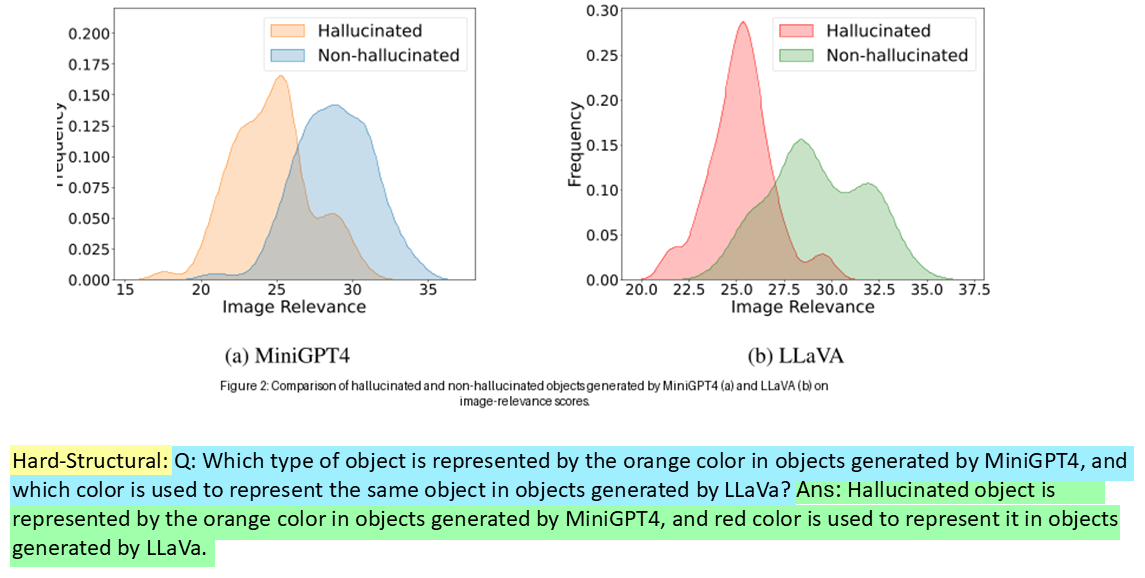}
    \caption{
        Example human-authored Hard-Structural QA pairs from PolyChartQA. The multi-chart is from \citet{xing-etal-2024-efuf}. }
    
    \label{H_Q_E_4}
\end{figure*}
\begin{figure*}[]
    \centering
\includegraphics[width=\textwidth]{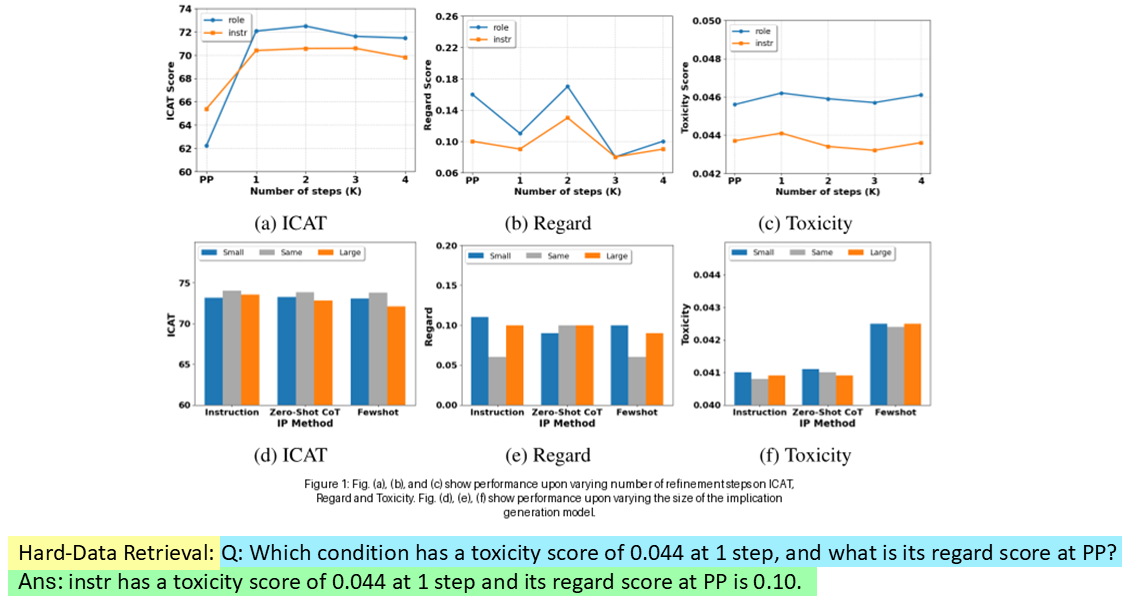}
    \caption{
        Example human-authored Hard-Data Retrieval QA pairs from PolyChartQA. The multi-chart is from \citet{furniturewala-etal-2024-thinking}.}
    
    \label{H_Q_E_5}
\end{figure*}

\subsection{MLM-generated Question-Answer Examples}\label{appendix_MLMQA_examples}
Figures \ref{L_Q_E_1}-\ref{L_Q_E_6} show examples of MLM-generated question-answer pairs from PolyChartQA.
\begin{figure*}[]
    \centering
\includegraphics[width=\textwidth]{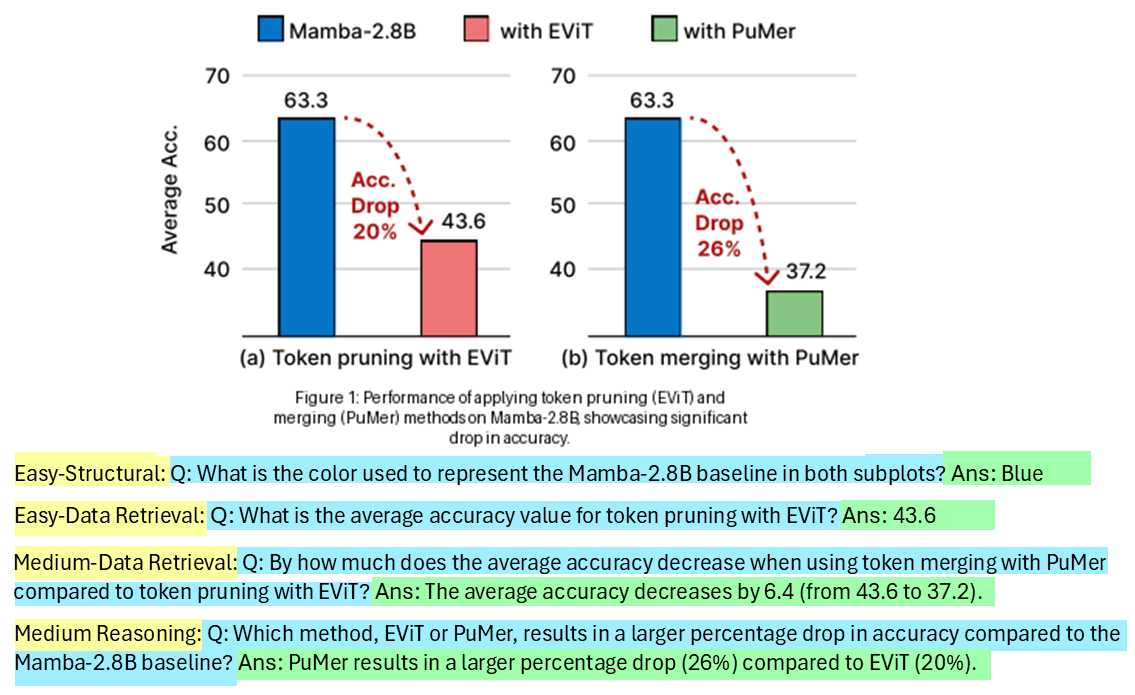}
    \caption{
        Example MLM-generated Easy-Structural, Easy-Data Retrieval, Medium-Data Retrieval, and Medium-Reasoning QA pairs from PolyChartQA. The multi-chart is from \citet{zhan-etal-2024-rethinking-token}.}
    
    \label{L_Q_E_1}
\end{figure*}
\begin{figure*}[]
    \centering
\includegraphics[width=\textwidth]{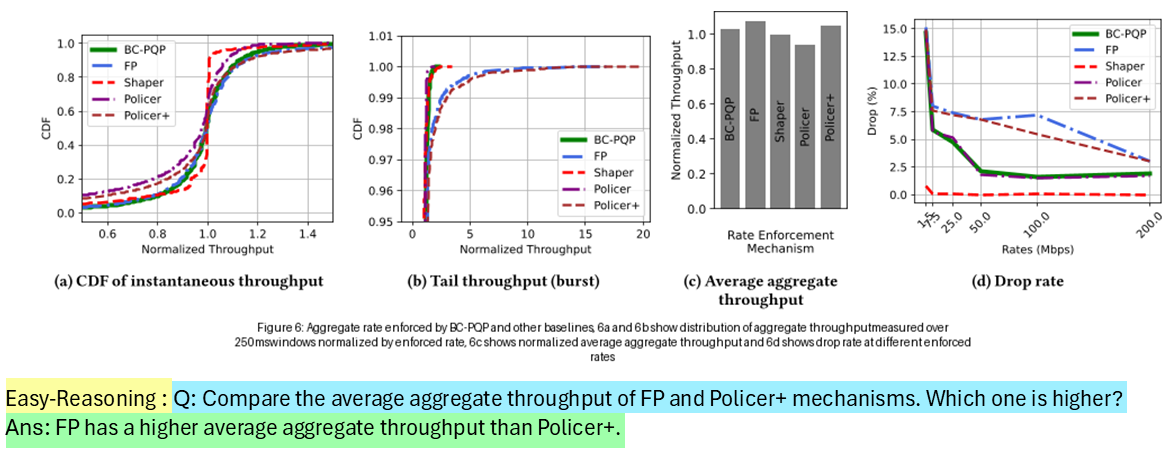}
    \caption{
        Example MLM-generated Easy-Reasoning QA pair from PolyChartQA. The multi-chart is from \citet{10.1145/3651890.3672267}.}
    
    \label{L_Q_E_2}
\end{figure*}
\begin{figure*}[]
    \centering
\includegraphics[width=\textwidth]{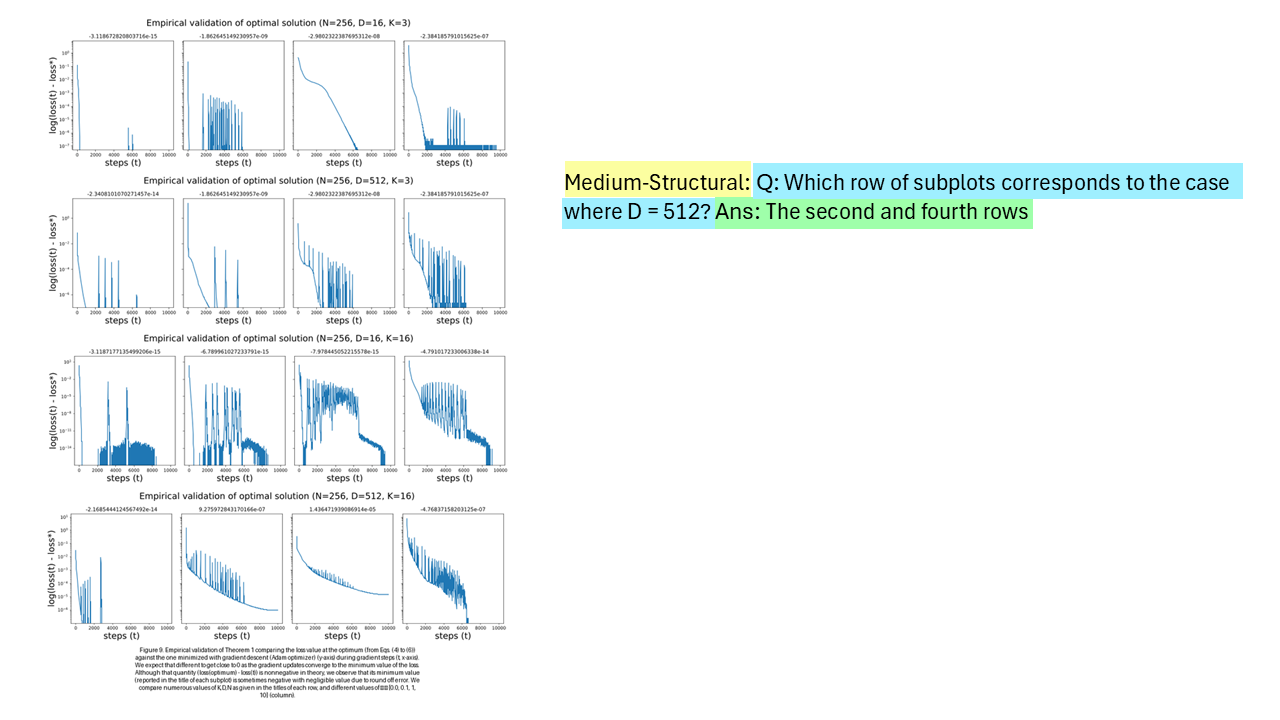}
    \caption{
        Example MLM-generated Medium-Structural QA pair from PolyChartQA. The multi-chart is from \citet{pmlr-v235-balestriero24b}.}
    
    \label{L_Q_E_3}
\end{figure*}

% \begin{figure*}[!t]
%     \centering
% \includegraphics[width=\textwidth]{L_Q_E_4.png}
%     \caption{
%         Example MLM-generated Hard-Structural QA pair from PolyChartQA. The multi-chart is from \citet{10.1145/3651890.3672222}}
    
%     \label{L_Q_E_4}
% \end{figure*}

\begin{figure*}[]
    \centering
\includegraphics[width=\textwidth]{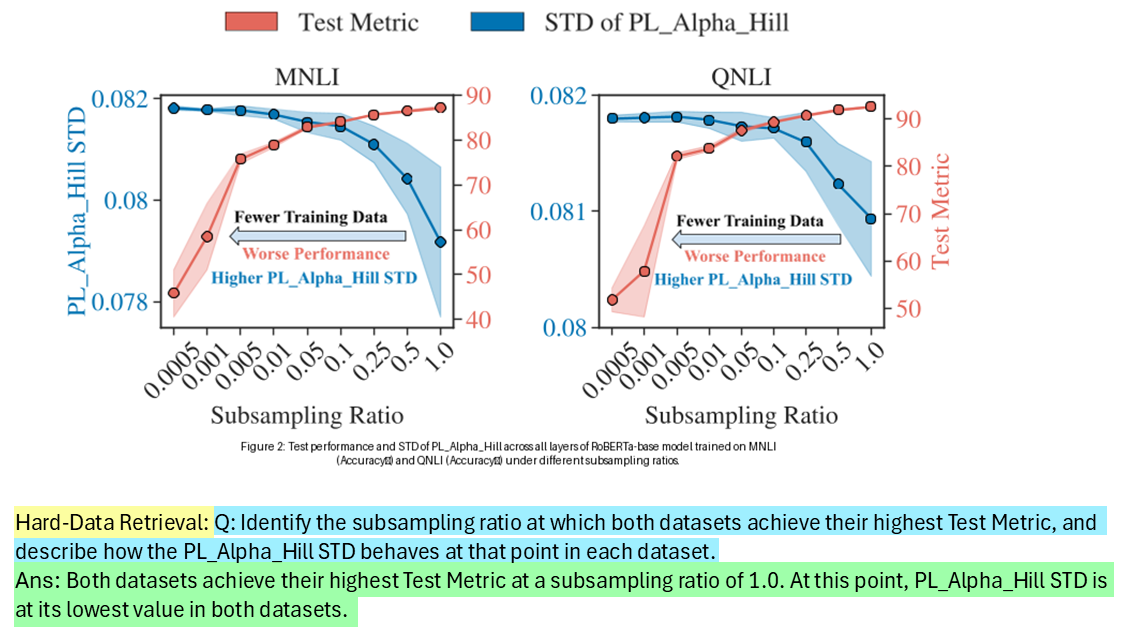}
    \caption{
        Example MLM-generated Hard-Data Retrieval QA pairs from PolyChartQA. The multi-chart is from \citet{liu-etal-2024-model}.}
    
    \label{L_Q_E_5}
\end{figure*}
\begin{figure*}[]
    \centering
\includegraphics[width=\textwidth]{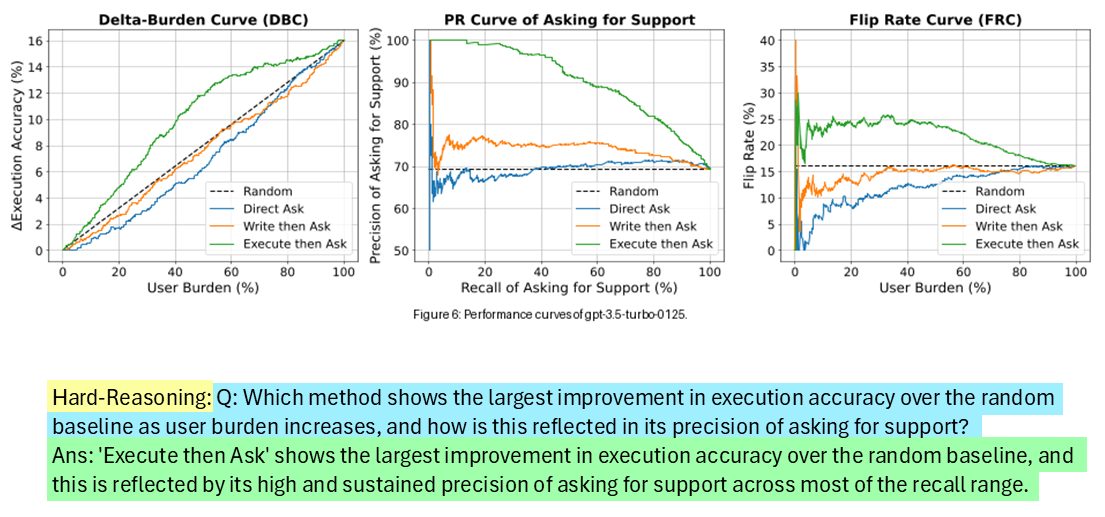}
    \caption{
        Example MLM-generated Hard-Reasoning QA pairs from PolyChartQA. The multi-chart is from \citet{wu-etal-2024-need}.}
    
    \label{L_Q_E_6}
\end{figure*}

\subsection{Prompt Examples \label{appendix_prompt_ex}}
Figures \ref{annot}-\ref{acc_prompt} show the prompts used in this experiment.
\begin{figure*}[]
    \centering
    \includegraphics[width=\textwidth]{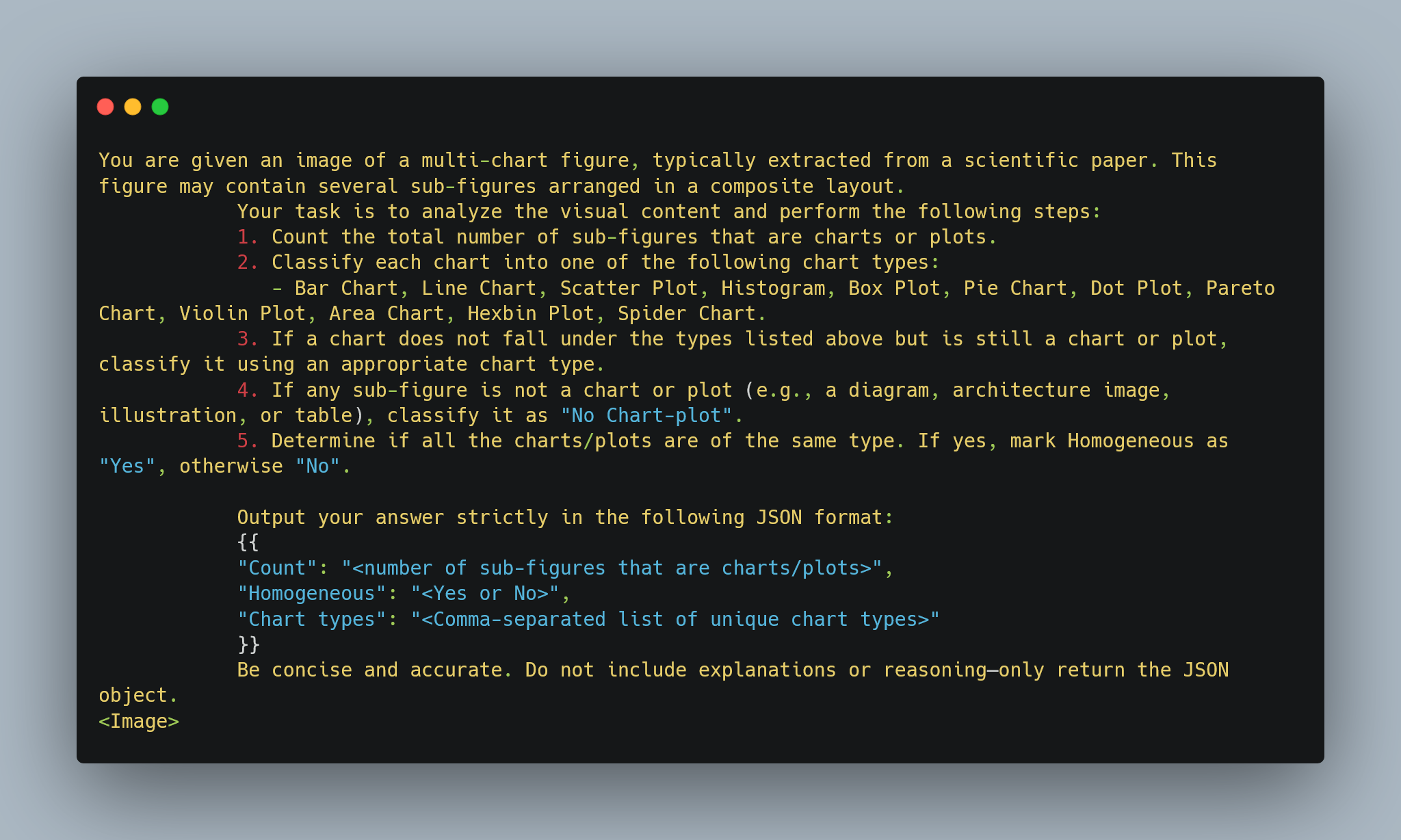}
    \caption{
        Prompt used to annotate multi-chart images.  
    }
    \label{annot}
\end{figure*}
\begin{figure*}[]
    \centering
    \includegraphics[width=0.9\textwidth]{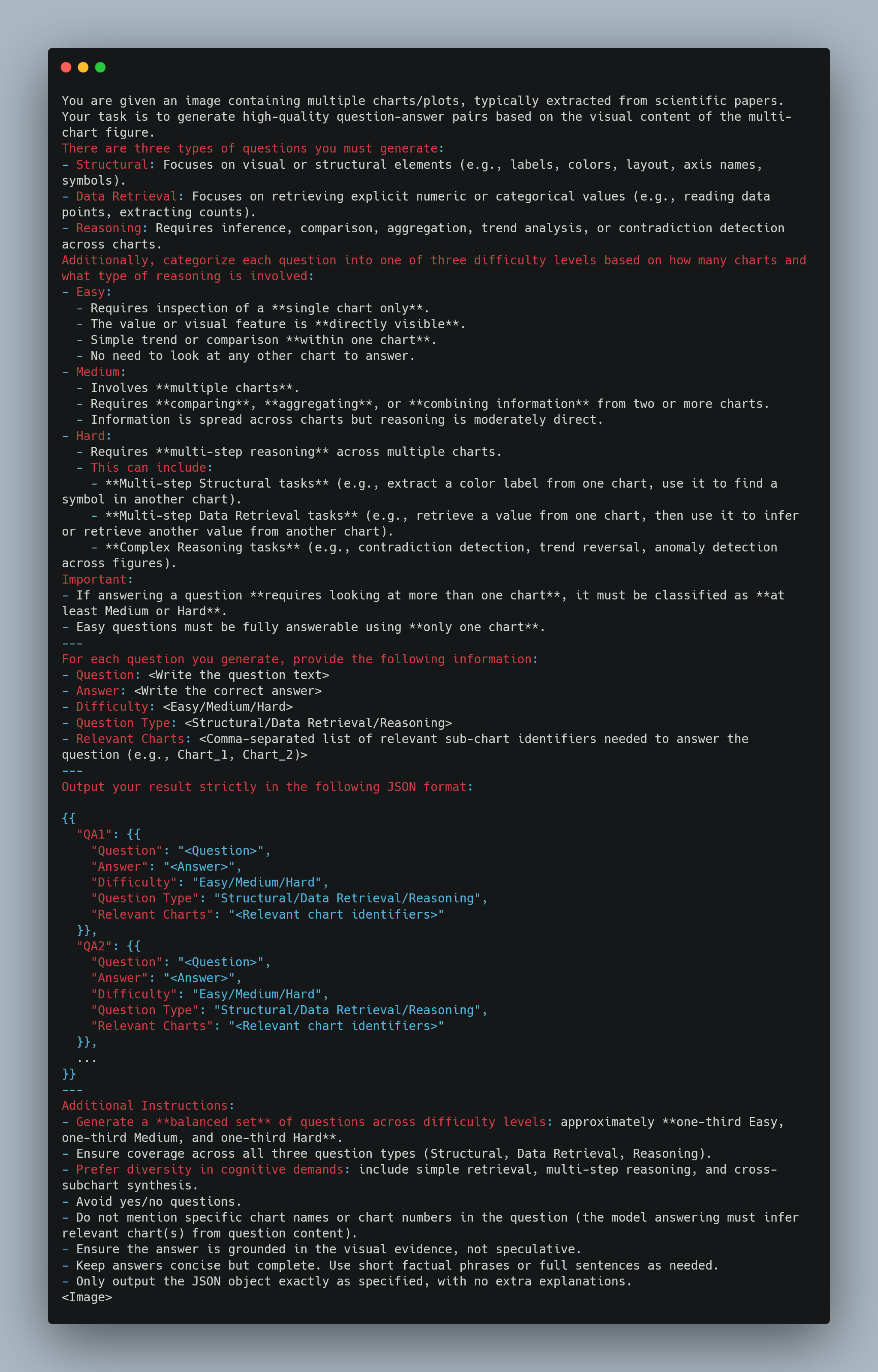}
    \caption{
        Prompt used to generate question-answers using MLM.
    }
    \label{qa_prompt}
\end{figure*}
\begin{figure*}[]
    \centering
    \includegraphics[width=\textwidth]{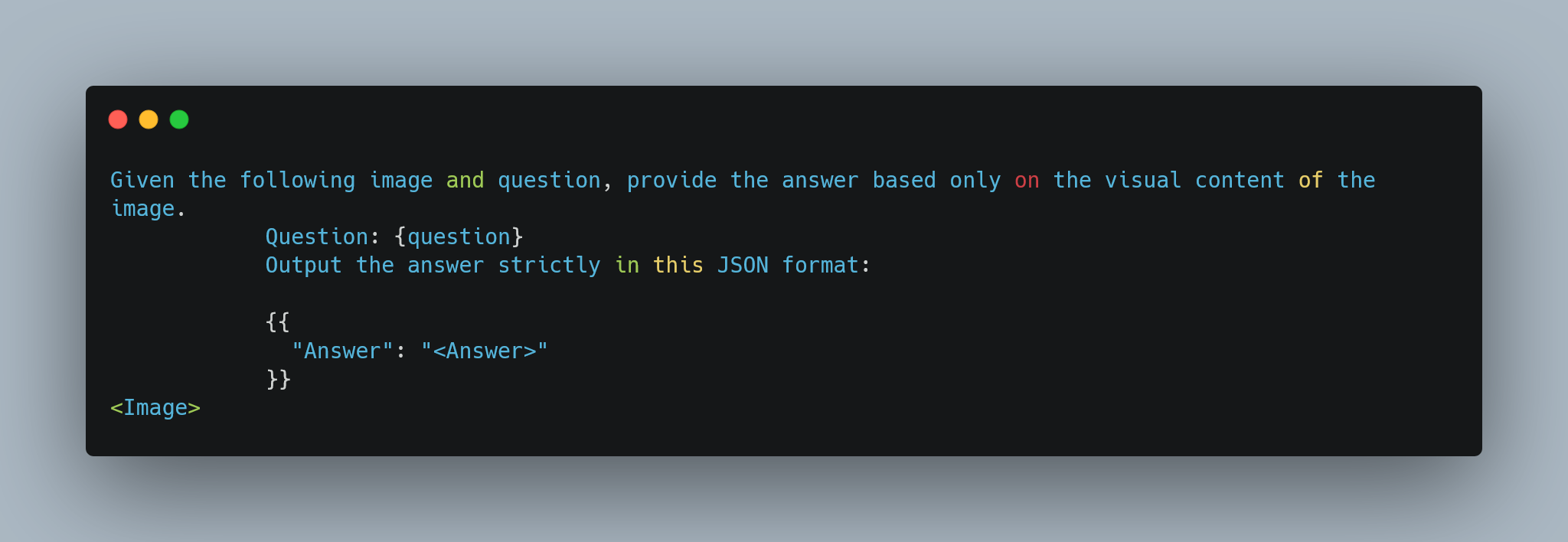}
    \caption{
        Zero-shot prompt used to evaluate MLM on multi-chart images.  
    }
    \label{zero}
\end{figure*}
\begin{figure*}[]
    \centering
    \includegraphics[width=\textwidth]{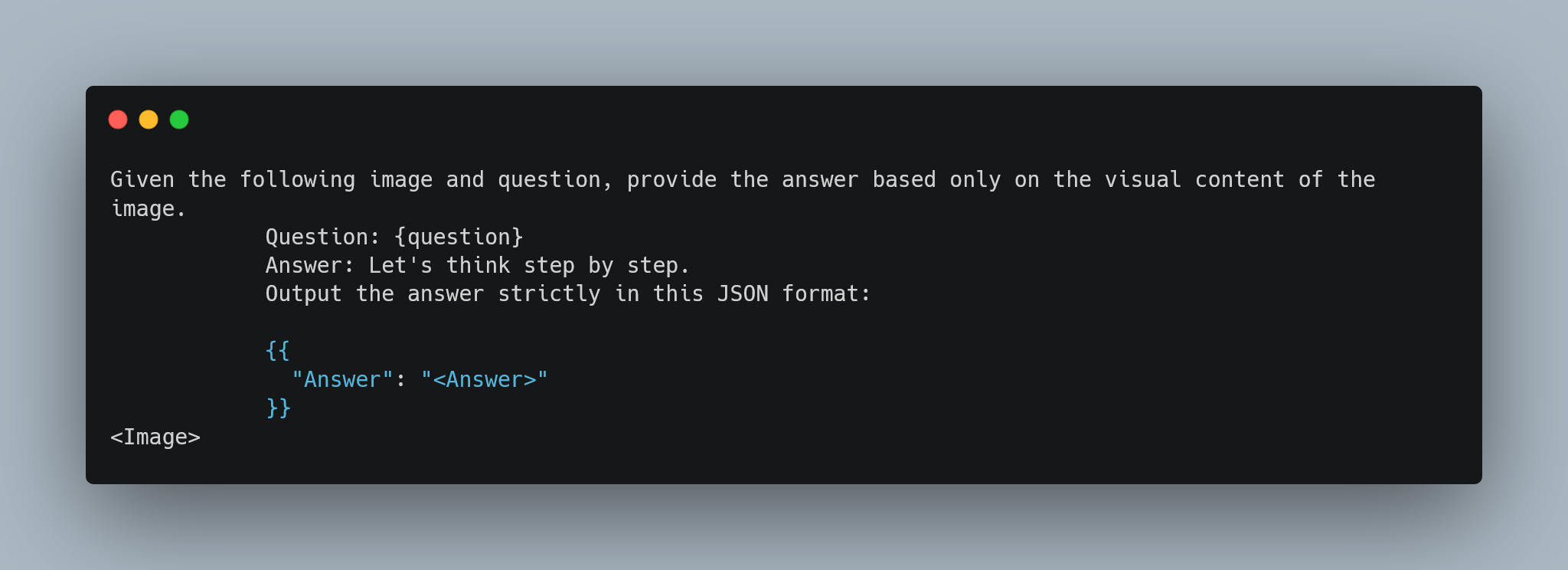}
    \caption{
        CoT prompt used to evaluate MLMs on multi-chart images.  
    }
    \label{cot}
\end{figure*}

\begin{figure*}[]
    \centering
    \includegraphics[width=\textwidth]{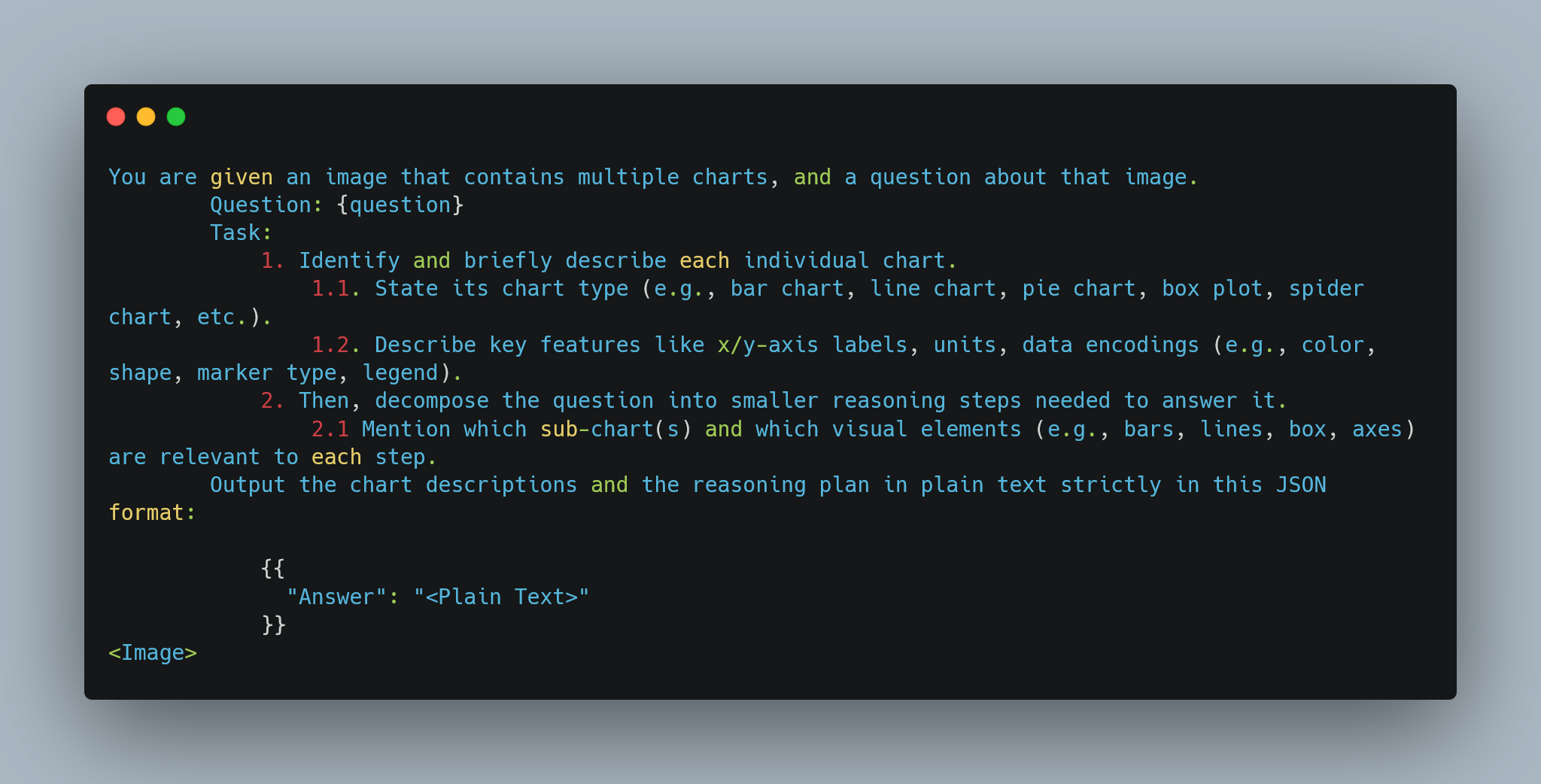}
    \caption{
       VDSP Stage 1 prompt used to evaluate MLM performance on multi-chart images.  
    }
    \label{stage1}
\end{figure*}
\begin{figure*}[]
    \centering
    \includegraphics[width=\textwidth]{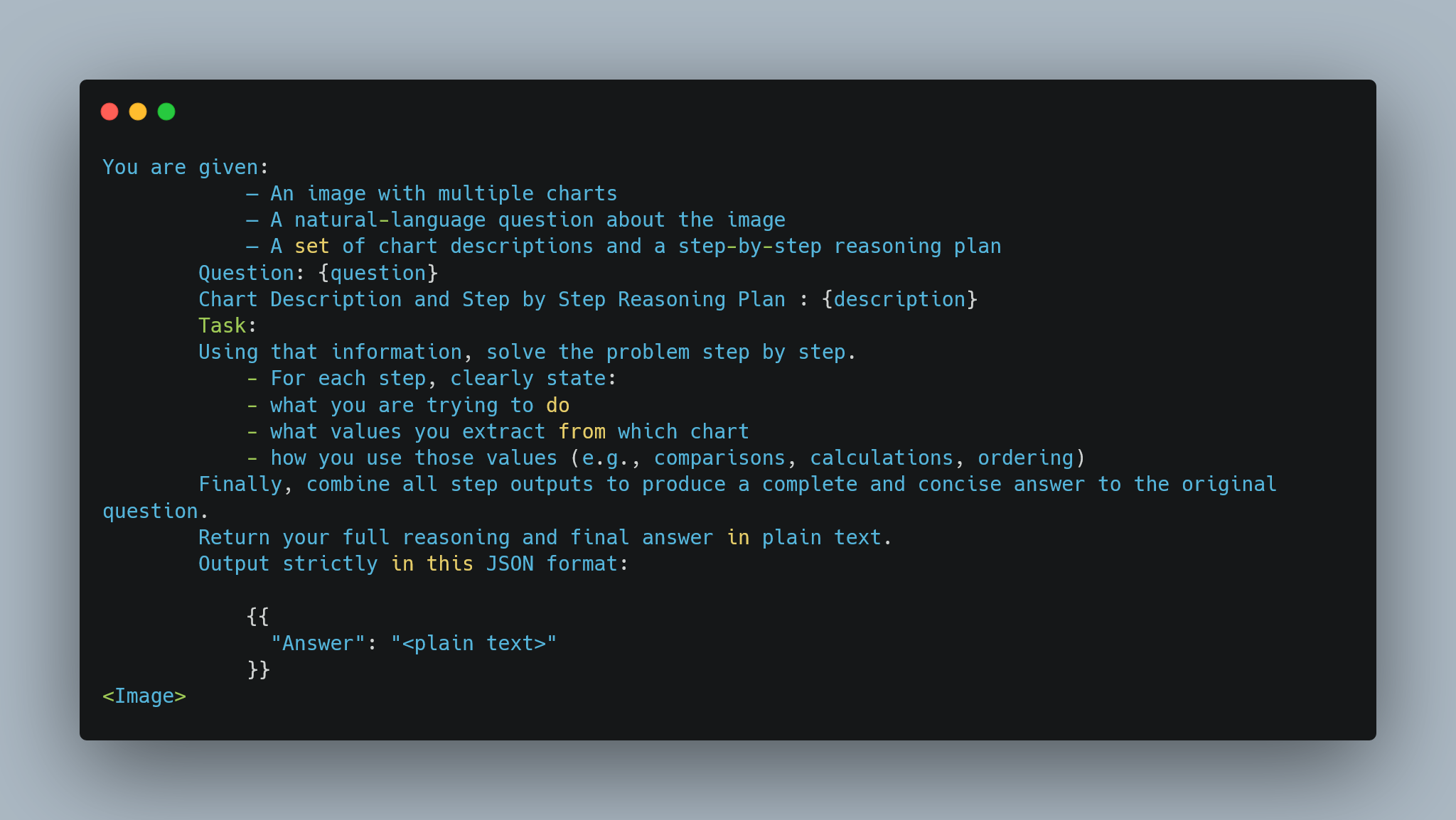}
    \caption{
        VDSP Stage 2 prompt used to evaluate MLM performance on multi-chart images.  
    }
    \label{stage_2}
\end{figure*}
\begin{figure*}[]
    \centering
    \includegraphics[width=\textwidth]{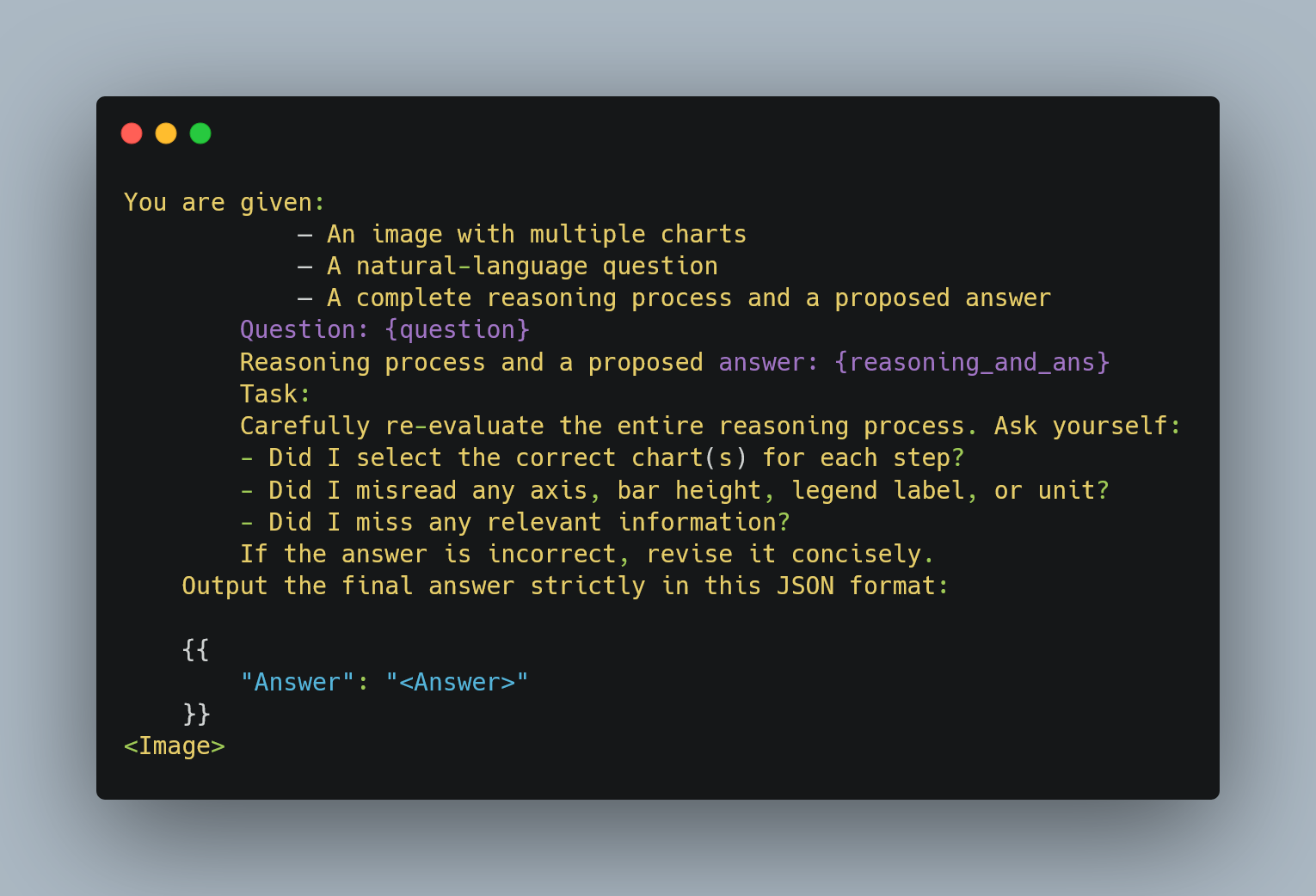}
    \caption{
        VDSP Stage 3 prompt used to evaluate MLM performance on multi-chart images.  
    }
    \label{stage3}
\end{figure*}
\begin{figure*}[]
    \centering
    \includegraphics[width=\textwidth]{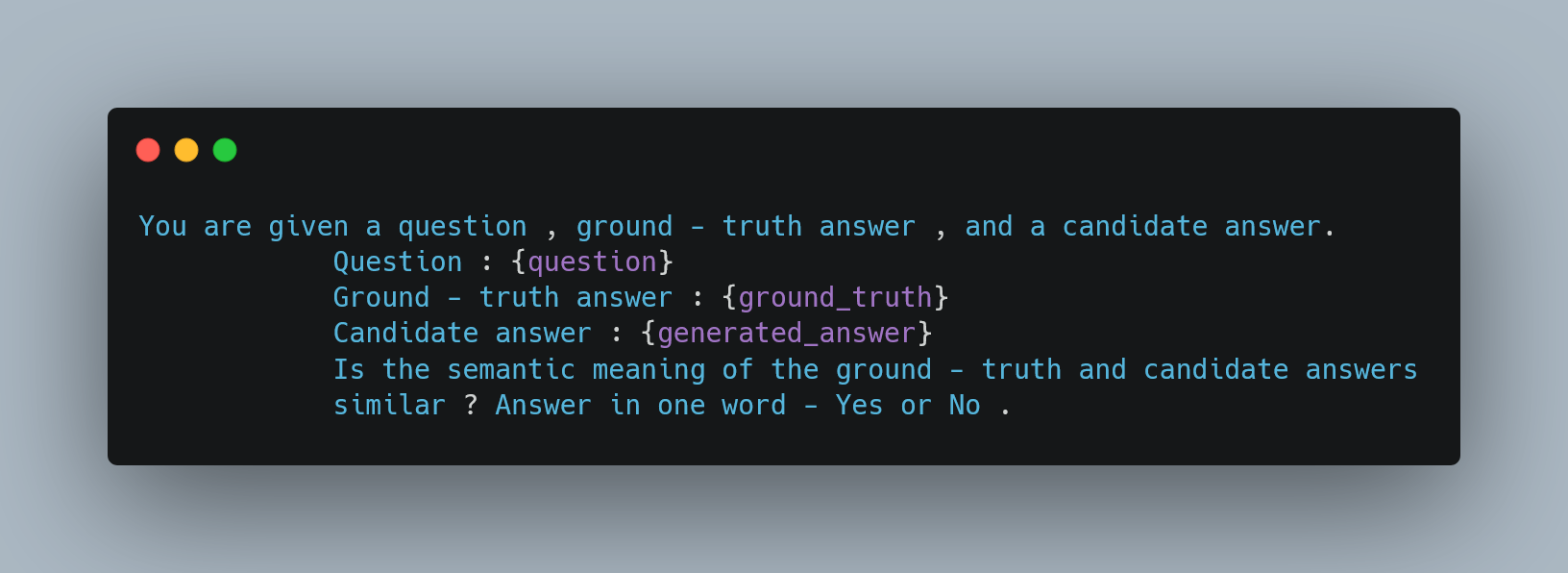}
    \caption{
        Prompt to compare the ground truth and an MLM response to derive L-Accuracy.  
    }
    \label{acc_prompt}
\end{figure*}

\begin{figure*}[]
    \centering
    \includegraphics[width=\textwidth]{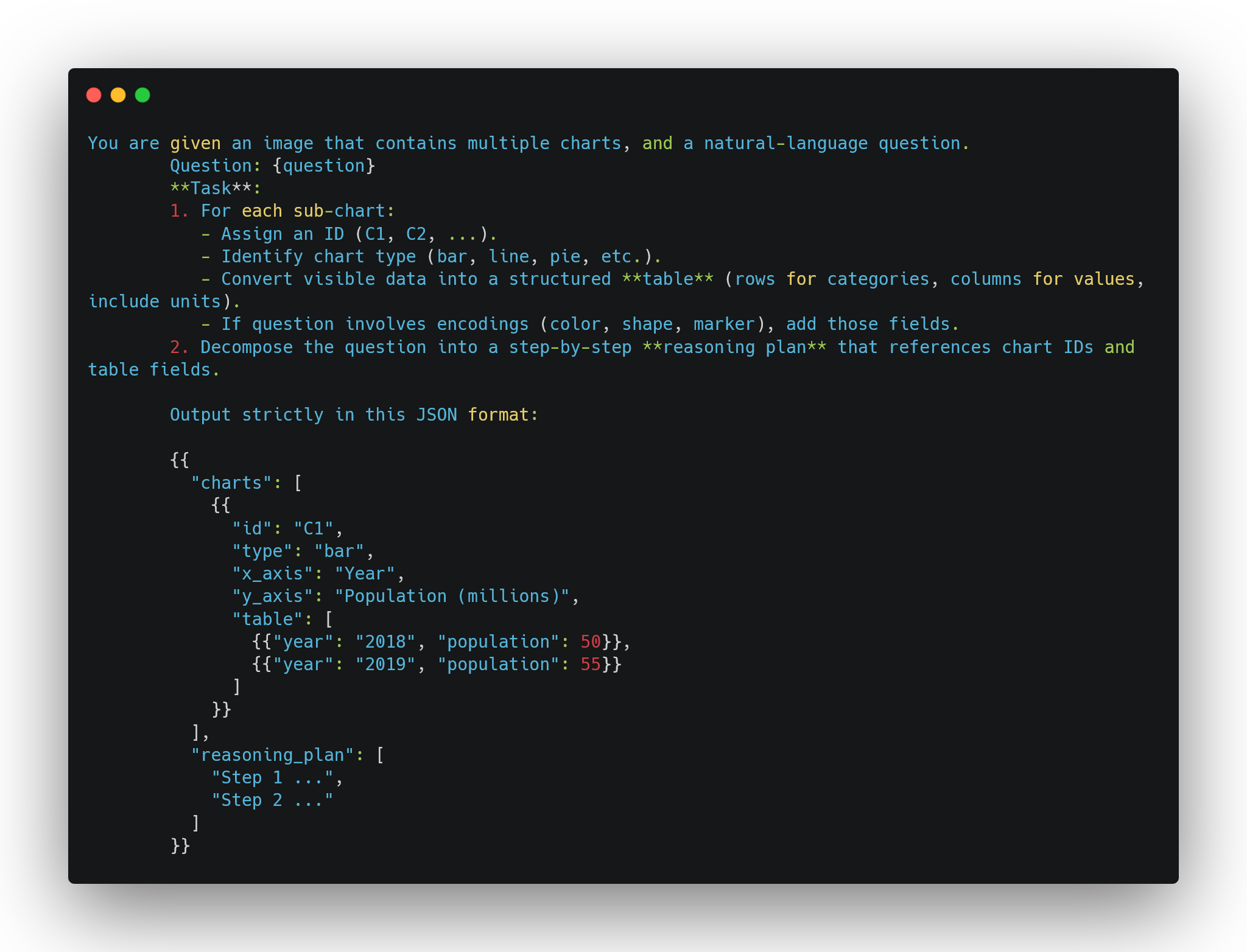}
    \caption{
       VDSP version 2 Stage 1 prompt used to evaluate MLM performance on multi-chart images.  
    }
    \label{vdsp_v2_s1}
\end{figure*}
\begin{figure*}[]
    \centering
    \includegraphics[width=\textwidth]{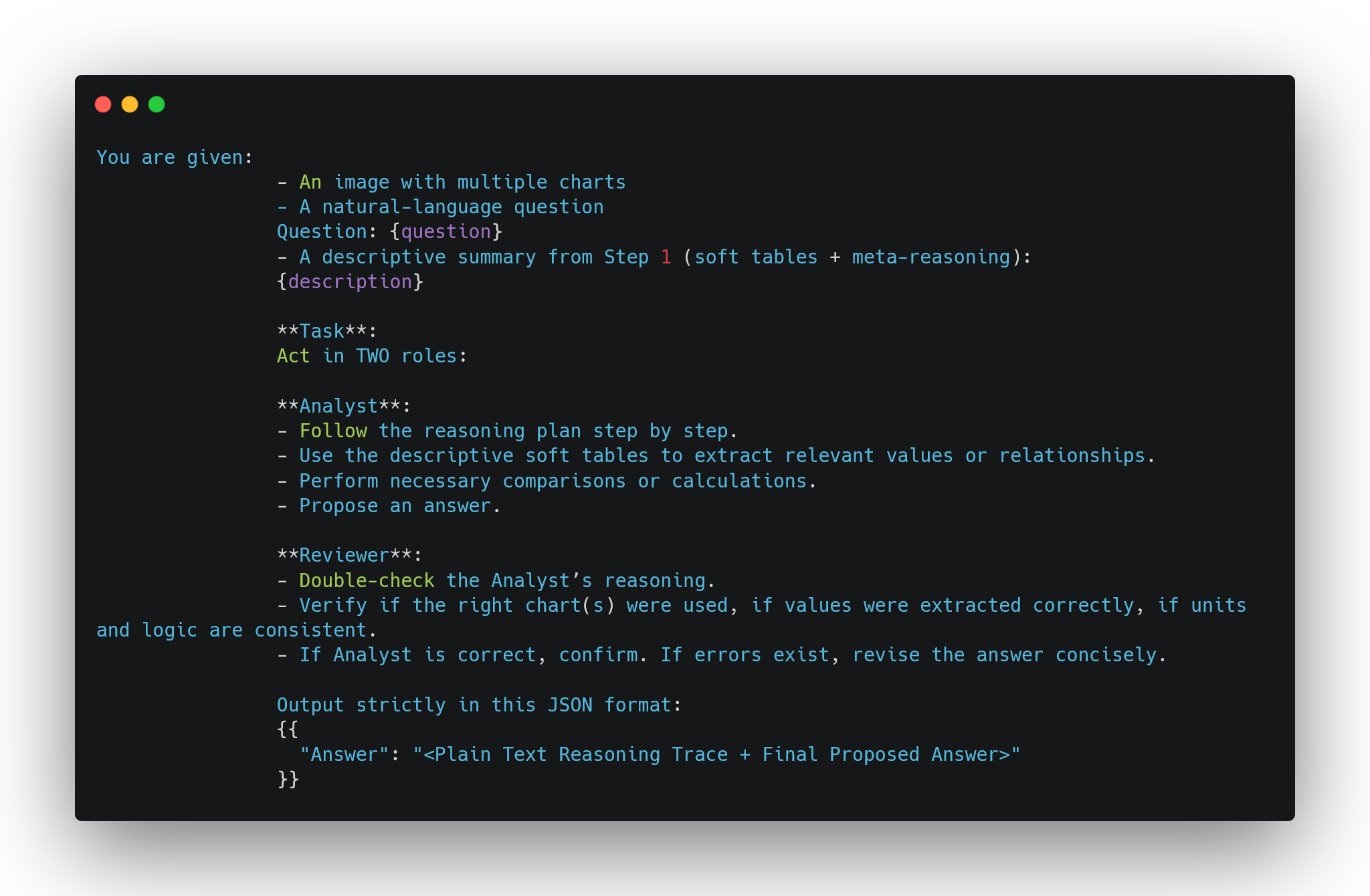}
    \caption{
       VDSP version 3 Stage 2 prompt used to evaluate MLM performance on multi-chart images.  
    }
    \label{vdsp_v3_s2}
\end{figure*}

\subsection{Example of VDSP's Intermediate Steps Output}\label{VDSP_int_steps_output}
Figures \ref{ex_step_out_vdsp}–\ref{int_step_ex_q} present the multi-chart image, the question, and the intermediate outputs along with the final answer generated by GPT-4.1 using VDSP.
\begin{figure*}[]
    \centering
\includegraphics[width=\textwidth]{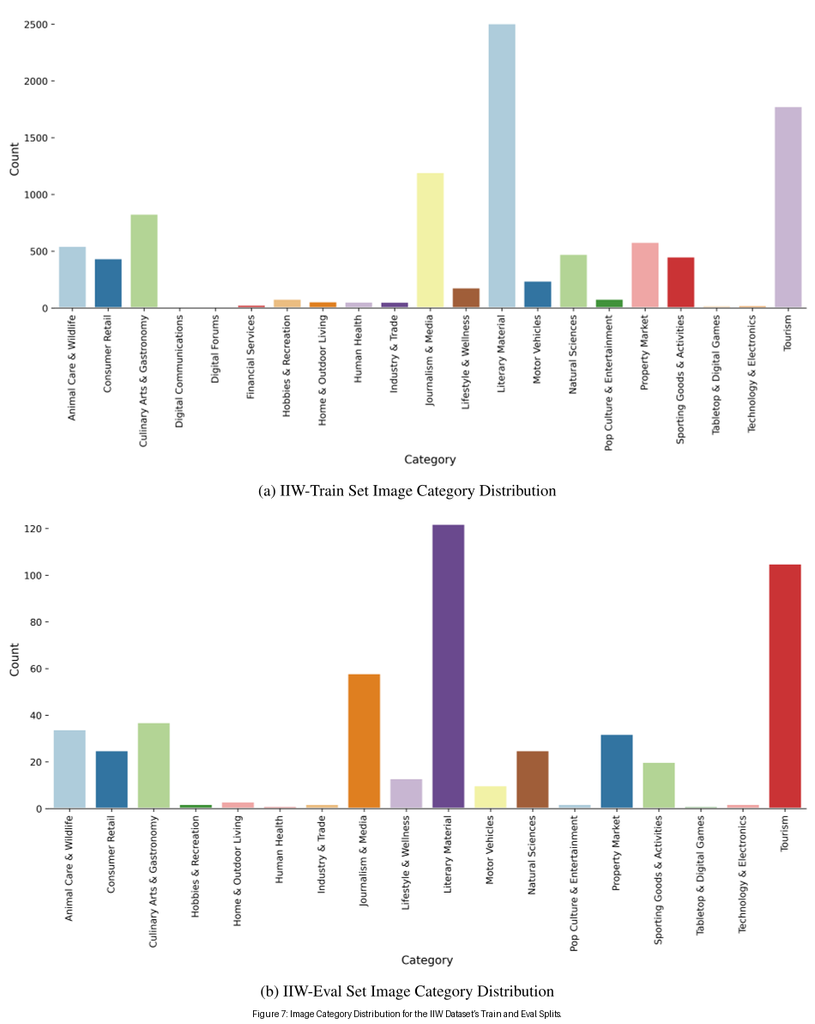}
    \caption{
        The multi-chart is from \citet{garg-etal-2024-imageinwords}.}
    \label{ex_step_out_vdsp}
\end{figure*}

\begin{figure*}[]
\centering
\begin{spverbatim}
Question: What is the count of Animal care & Wildlife in train set images?

Stage 1. Chart Structuring and Question Decomposition Output:
1. Chart Descriptions:
 1.1. The first chart (top) is a vertical bar chart titled 'IIW-Train Set Image Category Distribution'.
  - X-axis: 'Category' (various categories such as Animal Care & Wildlife, Consumer Retail, etc.)
  - Y-axis: 'Count' (number of images, unitless)
  - Data encoding: Each bar represents the count of images in a specific category. Bars are colored differently for each category. No explicit legend, but categories are labeled on the x-axis.

 1.2. The second chart (bottom) is a vertical bar chart titled 'IIW-Eval Set Image Category Distribution'.
  - X-axis: 'Category' (same categories as above)
  - Y-axis: 'Count' (number of images, unitless)
  - Data encoding: Each bar represents the count of images in a specific category. Bars are colored differently for each category. No explicit legend, but categories are labeled on the x-axis.

2. Reasoning Plan:
 2.1. Identify which chart corresponds to the 'train set' (the first/top chart labeled 'IIW-Train Set Image Category Distribution'). 
  - Relevant sub-chart: First/top bar chart.
 2.2. Locate the bar corresponding to 'Animal Care & Wildlife' on the x-axis of the train set chart.
  - Relevant visual element: The first bar from the left in the top chart.
 2.3. Read the height of this bar to determine the count value on the y-axis.
  - Relevant visual element: The height of the 'Animal Care & Wildlife' bar and its alignment with the y-axis tick marks.
 2.4. Report the count as the answer.
Summary: To answer the question, focus on the first/top bar chart, find the 'Animal Care & Wildlife' bar, and read its height to get the count.

Stage 2. Structured Reasoning Execution Output:
Step 1: Identify which chart corresponds to the 'train set'. The first (top) chart is labeled 'IIW-Train Set Image Category Distribution', so this is the relevant chart.

Step 2: Locate the bar corresponding to 'Animal Care & Wildlife' on the x-axis of the train set chart. This is the first bar from the left in the top chart.

Step 3: Read the height of this bar to determine the count value on the y-axis. The height of the 'Animal Care & Wildlife' bar aligns with the 500 mark on the y-axis.

Step 4: Report the count as the answer.
Final Answer: The count of Animal Care & Wildlife in train set images is 500.

Stage 3. Self-Verification and Final Answer Generation Output:
The count of Animal Care & Wildlife in train set images is approximately 500. The reasoning process correctly identified the relevant chart (the top one), the correct bar, and accurately read the y-axis value.
\end{spverbatim}
\caption{Question and model outputs under VDSP. The input question and the corresponding intermediate reasoning traces and final answer generated by GPT-4.1. 
}
\label{int_step_ex_q}
\end{figure*}

\subsection{Error Examples}\label{appendix_error}
Figure \ref{error_2} shows example question where all the models were wrong.

\begin{figure*}[]
    \centering
\includegraphics[width=\textwidth]{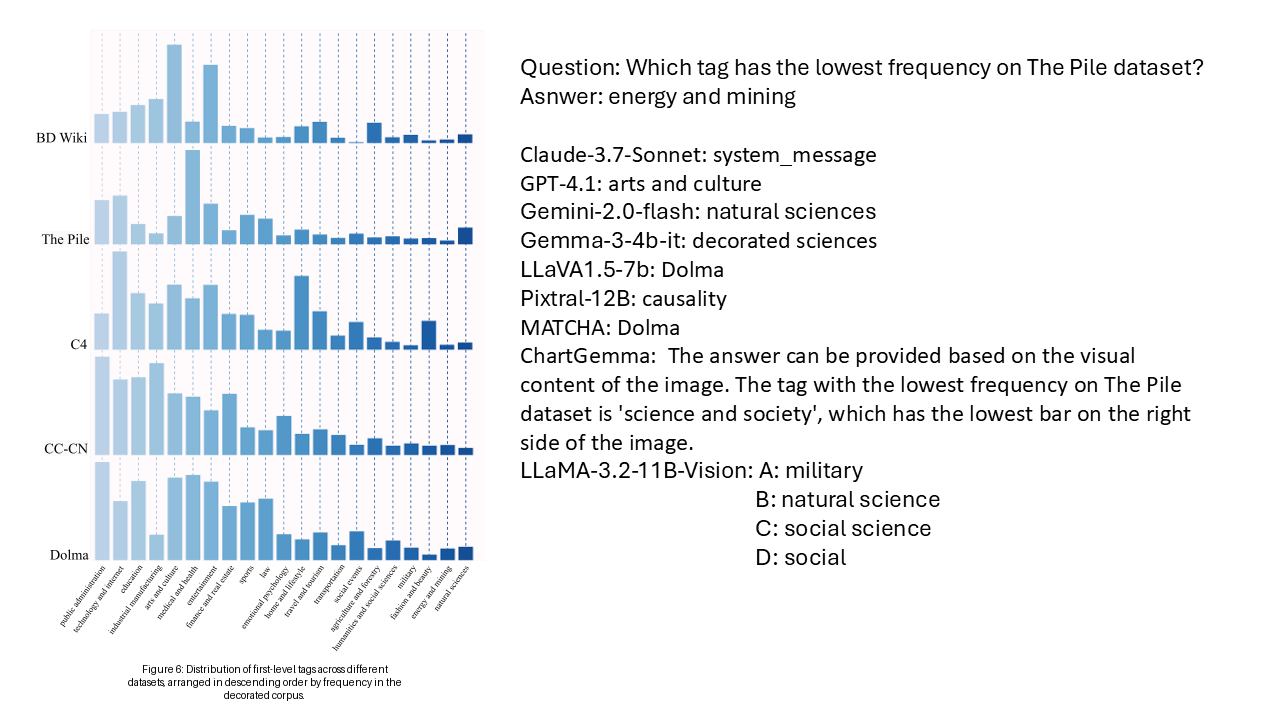}
    \caption{
        Examples of wrong answers by MLMs from PolyChartQA. The multi-chart is from \citet{zhao-etal-2024-decoratelm}.}
    \label{error_2}
\end{figure*}

\end{document}